%% file: gph_arxiv.tex
\documentclass[11pt,letterpaper]{article}
\newcommand{\arxivroot}{}
\IfFileExists{arxiv/layout.tex}{\renewcommand{\arxivroot}{arxiv/}}{}
\input{\arxivroot layout}
\usepackage[utf8]{inputenc}
\usepackage[T1]{fontenc}
\usepackage{amsmath,amssymb,amsfonts,amsthm}
\usepackage{booktabs}
\usepackage{graphicx}
\usepackage{microtype}
\usepackage{nicefrac}
\usepackage{xcolor}
\usepackage{enumitem}
\usepackage{multirow}
\usepackage{tabularx}
\usepackage{longtable}
\usepackage{float}
\usepackage{placeins}
\usepackage{wrapfig}
\usepackage{needspace}
\usepackage{hyperref}
\usepackage{url}
\hypersetup{hidelinks}
\graphicspath{{\arxivroot figs/}}
\newtheorem*{goldenpathhypothesis}{Golden Path Hypothesis}
\newcommand{\norm}[1]{\left\lVert#1\right\rVert}

\title{\textbf{The Golden Path Hypothesis:\\ Reusable Schedules in Diffusion Caching}}
\input{\arxivroot authors}

\begin{document}
\raggedbottom
\maketitle

\begin{abstract}
Diffusion caching accelerates generation by replacing transformer computation with cached or predicted features at selected denoising steps. We introduce the \emph{Golden Path Hypothesis} (GPH): under fixed inference conditions, prompt-independent cache schedules can achieve final-output quality comparable to the best prompt-specific schedules across prompts. We investigate the GPH across ten caching methods, four image and video models, and three cache ratios. Prompt-adaptive methods repeatedly select a small number of schedules, and reusing their most frequent schedules on new prompts closely matches the quality of prompt-specific choices. Exhaustive evaluation of 1.4 million schedules on four examples further identifies prompt-independent schedules that remain competitive on unseen prompts. To explain this transfer, we analyze denoising trajectories and the accumulation of caching errors. Latent-state trajectories exhibit similar structures across datasets and seeds, while an exact error decomposition shows that accumulated effects of earlier errors predict final latent-state error better than local approximation errors. This motivates searching for end-to-end schedules using final-output quality. With only a small set of examples, the resulting golden paths transfer across prompts and datasets, and can be tuned to the desired quality objective, including reconstruction fidelity or perceptual similarity.
\end{abstract}

\section{Introduction}
\label{sec:introduction}

Diffusion and flow-matching models generate images and videos through repeated
model evaluations \citep{ho2020denoising, lipman2022flow}. Diffusion caching
reduces this cost by reusing intermediate features \citep{ma2024deepcache} or
predicting them from earlier evaluations \citep{taylorseer2025}. A cache schedule
specifies which denoising steps use full computation or an approximation.
Because intermediate features depend on the prompt and noise
seed, many caching methods choose these steps adaptively during generation.
This suggests that high-quality schedules depend on the input. Yet
offline methods search for a schedule before generation and reuse it across
prompts, achieving high output quality in our image and video
evaluations~\citep{budcache2026,meancache2026}.
Our large-scale analysis shows that the search objectives used by existing
offline methods can be improved.
The approximation policy also affects a shared schedule's quality.
Existing work lacks a systematic analysis of these choices.
More fundamentally, it remains unclear why schedules found offline work
across prompts.
We therefore ask a basic question:
\emph{how much does a high-quality cache schedule actually need
to change across prompts?}
See Section~\ref{sec:related-work} for a detailed discussion of related work.

\begin{figure}[t]
  \centering
  \includegraphics[width=\linewidth]{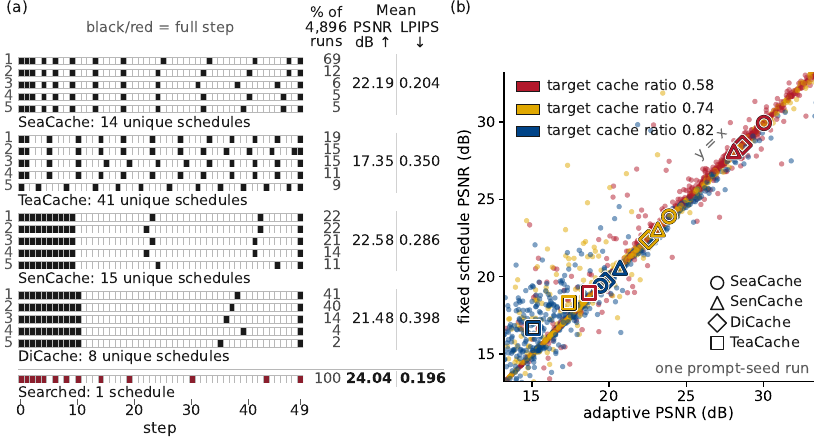}
  \let\normalsize\small
  \caption{\textbf{Shared schedules retain quality across prompts.}
  Both panels use FLUX.1-dev on PartiPrompts, with full-compute outputs
  for the same prompts and seeds as references.
  (a) Each method's five most frequent schedules are shown at a
  74\% target cache ratio.
  Black or dark-red cells mark full steps, and white cells mark cached steps.
  Percentages give schedule frequencies across 4,896 prompt--seed runs.
  The searched schedule is reused in every run.
  PSNR and Learned Perceptual Image Patch Similarity (LPIPS)~\citep{zhang2018lpips}
  are averaged over all runs for each method or searched schedule.
  Higher PSNR and lower LPIPS are better.
  Adaptive methods retain their approximation policies, while search uses
  SeaCache's feature reuse.
  (b) We reuse each method's most frequent schedule on 1,088 prompts excluded
  from selection, keeping its approximation policy unchanged.
  Small points show up to 300 runs per method and ratio where schedules differ.
  Large markers show mean PSNR over all such runs.
  Colors denote target cache ratios and large-marker shapes denote methods.
  The diagonal marks equal PSNR.}
  \label{fig:main}
\end{figure}

We propose the \emph{Golden Path Hypothesis} (GPH): under fixed inference
conditions, prompt-independent schedules can achieve final-output
quality comparable to the best prompt-specific schedules across prompts. Our
experiments provide several forms of evidence for this hypothesis. First,
prompt-adaptive methods repeatedly select a small number of schedules. For
example, SeaCache selects only 14 different schedules across 4,896 prompt--seed runs on
FLUX.1-dev at a 74\% cache ratio (Figure~\ref{fig:main}(a)). More generally, reusing an
adaptive method's most frequent schedule on unseen prompts closely matches the
quality of its prompt-specific choices (Figure~\ref{fig:main}(b)).
For each tested model and cache ratio, a shared schedule stays within
1.0~dB of each prompt's best evaluated PSNR on a majority of prompts,
as reported in Section~\ref{sec:shared}. Finally,
we exhaustively evaluate 1.4 million schedules on four examples and identify
a single schedule that remains competitive on thousands of unseen
prompt--seed runs, as shown in Section~\ref{sec:landscape} and
Figure~\ref{fig:exhaustive-existence}.

Why can the same schedule work for different inputs? We first examine
full-compute denoising trajectories obtained without caching.
Within each model, relative displacement, state norm, and changes in
direction evolve similarly across datasets and seeds.
Direction changes are smaller in the middle of denoising than near its ends.
Step displacements normalized by the distance between each trajectory's
initial and final states also vary little across prompts over much of denoising.
These patterns persist even though the trajectories have different spatial
directions.
These common changes support the use of shared cache schedules.
To evaluate such a schedule, we also need to understand how caching changes
the generation process. We derive an exact decomposition of the error in
the model output used to update the latent state at each step.
It separates the error introduced by the
approximation at that step from the effect of state changes caused by
earlier cached steps. The accumulated effects of these state changes
correlate much more strongly with final latent-state error than the errors
introduced by the individual approximations. These findings motivate scoring
schedules by final-output quality rather than accumulating prediction-error
costs as done in DPCache and MeanCache~\citep{dpcache2026,meancache2026}.

We then search for \emph{golden paths} on a small number of examples,
scoring candidates by final-output quality.
The resulting schedules retain their quality on unseen
prompts across four image datasets. Optimizing perceptual similarity instead
of pixel fidelity selects different schedules and improves perceptual
similarity to full-compute outputs on new prompts.

Our main contributions are:
\begin{itemize}[leftmargin=1.5em,itemsep=1pt,topsep=2pt]
    \item We formulate and test the \emph{Golden Path Hypothesis}.
    Large-scale comparisons and exhaustive evaluation show that shared
    schedules can approach each prompt's best evaluated output quality.

    \item We identify common patterns of relative step displacement and
    direction change in full-compute trajectories across prompts, datasets,
    and seeds within each model. These patterns provide empirical support
    for the GPH even when the trajectories have different spatial directions.

    \item We derive an exact decomposition of the error in the model output
    at each step and analyze how caching errors propagate to the final output.
    This analysis guides the scoring of candidate golden paths by final-output
    quality.
    We also show that changing the approximation policy can
    improve quality on a fixed schedule, and that this effect depends on the
    schedule.

    \item Motivated by the GPH, we search on a small set of examples using
    different output-quality objectives and find golden paths that retain
    high output quality on unseen prompts and datasets.
\end{itemize}

\section{The Golden Path Hypothesis}
\label{sec:setup}

A \emph{full step} computes the model's output without caching.
The schedule specifies the full steps and \emph{cached steps}.
At a cached step, an \emph{approximation policy} specifies how features or
model outputs are obtained in place of full model computation.
For example, it may reuse a stored feature or predict a feature from
earlier computations.
The \emph{cache ratio} is $K/N$ for $K$ cached steps out of $N$ total steps.
The \emph{inference conditions} comprise the model, sampler, sequence of
noise levels, cache ratio, approximation policy, and quality metric.
The quality metric used to rank schedules is the \emph{objective}.
\begin{goldenpathhypothesis}
Under fixed inference conditions, prompt-independent schedules can
achieve final-output quality comparable to the best prompt-specific schedule
across prompts.
\end{goldenpathhypothesis}
We call such a prompt-independent schedule a \emph{golden path}.
The comparison is with each prompt's best schedule among all schedules
that satisfy these inference conditions.
On new prompts, we compare shared schedules' output quality with
each prompt's best evaluated quality.

\noindent\textbf{Evaluation protocol.}
All experiments use 50 denoising steps. The main schedule comparisons target
cache ratios 0.58, 0.74, and 0.82, corresponding to $K=29$, $37$, and $41$.
The baseline comparisons retain each method's original approximation policy.
Schedule reuse retains the same approximation policy.
Our search experiments and tests against each prompt's best evaluated
schedule use \emph{residual reuse}.
The residual is the change in hidden features across the transformer blocks.
A cached step adds the residual from the most recent full step to its
current input features.
We calibrate each adaptive method to achieve a \emph{target cache ratio}
on average across prompt--seed runs. The \emph{realized cache ratio} is the
ratio achieved in an individual run.

We evaluate FLUX.1-dev~\citep{flux2024} and Qwen-Image~\citep{qwenimage2025}
on DrawBench~\citep{saharia2022drawbench},
GenEval-style~\citep{ghosh2023geneval}, PartiPrompts~\citep{yu2022parti}, and
DiffusionDB-clean10k~\citep{wang2022diffusiondb}.
We evaluate the video models HunyuanVideo~\citep{hunyuanvideo2024} and
Wan2.1~\citep{wan2025} on Penguin599~\citep{penguin2025} and
VBench944~\citep{vbench2024}.
The image and video baseline comparisons use three seeds per prompt.
PSNR and SSIM~\citep{wang2004ssim} increase as the cached output approaches its
full-compute reference.
LPIPS~\citep{zhang2018lpips} decreases.
Appendix~\ref{app:cache-experiment} describes the datasets, model configurations,
calibration procedures, and other experimental details.

A \emph{quality margin} is an allowed loss relative to a prompt's best
evaluated schedule, measured in the units of the quality metric.
We measure \emph{coverage} as the fraction of prompts on which a schedule
stays within that margin.
For each test, the \emph{candidate set} contains the schedules included in that comparison
under the same model, cache ratio, and approximation policy.
We report PSNR coverage at margins 0.25, 0.5, and 1.0 dB.
Appendix~\ref{app:coverage-margins} relates these margins to the quality change from
one additional cached step and gives the SSIM and LPIPS margins.

\section{Shared Schedules Recur and Transfer}
\label{sec:shared}
\label{sec:concentration}

We test the GPH through two comparisons of output quality.
We compare shared schedules with schedules chosen adaptively for each
prompt and with each prompt's best evaluated schedule.
We study SeaCache~\citep{seacache2026}, TeaCache~\citep{teacache2025},
SenCache~\citep{sencache2026}, and DiCache~\citep{dicache2025}, which choose
cached steps during generation.
Their choices concentrate on a small number of schedules, as illustrated
in Figure~\ref{fig:main}(a).
Our image study contains 891,720 prompt--seed runs across 96 combinations of
method, model, dataset, and cache ratio.
The four datasets include prompts about object counts, attributes, spatial
relations, and visual styles, as well as user-written prompts from DiffusionDB.
Across these 96 combinations, the fraction of runs using one of the three
most frequent schedules has a median of 89.8\% and a range of 40.7--100\%.
The decisions repeat across prompts despite being made separately for each run.

\subsection{Shared schedules retain quality on new prompts}
\label{sec:fixed-replay}

We first test how output quality changes when the same schedule replaces
prompt-specific adaptive decisions.
For each image model, method, and cache ratio, we select the most frequent
schedule with the target cache count on 544 prompts from PartiPrompts and
reuse it on the remaining 1,088 prompts.
We keep the approximation policy unchanged to compare the two ways of
choosing cached steps.
We compare outputs for the same prompt and seed, using runs where the
shared and adaptive schedules differ.
The large markers in Figure~\ref{fig:main}(b) lie close to the diagonal for
FLUX.1-dev: the two ways of choosing schedules produce similar mean PSNR.
The absolute difference between their mean PSNR values has a median of
0.10 dB across the 24 model--method--ratio combinations.
Mean PSNR with the shared schedule ranges from 0.81 dB lower to 1.56 dB
higher than with the adaptive method.
Table~\ref{tab:fixed-replay} summarizes the corresponding changes in
PSNR, SSIM, and LPIPS on the prompts excluded from schedule selection.

\input{\arxivroot tables/fixed_replay_main}

We repeat the comparison on HunyuanVideo and Wan2.1 using 150 prompts from
each of Penguin599 and VBench944, with one seed per prompt.
The most frequent schedules remain unchanged when selected using only
prompts outside this evaluation set.
We retain only runs where the schedules differ and the shared schedule
has the target cache count, leaving 19 model--method--ratio combinations.
In 17 of these 19 combinations, the shared schedule's mean PSNR is at most 0.25 dB below
the adaptive method's mean.
Shared schedules can therefore retain the mean quality of adaptive methods
on new image and video prompts.

\subsection{A shared schedule approaches each prompt's best evaluated quality}
\label{sec:landscape}

How closely can one shared schedule approach each prompt's best evaluated
output quality?
For FLUX.1-dev at cache ratio 0.82, we exhaustively evaluate 1,370,754
schedules on four prompts from PartiPrompts, using one seed per prompt.
We select the schedule with the highest mean PSNR across these four runs.
Table~\ref{tab:k41-per-run-optima} compares this shared schedule with
each selection run's best schedule in the fully enumerated space.
Its PSNR gap is at most 0.269 dB in three selection runs and
1.249 dB in the fourth.

\input{\arxivroot tables/k41_per_run_optima}

\begin{figure}[!htbp]
  \centering
  \includegraphics[width=0.64\linewidth]{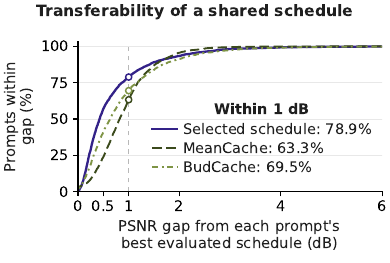}
  
  \caption{\textbf{A shared schedule transfers to new prompts.}
  For FLUX.1-dev at cache ratio 0.82, curves show the percentage of 2,381 prompts
  within each gap. We average each schedule's PSNR over three seeds for each prompt.
  The gap is the highest mean PSNR among 337 candidates minus the mean PSNR
  of the schedule shown.
  All three schedules use residual reuse. Circles mark 1 dB.}
  \label{fig:exhaustive-existence}
\end{figure}

We then evaluate this shared schedule on 2,381 new prompts from DrawBench,
GenEval-style, and PartiPrompts.
We combine high-scoring schedules from the exhaustive search,
uniformly sampled schedules, and control schedules such as those from
MeanCache and BudCache.
Removing duplicates yields 337 candidate schedules.
For each prompt, we compare the shared schedule's output PSNR with the
highest PSNR among these candidates.
Appendix~\ref{app:k41-search} details the candidate construction.
Figure~\ref{fig:exhaustive-existence} shows that the selected schedule stays
within 0.5 dB of the best evaluated PSNR on 57\% of the new prompts and
within 1.0 dB on 79\%.
A schedule selected on a few prompts can therefore approach the best
evaluated prompt-specific schedules on new prompts.

Appendices~\ref{app:schedule-distribution}--\ref{app:replay-videos} report
recurrence across seeds and datasets, additional quality metrics, and random
schedule controls.
In the coverage tests of Appendix~\ref{app:coverage}, a schedule selected
for each of four models and three cache ratios achieves majority coverage
at 1.0 dB. All 12 lower confidence bounds exceed 50\%.
Appendix~\ref{app:k41-search} gives the four-prompt selection rule and
search-space constraints.

\input{\arxivroot full_trajectory}

\section{Evaluating Schedules by Final-Output Quality}
\label{sec:mechanism}

The common patterns in Section~\ref{sec:trajectory} describe denoising
without caching. To compare candidate golden paths, we now examine how
reusing or predicting features changes later computations and the final
output. We ask what a quality score needs to capture when evaluating a
complete schedule. We define the \emph{output error} as the difference
between the cached and full-compute final latent states for the same
prompt and seed, and measure its magnitude using the L2 norm.

\subsection{Current approximations and earlier errors}
\label{sec:action-state}

\begin{figure}[!htbp]
  \centering
  \includegraphics[width=0.54\linewidth]{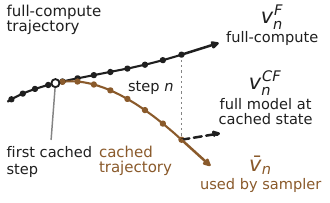}
  
  \caption{\textbf{The velocity error has current and state contributions.}
  Schematic illustration. Dots mark states. The dotted line joins the pair at cached step $n$.
  Arrows show velocity predictions.}
  \label{fig:action-state}
\end{figure}

How do current approximation errors and the effects of earlier cached
steps relate to output error?
Figure~\ref{fig:action-state} compares the full-compute state $z_n^F$ and
cached state $\widetilde z_n$ at denoising step $n=0,\ldots,49$, before
the update, for the same prompt and seed.
Earlier approximations affect the cached state.
The full model predicts velocity $v_n^F$ at $z_n^F$.
At $\widetilde z_n$, a full step would produce $v_n^{CF}$,
with all earlier cached steps unchanged.
The cached run uses $\bar v_n$ to update $\widetilde z_n$ to $\widetilde z_{n+1}$.
At a cached step, the approximation policy produces $\bar v_n$,
and we compute $v_n^{CF}$ separately for comparison.
At a full step, $\bar v_n=v_n^{CF}$.

The \emph{velocity error}, $\bar v_n-v_n^F$, has the exact decomposition:
\begin{equation}
 \bar v_n-v_n^F
 =\underbrace{\bar v_n-v_n^{CF}}_{a_n:\ \text{current contribution}}
 +\underbrace{v_n^{CF}-v_n^F}_{g_n:\ \text{state contribution}}.
 \label{eq:action-state}
\end{equation}
The current contribution $a_n$ compares approximate and full-model
predictions at the same cached state.
The state contribution $g_n$ compares full-model predictions at
the cached and full-compute states.
At a full step, $a_n=0$.

We measure both contributions on 100 PartiPrompts with FLUX.1-dev,
SeaCache, and TeaCache at cache ratios near 0.58 and 0.82.
For each prompt, we sum the magnitude of each contribution over denoising steps.
Across the four method--ratio combinations, the Spearman correlation of
the summed state contribution, $\sum_n\norm{g_n}$, with output-error magnitude
ranges from 0.988 to 0.996.
For the summed current contribution, $\sum_n\norm{a_n}$, it ranges from
0.579 to 0.748.
The summed state contribution is more strongly associated with output
error than the summed current contribution.
Earlier cached steps can affect the model's prediction even at later full
steps.
Appendix~\ref{app:closed-loop} gives the state recursion.

\subsection{Propagation from one cached step}
\label{sec:isolated-action}

\begingroup
\displaywidowpenalty=10000
We next isolate one cached step to trace how its feature error reaches the
final output. This explains which effects a local feature-error score omits.
We replace a feature only at step $k$ of a full-compute trajectory.
Let $d_k^F$ be that feature and $\widehat d_k$ its cached approximation.
Write their difference as $r_k=\widehat d_k-d_k^F$.
Let $B_k$ map feature error to velocity error.
Let $H_k$ map velocity error to the error in the next latent state through
the sampler update.
For a one-step sampler, let $J_j$ be the Jacobian of a later full-compute
transition.
Let $e_{50}^{(k)}$ be the output error with only step $k$ cached.
To first order,
\begin{equation}
 e_{50}^{(k)}\approx
 \underbrace{J_{49}\cdots J_{k+1}}_{\Phi_{50,k+1}}H_kB_kr_k.
 \label{eq:pathwise}
\end{equation}
Thus, output error depends on both the feature error and its propagation
through the model, sampler, and remaining steps.
\par\endgroup

The experiment uses 100 DrawBench prompts and three approximation policies
on FLUX.1-dev: residual reuse, first-order Taylor prediction, and
second-order Hermite prediction.
We measure two gains: the velocity-error norm divided by the feature-error
norm, and the output-error norm divided by the next-state error norm.
At each step, we average each gain over the other 99 prompts with the same
approximation policy, then multiply the two means by the Euler step-size
magnitude to obtain the propagation gain $h_k^{\mathrm{LOO}}$.
LOO denotes leave-one-prompt-out estimation.
Appendix~\ref{app:pathwise-derivation} gives the derivation and measurement procedure.

\input{\arxivroot tables/four_factor_ladder}

Table~\ref{tab:four-factor-ladder} shows that scores including
propagation factors estimated from other prompts
rank the measured output errors more accurately than local feature error
alone.
The step-only score is $-k$, so earlier cached steps rank higher.

The single-step experiment starts from a full-compute state.
In a complete cached trajectory, earlier approximations change the states
at which later approximations are made.
Adding scores measured independently on the full-compute trajectory would
omit this dependence on earlier choices.
Scoring final outputs includes these effects when comparing candidate golden paths.

\subsection{The approximation policy changes schedule quality}
\label{sec:spx}

\begin{figure}[!htbp]
  \centering
  \includegraphics[width=\linewidth]{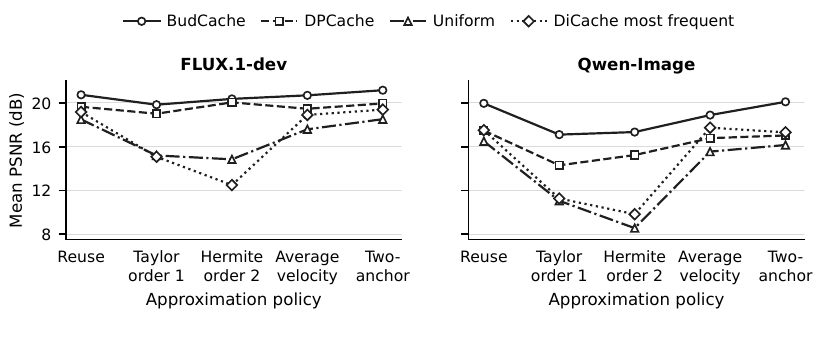}
  \caption{\textbf{The effect of changing the approximation policy depends on the schedule.}
  Both panels use cache ratio 0.82. Each curve holds one schedule fixed
  while changing the approximation policy. Points show mean PSNR over
  1,632 PartiPrompts and three seeds. Circles denote BudCache schedules,
  squares DPCache schedules, triangles uniform schedules, and diamonds
  DiCache's most frequent schedules.
  The policies are residual reuse, first-order Taylor prediction,
  second-order Hermite prediction, interval-average velocity prediction,
  and two-anchor prediction.
  Nonparallel curves show that the same change in approximation policy
  produces different PSNR changes across schedules.}
  \label{fig:schedule-policy-interaction}
\end{figure}

We test whether changing the approximation policy can improve output quality
on a fixed schedule, and whether the effect differs across schedules.
We compare four schedules under five approximation policies on FLUX.1-dev
and Qwen-Image, using 1,632 prompts from PartiPrompts and three seeds.
For FLUX.1-dev at cache ratio 0.58, keeping the BudCache schedule fixed and
replacing residual reuse with second-order Hermite prediction raises mean
PSNR by 0.48 dB.
At cache ratio 0.82 on FLUX.1-dev, replacing residual reuse with
second-order Hermite prediction lowers mean PSNR by 0.38 dB on the BudCache
schedule and by 6.68 dB on DiCache's most frequent schedule.
Figure~\ref{fig:schedule-policy-interaction} shows these comparisons
at cache ratio 0.82 for both models.

We use an additive model to measure this dependence on the schedule.
In this model, replacing one approximation policy with another changes
predicted PSNR by the same amount for every schedule.
We fit the model to the 20 mean PSNR values for each image model and cache ratio.
The differences between measured and fitted means quantify the interaction
between schedule and approximation policy.
At cache ratios 0.74 and 0.82, these differences account for 13--27\% of the
variation among the 20 means.
A schedule found by offline search can achieve higher mean PSNR
with a different approximation policy. The gain or loss from changing the
policy depends on the schedule.
We therefore evaluate a candidate golden path with the approximation policy
intended for generation.
Table~\ref{tab:spx-image-decomposition} reports the interaction results
for each model and cache ratio.

\section{Searching for Golden Paths}
\label{sec:search}

Motivated by the GPH, we test whether standard search can find
transferable schedules without exhaustive enumeration.
We score complete schedules on a few examples by final-output quality, following
Section~\ref{sec:mechanism}.
We also compare schedules and output quality under different objectives.

\noindent\textbf{Finding schedules that transfer.}
On FLUX.1-dev and Qwen-Image, we search under residual reuse using hill
climbing, simulated annealing, and greedy coordinate ascent, with random
search as a control.
For each model and cache ratio, we score candidates by mean PSNR on eight
prompt--seed runs. Among candidates from all four procedures, we select
the schedule with the highest mean PSNR on 50 separate COCO validation
prompts~\citep{lin2014coco}.
The scoring and validation prompts are separate from the evaluation datasets.
We evaluate each selected schedule on the four image datasets with three
seeds per prompt.
See Appendix~\ref{app:search} for search and selection details.

\input{\arxivroot tables/search_paired_main}

The searched schedules have higher mean PSNR than BudCache and random search
in every comparison in Table~\ref{tab:search-summary}.
They also improve on MeanCache in 19 of the 24 comparisons.
In the remaining five, mean PSNR is only 0.02--0.09 dB lower than MeanCache's.
Thus, a small set of scoring examples can identify schedules that remain
useful across a much larger set of prompts.

\noindent\textbf{Choosing the quality objective.}
Existing offline methods use different objectives.
DPCache~\citep{dpcache2026} accumulates feature prediction errors, while
MeanCache~\citep{meancache2026} uses errors in predicting the average velocity
over a denoising interval.
Both measure these errors on full-compute trajectories, which omit the state
changes caused by earlier cache approximations.
BudCache~\citep{budcache2026} instead scores complete cached generations
using final latent MSE.
Latent MSE measures a different quantity from pixel fidelity or perceptual similarity.

We repeat hill climbing and simulated annealing using LPIPS instead of
PSNR, while keeping the approximation policy fixed.
For each procedure, we compare its LPIPS-selected schedule with its
PSNR-selected schedule.
This gives 48 comparisons across two procedures, two models, three cache
ratios, and four datasets.
Table~\ref{tab:search-objectives} shows that the LPIPS-selected schedules
achieve lower mean LPIPS and higher mean SSIM
than the PSNR-selected schedules in all 48 comparisons.
Their mean PSNR is higher in 15 comparisons and lower in the remaining 33.
Golden-path search can therefore improve perceptual similarity on new
prompts by changing the output-quality objective.

\input{\arxivroot tables/search_objective_main}

\section{Related Work}

\label{sec:related-work}

\paragraph{Feature reuse and prediction.}
DeepCache reuses U-Net features~\citep{ma2024deepcache}.
FORA~\citep{selvaraju2024fora} and $\Delta$-DiT~\citep{deltadit2024}
extend feature or residual reuse to diffusion transformers.
ToCa selects cacheable tokens within layers~\citep{toca2025}.
For feature prediction, TaylorSeer uses finite differences~\citep{taylorseer2025}, while
HiCache uses Hermite interpolation~\citep{hicache2026}.
L2P instead learns the predictor~\citep{l2p2026}.
FoCa uses multistep prediction and correction~\citep{foca2026}.
HyCa assigns solvers to feature dimensions~\citep{hyca2026}.
SVD-Cache separates principal and residual subspaces~\citep{svdcache2026}.
RFC estimates output changes from module inputs~\citep{rfc2026}.
DiCache extrapolates from the two most recent full-step residuals, using the
first transformer block's output to guide the extrapolation~\citep{dicache2025}.
We extend this line of work by studying how approximation policies and
schedules jointly affect output quality. Controlled comparisons show
that changing the policy can improve output quality without changing the
schedule, and that the effect depends on the schedule. This identifies
policy choice as another way to improve the output quality of candidate
golden paths.

\paragraph{Schedule selection.}
TeaCache~\citep{teacache2025}, SeaCache~\citep{seacache2026},
SenCache~\citep{sencache2026}, and DiCache~\citep{dicache2025}
select cached steps during generation.
DPCache~\citep{dpcache2026} selects fixed schedules using accumulated
feature prediction errors.
MeanCache~\citep{meancache2026} uses errors in predicted interval-average velocity.
Both measure these errors on full-compute trajectories and omit
the state changes caused by earlier cache approximations.
BudCache~\citep{budcache2026} selects its cache schedule using final latent MSE.
OTCache interpolates schedules across computation limits after searching
with an LPIPS objective~\citep{otcache2026}.
ReCache uses reinforcement learning to select schedules under compute
constraints~\citep{recache2026}.
We extend the study of reusable schedules by testing the GPH through
large-scale comparisons of the quality obtained with shared and
prompt-specific schedules. Shared schedules can approach each prompt's
best evaluated quality. Our error analysis explains why candidate
schedules should be evaluated by final-output quality. Controlled
comparisons then show how the choice of output-quality objective affects
the schedules selected and their quality on new prompts.

\paragraph{Trajectory geometry.}
\citet{chen2024trajectory} use diffusion trajectory geometry to choose
sampling times.
OmniCache studies cache-specific curvature~\citep{chu2025omnicache}.
Fresco instead reduces spatial resolution at early steps~\citep{fresco2026}.
Our large-scale study of full-compute trajectories covers image and video
models across prompts, datasets, and seeds. Within each model, relative
step displacements and direction changes follow similar patterns even
when spatial directions differ. These shared patterns provide empirical
support for the GPH across different inputs.

\section{Discussion and Conclusion}
\label{sec:discussion}

\noindent\textbf{Discussion and conclusion.}
Our experiments support the GPH: shared schedules can achieve quality
comparable to the best evaluated prompt-specific schedules.
Common step-by-step denoising patterns within each model provide empirical
support for this reuse, even when trajectories differ in spatial direction.
Our error analysis motivates scoring complete schedules by final-output
quality, since local approximation errors omit the effects of earlier
state changes.
The effect of an approximation policy also depends on the schedule, so
evaluation should use the policy intended for generation.
Searches on a few examples demonstrate a practical use of the GPH.
They find schedules that transfer to new prompts, and changing the
objective can improve the chosen quality metric on those prompts.
Future work will study whether the distribution of golden paths in the
schedule space can guide searches that require fewer candidate evaluations.
We will consider evolutionary search and ant colony optimization.

\textbf{Limitations.}
Our experiments cover transformer-based image and video models.
We have not tested the GPH on non-transformer diffusion models or other data modalities.
We have not systematically studied how the number and diversity of scoring
examples affect output quality on new prompts.
We have not systematically characterized the distribution of golden paths
or the distances between them.
Whether small schedule changes can connect them at a fixed cache ratio
while preserving high output quality across prompts remains an open question.

\subsection*{AI Disclosure}

We used Claude Code and Codex with Fable 5.1 and GPT-5.6 to
implement baseline methods and experimental code, write Slurm scripts,
clean and reformat data, and produce figures.
For English drafting and editing, we used Fable 5, Fable 5.1, GPT-5.6,
and GPT-6.

We initially used AI agents to write all experimental code.
The authors independently made the original observations, formulated the GPH,
designed the methodology, developed the mathematical claims, and interpreted
the results.
AI tools assisted with subsequent discussion and review.
The authors made the final scientific decisions and specified the experimental
conditions and scope.
We recorded our exploratory experiments in hundreds of Markdown files.
The authors specified each experiment's design and procedure in these files,
and AI agents implemented the experiments and submitted jobs to the
compute clusters.
The agents updated the corresponding files with results under the authors'
authorization.
The authors substantially rewrote the initial AI-edited English drafts.
We take responsibility for the final paper, code, and reported results.

\subsection*{Reproducibility Statement}

Code and instructions for reproducing the experiments are available at\\
\url{https://github.com/nanguoyu/Golden-Path-Hypothesis}.
The repository includes model-specific implementations, generation and
evaluation scripts, and the search and analysis code used in this paper.
It also provides environment setup and data preparation instructions.

\bibliography{\arxivroot cache_refs}
\bibliographystyle{plainnat}

\appendix
\arxivappendixlayout
\raggedbottom
\input{\arxivroot appendix}

\end{document}

%% file: layout.tex
% Independent preprint layout. This file is maintained directly.
\usepackage[letterpaper,margin=1in]{geometry}
\usepackage{times}
\usepackage[round,authoryear]{natbib}
\setcitestyle{authoryear,round,citesep={;},aysep={,},yysep={;}}

\setlength{\parindent}{1em}
\setlength{\parskip}{0.3em}
\setlength{\floatsep}{12pt plus 2pt minus 2pt}
\setlength{\textfloatsep}{16pt plus 3pt minus 3pt}
\setlength{\intextsep}{14pt plus 2pt minus 2pt}
\setlength{\abovecaptionskip}{7pt}

\setcounter{topnumber}{6}
\setcounter{bottomnumber}{6}
\setcounter{totalnumber}{10}
\pagestyle{plain}
\date{}

% Use the standard article caption with a slightly smaller type size.
\makeatletter
\newcommand{\arxivappendixlayout}{%
  \setlength{\parskip}{0.2em}%
  \renewcommand\section{\@startsection{section}{1}{\z@}%
    {-2.5ex plus -.5ex minus -.2ex}{1ex plus .2ex}%
    {\normalfont\Large\bfseries}}%
  \renewcommand\subsection{\@startsection{subsection}{2}{\z@}%
    {-2ex plus -.5ex minus -.2ex}{.75ex plus .2ex}%
    {\normalfont\large\bfseries}}%
  \renewcommand\subsubsection{\@startsection{subsubsection}{3}{\z@}%
    {-1.75ex plus -.5ex minus -.2ex}{.75ex plus .2ex}%
    {\normalfont\normalsize\bfseries}}}
\setlength{\@fptop}{0pt}
\setlength{\@fpsep}{14pt plus 2pt minus 2pt}
\setlength{\@fpbot}{0pt plus 1fil}
\AtBeginDocument{%
  \setlength{\LTpre}{6pt plus 1pt minus 1pt}%
  \setlength{\LTpost}{8pt plus 1pt minus 1pt}%
  \def\LT@makecaption#1#2#3{%
    \LT@mcol\LT@cols c{\hbox to\z@{\hss\parbox[t]\LTcapwidth{%
      \reset@font
      \sbox\@tempboxa{#1{#2: }#3}%
      \ifdim\wd\@tempboxa>\hsize
        #1{#2: }#3%
      \else
        \hbox to\hsize{\hfil\box\@tempboxa\hfil}%
      \fi
      \endgraf\vskip4pt}%
    \hss}}}}
\long\def\@makecaption#1#2{%
  \vskip\abovecaptionskip
  \begingroup\small
  \sbox\@tempboxa{\textbf{#1:} #2}%
  \ifdim\wd\@tempboxa>\hsize
    \textbf{#1:} #2\par
  \else
    \global\@minipagefalse
    \hb@xt@\hsize{\hfil\box\@tempboxa\hfil}%
  \fi
  \endgroup
  \vskip\belowcaptionskip}
\makeatother

%% file: authors.tex
% Author order supplied by the authors. Academic contacts checked against
% FOSL (CPAL 2026), GRAIL (CPAL 2026), and institutional staff pages.
\author{%
\begin{tabular}{c}
Dong Wang\textsuperscript{1}\enspace Wenwu Tang\textsuperscript{1}\enspace Francesco Corti\textsuperscript{1}\enspace
Yun Cheng\textsuperscript{2}\enspace Lothar Thiele\textsuperscript{3}\enspace Olga Saukh\textsuperscript{1,4}\\[0.6em]
\small\textsuperscript{1}Graz University of Technology, Austria\\
\small\textsuperscript{2}Swiss Data Science Center, Switzerland\quad
\textsuperscript{3}ETH Zurich, Switzerland\\
\small\textsuperscript{4}Complexity Science Hub, Austria\\[0.4em]
\footnotesize\texttt{\{dong.wang,wenwu.tang,francesco.corti,saukh\}@tugraz.at}\\
\footnotesize\texttt{yun.cheng@sdsc.ethz.ch}\quad\texttt{thiele@tik.ee.ethz.ch}
\end{tabular}}
\hypersetup{pdfauthor={Dong Wang, Wenwu Tang, Francesco Corti, Yun Cheng, Lothar Thiele, Olga Saukh}}

%% file: tables/fixed_replay_main.tex
\begin{table}[!htbp]
\caption{\textbf{The most frequent schedules retain quality on new prompts.}
The evaluation uses the 1,088 prompts excluded from schedule selection.
For each metric, we subtract the value obtained with the adaptive method
from the value obtained with the fixed schedule on the same prompt and seed.
We average these differences within each model--ratio combination, using
only runs where the two schedules differ, with 34 to 2,901 runs per combination.
Each entry gives the median [minimum, maximum] of the six means from
two image models and three target cache ratios.
Arrows show which direction favors the fixed schedule.
Bold marks the largest median $\Delta$PSNR and $\Delta$SSIM and the
smallest median $\Delta$LPIPS.
Table~\ref{tab:fixed-replay-image-full} in Appendix~\ref{app:replay-images} gives each combination.
}
\label{tab:fixed-replay}
\centering
\small
\setlength{\tabcolsep}{3pt}
\renewcommand{\arraystretch}{1.15}
\begin{tabular}{@{}lrrr@{}}
\toprule
 & \multicolumn{3}{c}{Fixed $-$ adaptive} \\
\cmidrule(lr){2-4}
Method & \multicolumn{1}{c}{$\Delta$PSNR, dB $\uparrow$} & \multicolumn{1}{c}{$\Delta$SSIM $\uparrow$} & \multicolumn{1}{c@{}}{$\Delta$LPIPS $\downarrow$} \\
\midrule
SeaCache & $-0.03\;[-0.07, +0.02]$ & $-0.0011\;[-0.0025, +0.0034]$ & $+0.0009\;[-0.0019, +0.0049]$ \\
TeaCache & $\boldsymbol{+0.11}\;[-0.81, +1.56]$ & $\boldsymbol{+0.0011}\;[-0.0333, +0.0268]$ & $\boldsymbol{-0.0026}\;[-0.0390, +0.0333]$ \\
SenCache & $-0.02\;[-0.26, +0.30]$ & $+0.0008\;[-0.0182, +0.0122]$ & $-0.0014\;[-0.0153, +0.0203]$ \\
DiCache & $-0.06\;[-0.16, +0.18]$ & $0.0000\;[-0.0062, +0.0047]$ & $-0.0005\;[-0.0075, +0.0133]$ \\
\bottomrule
\end{tabular}
\end{table}

%% file: tables/k41_per_run_optima.tex
\begin{table}[!htbp]
\caption{\textbf{The shared schedule is within 0.269 dB of the best schedule
for three of the four selection runs.}
Each run uses FLUX.1-dev at cache ratio 0.82 with residual reuse.
The best PSNR is the maximum over all 1,370,754 schedules with steps
0, 1, 2, and 49 fixed as full computations.
The shared schedule maximizes mean PSNR across the four selection runs.
The gap is the best PSNR minus the shared schedule's PSNR, computed
before rounding. All quality values are in dB.}
\label{tab:k41-per-run-optima}
\centering
\small
\begin{tabular}{rrrrr}
\toprule
Prompt index & Seed & Best PSNR & Shared-schedule PSNR & Gap \\
\midrule
5 & 47 & 22.334 & 21.086 & 1.249 \\
8 & 50 & 28.621 & 28.590 & 0.030 \\
9 & 51 & 20.915 & 20.813 & 0.102 \\
15 & 57 & 20.765 & 20.496 & 0.269 \\
\bottomrule
\end{tabular}
\end{table}

%% file: full_trajectory.tex
\section{Common Patterns in Full-Compute Denoising}
\label{sec:trajectory}
\label{sec:shared-clock}

To understand why the shared schedules in Section~\ref{sec:shared} transfer
across prompts, we examine whether full-compute denoising trajectories share
step-by-step patterns.
We compare how latent states change and how trajectories turn, then examine
the shapes and directions of individual trajectories.
The measurements cover 74,310 image trajectories and 4,629 trajectories
from each video model.
Each dataset uses three noise seeds per prompt.
Appendix~\ref{app:sampling-conditions} gives the sampling configurations.

\subsection{Step-by-step changes across prompts}
\label{sec:trajectory-profiles}
\label{sec:trajectory-variation}

We first compare the relative distances that latent states move at the
same denoising step across prompts.
We measure how far each latent state lies from the line joining the initial
and final states, how far it moves at each step, and the state norm.
Let $z_n$ be the latent state after $n$ denoising steps.
There are 51 states, from initial noise $z_0$ to final state $z_{50}$.
The vector $\ell=z_{50}-z_0$ joins the endpoints.
Its length $L=\norm{\ell}$ is the \emph{chord length}, and
$\hat\ell=\ell/L$ is its direction.
The part of $z_n-z_0$ perpendicular to this line is the \emph{off-chord vector}
\begin{equation}
 \beta_n=(z_n-z_0)-\bigl((z_n-z_0)^\top\hat\ell\bigr)\hat\ell.
 \label{eq:off-chord}
\end{equation}
The normalized distance from the line is $\norm{\beta_n}/L$.
At step $n$, the normalized displacement is $\norm{z_{n+1}-z_n}/L$.
Both distances are expressed relative to the trajectory's chord length.
We measure the state norm as $\norm{z_n}/\sqrt{d}$, where $d$ is the number
of latent coordinates.
For each model and dataset, we also measure how much these quantities vary
across prompt--seed runs at each state or step.
We use the \emph{coefficient of variation}, abbreviated CV: the population
standard deviation divided by the mean across runs at that index.
We report CV as a percentage.

\begin{figure}[!htbp]
  \centering
  \includegraphics[width=\linewidth]{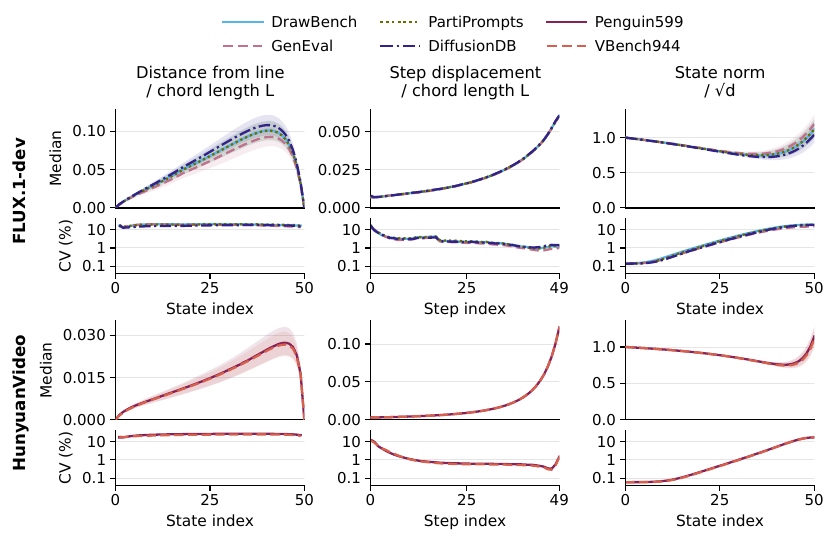}
  \caption{\textbf{Full-compute trajectories follow similar patterns across datasets
  within FLUX.1-dev and HunyuanVideo.}
  Rows show FLUX.1-dev and HunyuanVideo from top to bottom.
  Colors and line styles identify datasets within each model.
  Distance is measured from the line through each trajectory's endpoints.
  Distance and displacement are divided by each trajectory's chord length $L$.
  State norm is divided by $\sqrt{d}$, where $d$ is the latent dimension.
  Upper curves show medians across prompt--seed runs for each dataset,
  with bands spanning the middle 50\% at each index.
  Lower curves show the coefficient of variation across these runs at each index on a
  shared logarithmic percentage scale.
  Distance CV omits states 0 and 50, where it is undefined.
  GenEval and DiffusionDB denote GenEval-style and DiffusionDB-clean10k.
  FLUX.1-dev uses 37,155 prompt--seed runs from four datasets,
  and HunyuanVideo uses 4,629 runs from two datasets.
  Each dataset uses three seeds per prompt.
  Step $n$ updates state $n$ to $n+1$.}
  \label{fig:full-trajectory-summary}
\end{figure}

Within each model, the upper curves in
Figure~\ref{fig:full-trajectory-summary} show similar trends across
datasets, although their magnitudes differ.
Figure~\ref{fig:full-trajectory-summary-other-models} shows the
corresponding measurements for Qwen-Image and Wan2.1.
The distance from the line rises gradually and falls
rapidly near the end.
Step displacements generally increase, while the state norm first decreases
and then increases.
The profiles reflect both the model and its sampler.
The lower curves in Figure~\ref{fig:full-trajectory-summary} show similar
cross-prompt variation across these datasets.
Normalized step displacement has low variation over much of the trajectory.
The median CV over the 50 steps ranges from 1.94\% to 2.35\%
across the four FLUX.1-dev datasets and from 0.59\% to 0.63\%
across the two HunyuanVideo datasets.
Normalized step displacement varies more across prompts at the start.
State norms also vary little across prompts at the early and middle
states, with more variation near the final output.
Distances from the endpoint line vary more across prompts.
We conclude that normalized step displacements are similar across prompts
at most steps within each model. This consistency supports the GPH.

\subsection{Changes in direction}
\label{sec:trajectory-turning}

\begin{figure}[!htbp]
  \centering
  \includegraphics[width=\linewidth]{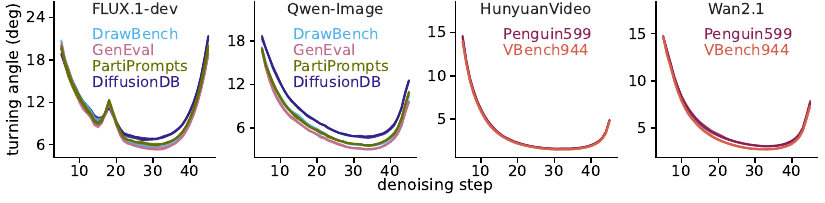}
  \caption{\textbf{Direction changes follow similar patterns across datasets
  within each model.}
  Curves show median turning angles across prompts for each dataset
  and seed. Colors identify datasets.
  }
  \label{fig:curvature-clock}
\end{figure}

We next compare how much trajectories change direction at the same
denoising steps across datasets and seeds.
The turning angle measures the change in direction around state $n$.
We compute it from the state changes over $w$ steps before and after $n$:
\begin{equation}
 \theta_n^{(w)}=\arccos
 \frac{(z_n-z_{n-w})^\top(z_{n+w}-z_n)}
 {\norm{z_n-z_{n-w}}\norm{z_{n+w}-z_n}}.
 \label{eq:turn-angle}
\end{equation}
We use $w=5$ and call the sequence of angles over denoising steps a
\emph{turning profile}.
Figure~\ref{fig:curvature-clock} shows smaller direction changes in the
middle than at the ends.
The FLUX.1-dev profiles also share a local peak near step 18.
Appendix~\ref{app:turning-precision} gives the sampling details and compares window widths.
The common step-by-step patterns therefore include both direction changes
and relative displacement.

\subsection{The shape and direction of individual trajectories}
\label{sec:trajectory-planes}

\begin{figure}[!htbp]
  \centering
  \includegraphics[width=\linewidth]{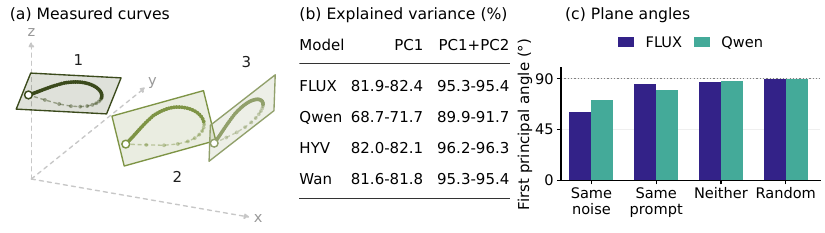}
  \caption{\textbf{Off-chord variation is low-dimensional, but its directions differ across inputs.}
  (a) Each FLUX.1-dev curve shows off-chord vectors in its own PC1--PC2
  coordinates, normalized by the full-compute trajectory's chord length.
  Solid lines show states 0--40, and pale dashed lines show states 40--50.
  Dots mark all 51 states.
  Open circles mark states 0 and 50.
  Planes are rotated and separated in display coordinates.
  (b) For each model, ranges give the minimum and maximum dataset medians
  of the percentage of centered off-chord variance explained by PC1 and
  by PC1 + PC2.
  (c) Bars show median first principal angles in the original latent space.
  Pairs share only initial noise, only the prompt, or neither.
  Colors identify models. The Random group shows random-plane references.
  The dotted line marks $90^\circ$.
  FLUX, Qwen, HYV, and Wan denote FLUX.1-dev, Qwen-Image,
  HunyuanVideo, and Wan2.1.
  Appendix~\ref{app:plane-samples} gives the sample construction and pair counts.}
  \label{fig:trajectory-geometry}
\end{figure}

We next ask whether the common patterns in normalized displacement and
turning angle also correspond to shared directions in latent space.
We use principal-component analysis, abbreviated PCA, to describe the
off-chord vectors of each trajectory.
For each trajectory, we subtract the mean of its 51 off-chord vectors and
fit the first two principal components, PC1 and PC2.
These components span a plane for the trajectory's off-chord vectors.
Across model--dataset combinations, the median fraction of centered
off-chord variance explained by these two components ranges from 89.9\%
to 96.3\%, as shown in Figure~\ref{fig:trajectory-geometry}(b).
Adding the chord direction gives a three-dimensional approximation of
the complete trajectory.
Figure~\ref{fig:trajectory-geometry}(a) shows three measured off-chord curves
whose projected endpoints coincide because the off-chord vector is zero
at both the initial and final states.

We then compare the directions of the fitted planes in the original latent
space.
Their \emph{first principal angle} is the smallest angle between a unit
direction in one plane and a unit direction in the other.
An angle near zero means that the planes share a direction.
An angle near $90^\circ$ means that they are nearly orthogonal.
When both the prompt and initial noise differ, the median first principal
angle is about $88^\circ$ in both image models.
In both image models, pairs that share initial noise have smaller median
first principal angles than pairs that share a prompt, as shown in
Figure~\ref{fig:trajectory-geometry}(c).
To summarize, trajectories within each model share patterns in normalized
displacement and turning angle, even when their off-chord planes have different orientations.
Together with the transfer results in Section~\ref{sec:shared}, these
common patterns provide empirical support for the GPH.

%% file: tables/four_factor_ladder.tex
\begin{table}[!htbp]
\caption{\textbf{Scores that include propagation rank output errors more accurately than local feature error alone.}
Entries give Spearman correlations between each score and the L2 distance
between the final latent states with and without caching, over 4,800 isolated
cached steps per approximation policy.
$\norm{r_k}$ is the feature-error norm, and $h_k^{\mathrm{LOO}}$ is the
propagation gain estimated from the other 99 prompts at the same step
and with the same approximation policy.
The step-only score is the negative cached-step index $-k$.
Bold marks the highest correlation in each row.}
\label{tab:four-factor-ladder}
\centering
\small
\begin{tabular}{lrrrr}
\toprule
\shortstack{Approximation\\policy} & \shortstack{Feature error\\$\norm{r_k}$} &
\shortstack{Propagation gain\\$h_k^{\rm LOO}$} &
\shortstack{Feature error $\times$ gain\\$\norm{r_k}h_k^{\rm LOO}$} &
\shortstack{Step only\\$-k$} \\
\midrule
Residual reuse & 0.528 & 0.803 & \textbf{0.886} & 0.867 \\
Hermite order 2 & 0.586 & 0.800 & \textbf{0.896} & 0.880 \\
Taylor order 1 & 0.676 & 0.788 & \textbf{0.899} & 0.880 \\
\bottomrule
\end{tabular}
\end{table}

%% file: tables/search_paired_main.tex
\begin{table}[!htbp]
\caption{\textbf{Searched schedules achieve higher or comparable mean PSNR
relative to all three reference schedules.}
For each prompt and seed, we subtract the reference schedule's output PSNR
from the searched schedule's output PSNR. Entries give the mean differences,
with positive values favoring the searched schedule.
All schedules use residual reuse.
The random-search schedule is selected from random search alone.
FLUX denotes FLUX.1-dev and Qwen denotes Qwen-Image.
DB, GE, PP, and DDB denote DrawBench, GenEval-style, PartiPrompts, and
DiffusionDB-clean10k, respectively.
We average over 600, 1,659, 4,896, and 30,000 prompt--seed runs, respectively.
Confidence
intervals and four additional metrics appear in Table~\ref{tab:search-paired-full}.}
\label{tab:search-summary}
\centering
\footnotesize
\setlength{\tabcolsep}{1.7pt}
\renewcommand{\arraystretch}{1.02}
\begin{tabular}{l@{\hspace{6pt}}c*{4}{r}@{\hspace{3pt}}*{4}{r}@{\hspace{3pt}}*{4}{r}}
\toprule
Model & Cache ratio & \multicolumn{12}{c}{Mean $\Delta$PSNR: searched $-$ comparison schedule, dB} \\
\cmidrule(lr){3-14}
 & & \multicolumn{4}{c}{MeanCache} & \multicolumn{4}{c}{BudCache} &
\multicolumn{4}{c}{Random search} \\
\cmidrule(lr){3-6}\cmidrule(lr){7-10}\cmidrule(lr){11-14}
 & & DB & GE & PP & DDB & DB & GE & PP & DDB & DB & GE & PP & DDB \\
\midrule
FLUX & 0.58 & +0.70 & +0.66 & +0.28 & +0.37 & +5.49 & +6.21 & +4.35 & +3.26 & +2.83 & +2.94 & +2.36 & +1.99 \\
FLUX & 0.74 & +0.49 & +0.50 & +0.36 & +0.32 & +1.14 & +1.17 & +0.91 & +0.74 & +0.57 & +0.58 & +0.57 & +0.40 \\
FLUX & 0.82 & +0.41 & +0.58 & +0.42 & +0.44 & +0.47 & +0.70 & +0.31 & +0.12 & +0.60 & +0.83 & +0.61 & +0.46 \\
\addlinespace[1pt]
Qwen & 0.58 & -0.02 & -0.05 & +0.04 & +0.09 & +1.19 & +0.84 & +1.08 & +1.15 & +3.95 & +3.71 & +3.93 & +3.88 \\
Qwen & 0.74 & +0.07 & -0.09 & -0.02 & -0.03 & +0.41 & +0.28 & +0.35 & +0.30 & +0.87 & +0.70 & +0.68 & +0.49 \\
Qwen & 0.82 & +0.70 & +0.96 & +0.79 & +0.71 & +1.57 & +1.99 & +1.51 & +1.06 & +1.36 & +1.52 & +1.62 & +1.57 \\
\bottomrule
\end{tabular}
\end{table}

%% file: tables/search_objective_main.tex
\begin{table}[!htbp]
\caption{\textbf{Optimizing LPIPS selects different schedules and improves
the targeted metric.}  Each entry gives the range across eight comparisons,
using hill climbing and annealing on four datasets.
For each procedure and dataset, we subtract the mean PSNR, SSIM, or LPIPS of
the PSNR-selected schedule's outputs from the corresponding mean for the
LPIPS-selected schedule's outputs.
Positive PSNR and SSIM differences and negative LPIPS differences favor the
LPIPS-selected schedule.  Both means use the same 600 DrawBench, 1,659
GenEval-style, 4,896 PartiPrompts, or 30,000 DiffusionDB-clean10k prompt--seed
runs, with three seeds per prompt.}
\label{tab:search-objectives}
\centering
\footnotesize
\setlength{\tabcolsep}{3pt}
\begin{tabular}{llrrr}
\toprule
Model & Cache ratio & \multicolumn{3}{c}{LPIPS-selected $-$ PSNR-selected} \\
\cmidrule(lr){3-5}
 & & PSNR range (dB) & SSIM range & LPIPS range \\
\midrule
FLUX.1-dev & 0.58 & $-0.002$ to $+0.269$ & $+0.0023$ to $+0.0144$ & $-0.0224$ to $-0.0043$ \\
FLUX.1-dev & 0.74 & $-0.451$ to $+0.080$ & $+0.0009$ to $+0.0241$ & $-0.0493$ to $-0.0025$ \\
FLUX.1-dev & 0.82 & $-0.630$ to $-0.004$ & $+0.0010$ to $+0.0410$ & $-0.0992$ to $-0.0044$ \\
Qwen-Image & 0.58 & $-0.883$ to $+0.329$ & $+0.0007$ to $+0.0210$ & $-0.0261$ to $-0.0074$ \\
Qwen-Image & 0.74 & $-1.262$ to $-0.365$ & $+0.0059$ to $+0.0206$ & $-0.0454$ to $-0.0124$ \\
Qwen-Image & 0.82 & $-0.365$ to $+0.055$ & $+0.0068$ to $+0.0115$ & $-0.0199$ to $-0.0036$ \\
\bottomrule
\end{tabular}
\end{table}

%% file: appendix.tex
\section*{Appendix}

\begin{center}
\small
\begin{tabularx}{\linewidth}{@{}l X r@{}}
\toprule
Appendix & Contents & Page \\
\midrule
\ref{app:cache-experiment} & Experimental setup and baseline results & \pageref{app:cache-experiment} \\
\ref{app:shared-evidence} & Schedule frequencies, reuse, coverage, and exhaustive search & \pageref{app:shared-evidence} \\
\ref{app:trajectory-details} & Additional full-compute trajectory results & \pageref{app:trajectory-details} \\
\ref{app:mechanism} & Cache-error analysis and approximation policy comparisons & \pageref{app:mechanism} \\
\ref{app:search} & Schedule search, quality objectives, and additional results & \pageref{app:search} \\
\bottomrule
\end{tabularx}
\end{center}

\section{Experimental Setup and Baseline Results}
\label{app:cache-experiment}

\subsection{Sampling conditions}
\label{app:sampling-conditions}

The FLUX.1-dev trajectories use 50 FlowMatchEuler steps, BF16 inference,
$1024\times1024$ images, and guidance scale 3.5.  Their three base seeds are
41, 42, and 43.  The Qwen-Image trajectories use 50 FlowMatchEuler steps
with the checkpoint's dynamic noise-level shift, BF16 inference,
the model's native $1328\times1328$ image grid, classifier-free guidance 4.0,
and base seeds
42, 100042, and 200042.  The seed for prompt index $i$ is the base seed plus $i$.
Each prompt is therefore evaluated with three different seeds.

The four datasets contain 200 DrawBench prompts, 553 GenEval-style
prompts, 1,632 PartiPrompts, and 10,000 DiffusionDB prompts.
GenEval-style prompts are built from GenEval templates~\citep{ghosh2023geneval}.
DiffusionDB-clean10k is a 10,000-prompt subset of DiffusionDB~\citep{wang2022diffusiondb}.
Distances from the
line joining trajectory endpoints and principal-component variance summaries
use all prompts.  Turning profiles and comparisons
of plane orientations use the first 120 prompts from each dataset and seed,
giving 1,440 trajectories per image model.

Both video models generate 65 frames with 50 denoising steps in BF16.
HunyuanVideo uses the HYVideo-T/2-cfgdistill checkpoint at $640\times480$,
Euler sampling with flow shift 7, embedded guidance 6, and CFG scale 1.
Wan2.1 uses T2V-1.3B at $832\times480$, UniPC sampling with shift 5 and
guidance 5.  Each video model uses all 599 Penguin599 and 944 VBench944 prompts
under three base seeds, giving 4,629 full-compute trajectories.
The base seeds are 54, 55, and 56 for Penguin599 and 42, 43, and 44 for
VBench944.  Both models use the base seed plus the prompt index as the
generation seed.

\subsection{Methods}

Table~\ref{tab:method-families} compares how the methods choose cached steps
and how they approximate the skipped computation.
Adaptive methods choose a schedule during each prompt--seed run.
Offline searches select a fixed schedule before generation.

\begin{table}[!htbp]
\caption{\textbf{The methods differ in both schedule selection and approximation policy.}
The last row describes the search experiments in Section~\ref{sec:search}.}
\label{tab:method-families}
\centering
\footnotesize
\setlength{\tabcolsep}{3pt}
\begin{tabularx}{\linewidth}{@{}>{\raggedright\arraybackslash}p{0.23\linewidth}
>{\raggedright\arraybackslash}p{0.18\linewidth}
>{\raggedright\arraybackslash}p{0.25\linewidth}
>{\raggedright\arraybackslash}X@{}}
\toprule
Method & Schedule choice & Selection rule or objective & Approximation policy \\
\midrule
SeaCache~\citep{seacache2026} & Adaptive & Spectral feature change & Residual reuse \\
TeaCache~\citep{teacache2025} & Adaptive & Calibrated feature change & Residual reuse \\
SenCache~\citep{sencache2026} & Adaptive & Latent and timestep sensitivity & Residual reuse \\
DiCache~\citep{dicache2025} & Adaptive & First-block output change & Two-anchor prediction \\
TaylorSeer~\citep{taylorseer2025} & Fixed & Uniform spacing & First-order Taylor prediction \\
HiCache~\citep{hicache2026} & Fixed & Uniform spacing & Second-order Hermite prediction \\
L2P~\citep{l2p2026} & Fixed & Uniform spacing & Learned causal linear prediction \\
DPCache~\citep{dpcache2026} & Offline dynamic programming & Feature prediction error & Second-order feature prediction \\
BudCache~\citep{budcache2026} & Offline annealing & Final latent MSE & Residual reuse \\
MeanCache~\citep{meancache2026} & Offline graph search & Interval-average velocity error & Interval-average velocity prediction \\
\midrule
Our search experiments & Offline search & Final-output PSNR or LPIPS & Residual reuse \\
\bottomrule
\end{tabularx}
\end{table}

\subsection{Aggregation and evaluation measures}

For every model, method, cache ratio, dataset, and base seed,
the accelerated output is compared with a full-compute output generated from
the same prompt and seed.  PSNR measures pixel error on a logarithmic scale.
Higher values mean closer reconstruction.  SSIM compares local luminance,
contrast, and structure. Higher is better~\citep{wang2004ssim}.  LPIPS compares
deep visual features. Lower is better~\citep{zhang2018lpips}.  For video,
$\Delta_t$ compares the two videos' changes between consecutive frames.
For each video, we average LPIPS over all 64 adjacent frame pairs in its
65-frame sequence.  We subtract the full-compute video's average from the
accelerated video's average.  A value closer to zero is better.
CLIP Score measures text--image alignment and
ImageReward is a learned preference score~\citep{xu2023imagereward}.
CLIP Score uses image and text representations from CLIP~\citep{radford2021clip}.
VBench evaluates semantic and temporal video quality.

Tables~\ref{tab:flux-full-results}--\ref{tab:wan21-full-results} report each
dataset separately.  Every result is first averaged over prompts within a
base seed and then over the three base seeds.  Bold marks the best unrounded
mean within each dataset, cache ratio, and metric.
For image experiments that reuse an adaptive method's most frequent schedule, we compute
confidence intervals for the mean PSNR difference by resampling prompts.
We subtract the adaptive output's PSNR from the PSNR of the output generated
with the shared schedule on each run where the schedules differ, then average these differences
over the retained seeds for each prompt.

\subsection{Calibration, schedule search, and evaluation separation}
\label{app:calibration-protocol}

\paragraph{Preparing predictors and fixed schedules.}
Image methods use a separate set of 50 prompts whose normalized texts do
not overlap with the four image evaluation datasets.
Video methods use 50 motion prompts that are separate from Penguin599 and
VBench944.
Each fitted predictor or calibration table is shared across datasets for
the same model and method.
DPCache and MeanCache construct one schedule per model and cache ratio
on the preparation set and reuse it across evaluation datasets.
BudCache follows the same model-level protocol, using two preparation
prompts for FLUX and each video model and three for Qwen-Image.
Each BudCache search runs one restart, 200 simulated-annealing proposals, and at
most 20 local hill-climbing rounds.
BudCache ranks candidate schedules by final latent MSE.

\paragraph{Calibrating the cache ratio.}
Adaptive image methods initially use the same 50 preparation prompts.
For DiffusionDB, we calibrate the cache ratio using a separate set of 512
prompts whose normalized texts do not appear in the evaluation datasets.
If the realized cache ratio drifts, we adjust the threshold once using
cache counts from the evaluation data.
For video, we use 48 prompts from each evaluation corpus.
These prompts remain in the full 599-prompt and 944-prompt evaluations.
Ratio calibration uses cache decisions and realized cache counts.
The resulting threshold is fixed for each model, method, target ratio,
and prompt distribution across its prompt--seed runs.

\paragraph{Selecting shared schedules.}
For SeaCache, TeaCache, SenCache, and DiCache, we shuffle the 1,632
PartiPrompts indices with seed 42 and split them into three equal subsets.
We use the first subset of 544 prompts to select the most frequent schedule with
exactly $K$ cached steps for each model, method, and target ratio.
We combine the schedules from all three base seeds, counting each
prompt--seed run once.
We evaluate its reuse on the other 1,088 prompts.
The full 1,632-prompt result includes both selection and evaluation prompts.
These schedules also enter the comparisons of schedules and approximation
policies and provide initial candidates for the search in
Section~\ref{sec:search}.

\subsection{Complete per-model results}
\label{app:complete-tables}

The following tables present each model separately, with results by method,
dataset, target cache ratio, and quality metric.
Dataset abbreviations are DB for DrawBench, PP for PartiPrompts, GE for
GenEval-style, DDB for DiffusionDB-clean10k, P599 for Penguin599, and VB944
for VBench944.  Method labels shorten SeaCache, TeaCache,
SenCache, DiCache, TaylorSeer, HiCache, DPCache, BudCache, and MeanCache to Sea,
Tea, Sen, Di, Tay, Hi, DP, Bud, and Mean. O1 and O2 are prediction orders.  P,
S, L, C, and IR denote PSNR, SSIM, LPIPS, CLIP, and ImageReward.  The video
tables add temporal-LPIPS change $\Delta_t$ and VBench score VB.

\input{\arxivroot tables/full_results_flux}
\input{\arxivroot tables/full_results_qwen}
\input{\arxivroot tables/full_results_hunyuan_video}
\input{\arxivroot tables/full_results_wan21}

\subsection{Cache-ratio alignment}
\label{app:ratio-alignment}

The six fixed-schedule image methods execute exactly $K$ cached steps.  The four
adaptive methods choose full and cached steps for each prompt--seed run and are calibrated
toward the same target mean.  Tables~\ref{tab:actual-k-flux} and
\ref{tab:actual-k-qwen} report their realized mean cache counts separately for
every dataset represented in the complete quality tables.

\begin{table}[!htbp]
\caption{FLUX.1-dev mean realized cache counts for the four adaptive methods.}
\label{tab:actual-k-flux}
\centering
\small
\begin{tabular}{llrrr}
\toprule
Dataset & Method & $K_{\rm tgt}=29$ & $K_{\rm tgt}=37$ & $K_{\rm tgt}=41$ \\
\midrule
DrawBench & SeaCache & 28.998 & 37.000 & 41.000 \\
DrawBench & TeaCache & 29.013 & 36.917 & 40.993 \\
DrawBench & SenCache & 28.718 & 37.000 & 40.793 \\
DrawBench & DiCache & 28.668 & 37.000 & 41.000 \\
\midrule
Parti & SeaCache & 28.991 & 37.000 & 41.000 \\
Parti & TeaCache & 29.021 & 36.864 & 40.989 \\
Parti & SenCache & 28.757 & 36.999 & 40.825 \\
Parti & DiCache & 28.716 & 37.000 & 41.000 \\
\midrule
GenEval & SeaCache & 28.997 & 37.000 & 41.000 \\
GenEval & TeaCache & 29.010 & 36.903 & 40.987 \\
GenEval & SenCache & 28.457 & 37.000 & 40.731 \\
GenEval & DiCache & 28.558 & 36.999 & 41.000 \\
\midrule
DiffusionDB & SeaCache & 28.991 & 37.000 & 41.000 \\
DiffusionDB & TeaCache & 29.023 & 36.854 & 40.994 \\
DiffusionDB & SenCache & 29.093 & 37.000 & 40.912 \\
DiffusionDB & DiCache & 28.823 & 37.000 & 41.000 \\
\bottomrule
\end{tabular}
\end{table}

\begin{table}[!htbp]
\caption{Qwen-Image mean realized cache counts for the four adaptive methods.}
\label{tab:actual-k-qwen}
\centering
\small
\begin{tabular}{llrrr}
\toprule
Dataset & Method & $K_{\rm tgt}=29$ & $K_{\rm tgt}=37$ & $K_{\rm tgt}=41$ \\
\midrule
DrawBench & SeaCache & 29.003 & 37.003 & 41.000 \\
DrawBench & TeaCache & 29.000 & 37.195 & 40.875 \\
DrawBench & SenCache & 28.915 & 37.000 & 41.000 \\
DrawBench & DiCache & 28.932 & 37.000 & 41.000 \\
\midrule
Parti & SeaCache & 29.002 & 37.002 & 41.000 \\
Parti & TeaCache & 28.999 & 37.175 & 40.894 \\
Parti & SenCache & 28.925 & 37.000 & 41.000 \\
Parti & DiCache & 28.944 & 37.000 & 41.000 \\
\midrule
GenEval & SeaCache & 29.000 & 36.998 & 41.000 \\
GenEval & TeaCache & 29.000 & 37.051 & 40.827 \\
GenEval & SenCache & 28.993 & 37.000 & 41.000 \\
GenEval & DiCache & 28.879 & 37.000 & 41.000 \\
\midrule
DiffusionDB & SeaCache & 29.001 & 37.001 & 41.000 \\
DiffusionDB & TeaCache & 29.000 & 37.238 & 41.000 \\
DiffusionDB & SenCache & 28.964 & 37.000 & 41.000 \\
DiffusionDB & DiCache & 28.995 & 37.000 & 41.000 \\
\bottomrule
\end{tabular}
\end{table}

\input{\arxivroot tables/actual_k_video}

The realized means remain close to their targets, but adaptive methods do not
always execute exactly the target cache ratio.  The largest image deviation is
0.543 cached steps for FLUX SenCache on GenEval-style at target 29.  Tables
\ref{tab:actual-k-flux}--\ref{tab:actual-k-wan21} give the realized cache counts
for each quality comparison.

\subsection{Online Runtime and Offline Preparation Cost}
\label{app:compute-costs}

The online and offline costs answer different questions.  Online latency is
paid for every generated sample.  Calibration, predictor fitting, and schedule
search are one-time costs that can be amortized over later generations.  We
therefore report them separately.

The image benchmark uses one H100 per image and a batch size of one.
Each model--method--ratio configuration uses the same 16 prompts at base seed
314159.  The first four prompt indices are excluded as warm-up, and latency
is the arithmetic mean over the other 12 prompt--seed runs.  These are 12
different prompts.  Timing includes prompt encoding, denoising, VAE decoding,
and conversion to an image.  It excludes model loading and image-file saving.
GPU work is synchronized before and after the timed computation.
Speedup divides the full-compute mean latency by the accelerated mean.

Video latency is measured during generation on H100 GPUs at batch size one.
Each configuration generates one video for each prompt--seed run, giving
599 videos per Penguin599 base seed and 944 per VBench944 base seed.
Timing includes prompt encoding, denoising, decoding, and output-file saving.
It excludes model loading.  The first video of each generation worker is
excluded as warm-up.  We average the retained per-video times within each
dataset and base seed, then give the six dataset--seed means equal weight.
The image and video timings therefore use different sample populations and
timing boundaries.
Tables~\ref{tab:latency-flux}--\ref{tab:latency-wan21} give the four online
latency tables.

\input{\arxivroot tables/compute_costs}

Offline preparation is paid once per model.  SeaCache, DiCache, TaylorSeer, and
HiCache do not fit model-specific predictors. TeaCache uses published coefficients
on FLUX.1-dev and requires no additional coefficient fitting there.
BudCache runs the largest search.
It evaluates 179 to 727 candidate schedules per cache count on the image
models, which takes 0.19 to 1.65 H100 hours on FLUX.1-dev and 0.91 to 5.82
hours on Qwen-Image.  On the video models it evaluates 162 to 446 schedules,
which takes 4.8 to 18.3 RTX 6000 Pro GPU hours on HunyuanVideo and 1.4 to 8.2
H100 hours on Wan2.1.  The other fitted methods are cheaper.  SenCache, L2P,
and MeanCache take 2.5 to 7.2 RTX 6000 Pro GPU hours on HunyuanVideo and 1.8
to 2.9 H100 hours on Wan2.1. Image calibration for these methods and DPCache
uses 50 prompts. Comparable preparation times were recorded
only for the video implementations.
Calibrating adaptive methods to the target cache ratios is the largest
offline cost for the evaluated video methods: 67.5 H100 hours for HunyuanVideo and
68.0 hours for Wan2.1.  RTX 6000 Pro hours are not converted to H100 hours.

The exhaustive schedule search of Appendix~\ref{app:k41-search} is far larger
than any of these.  It generated 5,483,016 cached images.  At the 2.37 seconds per image measured
for a fixed schedule with residual reuse at $K=41$ on FLUX.1-dev in
Table~\ref{tab:latency-flux}, that is about 3,600 H100 hours.

\section{Additional Evidence for Shared Schedules}
\label{app:shared-evidence}

\subsection{Schedule-distribution statistics}
\label{app:schedule-distribution}

The image study comprises 891,720 prompt--seed runs across 96 combinations
of method, model, dataset, and cache ratio.
The number of unique schedules ranges from 2 to 62 per combination,
with a median of 11 across the 96 combinations.
Across these 96 combinations, the fraction of runs using one of the three
most frequent schedules has a median of 89.8\% and a range of 40.7--100\%.
The same schedule is most frequent under all three seeds in 85 of the
96 image combinations.

We compute schedule frequencies from the exact sequences of 50 full or
cached decisions, combining the three base seeds within each model, dataset,
cache ratio, and method. If schedule $u$ occurs on a fraction $p(u)$ of runs, the
effective schedule count is $\exp[-\sum_u p(u)\log p(u)]$.
It is the number of equally frequent schedules with the same entropy.
For example, if all observed schedules are equally frequent, their effective
count equals their unique count.

\begin{table}[!htbp]
\caption{Schedules chosen by the four adaptive image methods.
Within each model, dataset, and target cache ratio, we combine the three seeds
and compute schedule frequencies.
Top-$r$ is the fraction of runs using one of the $r$ most frequent schedules
within that combination.
The table averages these fractions equally over 24 combinations, each with
600 to 30,000 prompt--seed runs.
Unique and effective counts are averaged over the same combinations, with
their minimum and maximum in brackets. Runs gives the total across combinations.}
\label{tab:gate-paths}
\centering
\small
\setlength{\tabcolsep}{4pt}
\begin{tabular}{lrrrrrr}
\toprule
Method & Runs & Unique & Effective & Top-1 & Top-3 & Top-10 \\
\midrule
SeaCache & 222,930 & 8.75 [2,18] & 2.23 [1.04,3.77] & 0.746 & 0.960 & 0.9996 \\
TeaCache & 222,930 & 18.25 [3,62] & 5.56 [1.60,13.75] & 0.456 & 0.778 & 0.9752 \\
SenCache & 222,930 & 10.46 [3,25] & 3.99 [1.23,8.64] & 0.561 & 0.857 & 0.9918 \\
DiCache & 222,930 & 16.46 [5,51] & 5.55 [2.28,12.89] & 0.426 & 0.777 & 0.9811 \\
\bottomrule
\end{tabular}
\end{table}

The most frequent schedule of an image method is identical across all four
image datasets in five of six model--ratio combinations for SeaCache.  The counts
are four of six for SenCache, two of six for TeaCache, and one of six for
DiCache.  On video, the same schedule is most frequent under all three
base seeds in 46 of 48 model--method--dataset--ratio combinations.  The two
disagreements are TeaCache on HunyuanVideo VBench944 at cache ratio 0.58 and
SenCache on Wan2.1 Penguin599 at cache ratio 0.74.
Across the two video models, the most frequent schedules differ at 4 to 23
step positions.  The two video datasets share the most frequent schedule
in 18 of the 24 model--method--ratio combinations.

The four datasets have different sample sizes, which affects their unique
schedule counts in Table~\ref{tab:gate-paths}. Top-$r$ frequencies and effective-count
summaries are more directly comparable across datasets.

Each method contributes 222,930 prompt--seed runs across 24 combinations.
Each model--dataset--ratio combination contains 600 runs for DrawBench,
1,659 for GenEval, 4,896 for PartiPrompts, or 30,000 for DiffusionDB.
These combinations receive equal weight in the summary, regardless of their size.

Schedule concentration varies across methods.  SeaCache has the smallest mean
effective count, 2.23, followed by SenCache at 3.99.  TeaCache and DiCache
average 5.56 and 5.55 effective schedules, respectively.  The broadest observed
combination has 13.75 effective schedules.  The schedule-reuse experiment in
Section~\ref{sec:fixed-replay} evaluates each method's most frequent schedule
on new prompts.

Fixed schedules can also differ between models.  Replacing HunyuanVideo with
Wan2.1 changes the BudCache schedule by 22, 12, and 6 steps at cache ratios
0.58, 0.74, and 0.82.  The corresponding MeanCache changes are 18, 8, and 0
steps.  MeanCache therefore uses the same schedule on both models at cache
ratio 0.82.

\subsection{Reusing adaptive schedules on images}
\label{app:replay-images}
We compare SeaCache, TeaCache, SenCache, and DiCache with fixed reuse of
their most frequent schedules.
Each model--method--ratio combination uses 1,632 prompts under three seeds,
giving 4,896 prompt--seed runs.
We select the most frequent schedule on 544 prompts and evaluate its reuse
on the remaining 1,088 prompts.
For each run, we subtract the adaptive output's PSNR from the PSNR of the
output generated with the fixed schedule and the same approximation policy.
Runs with identical schedules have zero difference, so the results below
use only runs where the two schedules differ.

We first report mean quality changes across the 24 model--method--ratio combinations.
Using all prompts gives 54 to 4,313 retained runs per combination.
The mean PSNR differences range from $-0.79$ to $+1.60$ dB, where positive
values favor the fixed schedule. The median absolute difference in mean PSNR is 0.09 dB.
Restricting evaluation to the 1,088 prompts excluded from selection gives
34 to 2,901 retained runs per combination.
On these prompts, the mean PSNR differences range from $-0.81$ to $+1.56$ dB,
with a median absolute difference in mean PSNR of 0.10 dB.

We separately measure the size of the change within individual runs.
For each combination, we take the absolute PSNR difference on each retained
run from the 1,088 evaluation prompts and report the median.
These 24 medians range from 0.00 to 1.53 dB, with a median of 0.08 dB.
Tables~\ref{tab:fixed-replay-image-full} and \ref{tab:fixed-replay-image-seeds}
report the combined-seed and per-seed results.

\input{\arxivroot tables/fixed_replay_image}
\input{\arxivroot tables/fixed_replay_image_seeds}

Table~\ref{tab:fixed-replay-image-full} gives 95\% confidence intervals for
the mean PSNR differences on prompts excluded from selection.
We first average each prompt's PSNR differences over the seeds where its
adaptive and fixed schedules differ,
then resample prompts 20,000 times.
Each resampled mean weights a prompt by its number of retained runs,
matching the reported mean paired difference.
The interval spans the 2.5th to 97.5th percentiles of these resampled means.
Twenty-two of the 24 intervals exclude zero.  The two that do not
are Qwen-Image DiCache at $K=29$ and Qwen-Image SeaCache at $K=37$, the
latter containing 34 runs.

The lower block of Table~\ref{tab:fixed-replay-image-full} replaces the fixed
schedule with a random schedule of the same $K$, on the same held-out runs
where the fixed and adaptive schedules differ. We draw each random schedule's
$K$ cached steps uniformly without replacement from steps 3--48, leaving
steps 0, 1, 2, and 49 full.
Each random schedule uses the approximation policy of the method it is compared against:
residual reuse for SeaCache, SenCache, and TeaCache, and the two-anchor
extrapolation for DiCache. There are five draws per model at $K=29$ and
$K=37$ and two at $K=41$.
The first two draws at each cache count use three seeds. The remaining
draws at $K=29$ and $K=37$ use only base seed 42.
For each paired run, we subtract the adaptive output's PSNR from the random
schedule output's PSNR.
We report the median of these differences across draws, models, cache counts,
available seeds, and evaluation prompts where the adaptive and fixed schedules differ.
A prompt can contribute to several draws and seeds.
The median paired PSNR difference is negative for SeaCache, SenCache, and DiCache.
For TeaCache, the random schedules have higher PSNR but lower SSIM and
higher LPIPS than the adaptive schedules.

\subsection{Reusing adaptive schedules on videos}
\label{app:replay-videos}
The video comparison covers the same four methods. Each comparison contains
150 prompts from each of Penguin599 and VBench944, with one seed per prompt.
Both models use the first 150 prompts in each dataset's evaluation list,
ordered by the SHA-256 hash of the prompt text.
For Penguin599, we apply Unicode NFKC normalization, lowercase the text,
remove leading and trailing whitespace, and replace internal runs of
whitespace with a single space before hashing.
For VBench944, we hash the original text. Both use UTF-8 encoding.
Generation uses the original prompt text.

For each model, method, and target cache ratio, we combine schedule counts
from both datasets and all three base seeds, counting each prompt--seed run once.
We choose the most frequent schedule with exactly $K$ cached steps.
If no schedule has that cache count, we choose the most frequent schedule
across all cache counts.
Restricting selection to prompts outside the 300 evaluation prompts,
excluding those prompts under all three seeds, gives the same schedule
in all 24 model--method--ratio combinations.
Of the 24 model--method--ratio combinations, three have a fixed schedule
whose cache count differs from the target. In two other combinations,
the adaptive method already follows the fixed schedule on every evaluation prompt.
Table~\ref{tab:fixed-replay-video-full} separates these groups.
We compare mean PSNR on prompts with different schedules in the remaining
19 combinations. The fixed schedule's mean PSNR is at most 0.25 dB lower
than the adaptive method's in 17 of them.
The two exceptions are SenCache on HunyuanVideo at ratio 0.58 and SenCache on
HunyuanVideo at ratio 0.82.
Table~\ref{tab:fixed-replay-video-datasets} gives the Penguin599 and VBench944
results separately.

The video random control uses uniformly sampled schedules with the same cache
ratio and residual-reuse approximation policy.
For each model and cache ratio, we draw two schedules by sampling the $K$
cached steps uniformly without replacement from steps 3--48.
Steps 0, 1, 2, and 49 remain full.
We compare them with SeaCache,
TeaCache, and SenCache on both video models. For each paired run, we subtract
the adaptive output's PSNR from the random-schedule output's PSNR.
For each model and method, we take the median of these differences across
random draws, cache ratios, dataset-specific base seeds, and prompts.
A prompt can contribute to several draws and cache ratios.
The six medians range from $-5.9$ to $-0.3$ dB, favoring the adaptive methods.
DiCache is excluded because its adaptive run uses two-anchor prediction
rather than residual reuse.

\input{\arxivroot tables/fixed_replay_video}
\input{\arxivroot tables/fixed_replay_video_datasets}

\subsection{The coverage test}
\label{app:coverage}

\subsubsection{Candidate sets and schedule selection}
\label{app:coverage-candidates}

A prompt's best evaluated result depends on the schedules in its candidate set.
We compare two candidate sets to examine how this choice affects coverage.
The smaller sets contain 9--17 schedules from offline search methods, uniform
spacing, frequent adaptive choices, and variants of the MeanCache schedule.
The enlarged sets contain every available distinct schedule at the same
cache ratio under residual reuse, including further MeanCache variants and
random schedules. They contain 13--32 schedules.
Section~\ref{sec:shared} summarizes the coverage results for these enlarged sets.
Within each set, we compare a fixed schedule's quality with the highest
quality reached by any candidate on the same prompt.

For the image coverage tests, we construct variants of the MeanCache schedule
by exchanging one, two, or
four cached steps with full steps while preserving the cache count.
For each available swap count, the first variant restricts every exchanged
position to steps after the first cached step and is evaluated under three seeds.
Further variants at cache ratios 0.58 and 0.74 use base seed 42, with swaps
either restricted in this way or allowed throughout steps 3--48.
The coverage tests include the variants with available quality measurements.

Formally, let $q(u;x)$ be the seed-averaged quality of schedule $u$ on prompt
$x$.  The comparison value is $b(x)=\max_{u\in\mathcal U}q(u;x)$ for
candidate set $\mathcal U$.
A binary schedule has $u_n=1$ at a cached step.
The fixed-budget space is
$\mathcal S_K=\{u\in\{0,1\}^{N}:\norm{u}_0=K\}$.
Each experiment specifies any required full steps within this space.

For each image prompt and schedule, we average scores over the available base seeds.  Most
configurations use three base seeds. Additional MeanCache variants and random
draws use only base seed 42, as described above and in
Appendix~\ref{app:replay-images}. We also
evaluate all image schedules at their common base seed 42 below.  Video
scores use one seed per prompt throughout each candidate set.

For each candidate set, we select the schedule with the highest coverage
at 0.25 dB and report that same schedule at all three margins.
We repeat this selection for the enlarged set.
To account for selecting among $m$ schedules, we compute a one-sided
binomial lower confidence bound at the Bonferroni level $0.05/m$.
At each margin, we count the model--ratio combinations whose corrected
lower bound on coverage exceeds 0.5.

\subsubsection{PSNR coverage}
\label{app:coverage-psnr}

For the smaller candidate sets, Table~\ref{tab:oracle-coverage} reports
12 model--ratio combinations, six image and six video.
All twelve corrected bounds exceed 0.5 at the 0.5 dB margin, the lowest being
0.529.  Eight of the twelve pass at the 0.25 dB margin under the same
correction.  Random schedules of the same $K$ reach at most 6\% of prompts at
0.25 dB.

The enlarged candidate sets provide a stronger per-prompt reference.
Under the same correction, the lower bound exceeds 0.5 in 7 of 12
combinations at 0.25 dB, 9 of 12 at 0.5 dB, and all 12 at 1.0 dB.
These are the enlarged-set results summarized in Section~\ref{sec:shared}.
The separate 337-schedule comparison in Appendix~\ref{app:k41-search} examines
FLUX.1-dev at cache ratio 0.82 on 2,381 new prompts.
One schedule stays within 0.5 dB of each prompt's best candidate on 62\% of
these prompts. The schedule selected from four scoring runs reaches 57\%.

To compare all image schedules under the same seed, we repeat the enlarged-set
analysis at base seed 42 in Table~\ref{tab:coverage-common-seed}.
We again select one schedule at the 0.25 dB margin before reporting all three margins.
Each of the six reported schedules stays within 0.5 dB of its
candidate set's per-prompt best on a majority of prompts, including after correcting
the lower bounds for candidate set size.

\input{\arxivroot tables/coverage_common_seed}

\begin{table}[!htbp]
\caption{One shared schedule in each smaller candidate set stays within 0.5 dB
of the per-prompt best on most prompts.
These sets contain 9--17 schedules.
For every model and cache ratio, the per-prompt best is the highest
seed-averaged PSNR reached by any schedule in the candidate set, all of which use
residual reuse.  Each entry gives the share of prompts inside the margin, with
the unadjusted one-sided 95\% binomial lower bound in brackets.
The reported counts of combinations whose lower bounds exceed 0.5 use
Bonferroni-adjusted bounds. We report the candidate set
member with the largest share at 0.25 dB. The reported MeanCache variant
exchanges one cached step with one full step.
Image rows use the 1,088 PartiPrompts prompts excluded from the 544-prompt
recurring-schedule selection split.
Video rows use 300 prompts split between Penguin599 and VBench944.}
\label{tab:oracle-coverage}
\centering
\small
\setlength{\tabcolsep}{4pt}
\begin{tabular}{llrlrrr}
\toprule
Model & $K$ & Candidates & Fixed schedule & 0.25 dB & 0.5 dB & 1.0 dB \\
\midrule
FLUX.1-dev & 29 & 17 & MeanCache & 0.694 [0.670] & 0.776 [0.754] & 0.869 [0.851] \\
FLUX.1-dev & 37 & 17 & MeanCache variant & 0.443 [0.418] & 0.571 [0.546] & 0.715 [0.692] \\
FLUX.1-dev & 41 & 10 & BudCache & 0.653 [0.629] & 0.752 [0.729] & 0.882 [0.865] \\
Qwen-Image & 29 & 17 & MeanCache & 0.482 [0.456] & 0.583 [0.558] & 0.767 [0.744] \\
Qwen-Image & 37 & 17 & MeanCache & 0.530 [0.505] & 0.617 [0.592] & 0.763 [0.741] \\
Qwen-Image & 41 & 10 & MeanCache & 0.756 [0.734] & 0.849 [0.830] & 0.949 [0.937] \\
\midrule
HunyuanVideo & 29 & 11 & MeanCache & 0.530 [0.481] & 0.637 [0.588] & 0.817 [0.776] \\
HunyuanVideo & 37 & 10 & MeanCache & 0.620 [0.571] & 0.703 [0.657] & 0.807 [0.765] \\
HunyuanVideo & 41 & 10 & BudCache & 0.630 [0.582] & 0.713 [0.667] & 0.860 [0.823] \\
Wan2.1 & 29 & 11 & MeanCache & 0.697 [0.650] & 0.760 [0.716] & 0.840 [0.801] \\
Wan2.1 & 37 & 10 & MeanCache & 0.637 [0.588] & 0.740 [0.695] & 0.907 [0.874] \\
Wan2.1 & 41 & 9 & BudCache & 0.673 [0.626] & 0.827 [0.787] & 0.943 [0.916] \\
\bottomrule
\end{tabular}
\end{table}

\subsubsection{Quality margins and coverage under SSIM and LPIPS}
\label{app:coverage-margins}

To interpret the PSNR margins in Table~\ref{tab:oracle-coverage}, we compare
them with the quality change across cache ratios for the schedules from
BudCache, DPCache, MeanCache, and uniform spacing.
For each schedule, image model, and ratio, we average
PSNR over the available seeds within each prompt and take the median across
the 1,088 held-out PartiPrompts prompts.
We subtract the higher-ratio median from the lower-ratio median.
Adjacent ratios differ by 8 and by 4 cached steps, so we divide
each drop by the number of additional cached steps. This gives the average quality
change per additional cached step between the tested ratios.
There are 16 values, four schedules by two ratio pairs by two models.
They run from 0.171 to 1.166 dB, with a median of 0.592 dB.  The margins are
0.25, 0.5, and 1.0 dB.  These are 0.42, 0.84, and 1.69 times that median.

The coverage test in Table~\ref{tab:oracle-coverage} uses PSNR to measure quality.
We repeat the test with SSIM and LPIPS, using the same candidate sets and prompts.
Each measure needs its own margins.  On the same four schedules, we take the
median of seed-averaged prompt quality over the same 1,088 held-out prompts
at each cache ratio.  We divide the SSIM decrease or LPIPS increase between
adjacent tested ratios by the number of additional cached steps.
There are again 16
values.  For SSIM they run from 0.0054 to 0.0367, with a median of 0.0138.
For LPIPS they run from 0.0080 to 0.0495, with a median of 0.0202.
For each metric, the three margins are about one half, one, and two times
its median change per additional cached step.
LPIPS is smaller when quality is better, so we negate it before
the test and state its margins as positive LPIPS differences.
Table~\ref{tab:oracle-coverage-metrics} gives the result.

\begin{table}[!htbp]
\caption{Shared schedules also approach prompt-specific choices under SSIM and LPIPS.
Candidate sets and evaluation prompts are those of Table~\ref{tab:oracle-coverage}.
For each metric, we select the schedule with the highest coverage at the
tightest margin and report it at all three margins.
Cells show coverage and the unadjusted one-sided 95\% lower bound in brackets.
Each prompt's reference is its best seed-averaged score among the candidates.
All schedules use residual reuse.
The MeanCache variant exchanges one cached step with one full step.
LPIPS is negated for the test, while margins are reported as positive LPIPS differences.}
\label{tab:oracle-coverage-metrics}
\centering
\small
\setlength{\tabcolsep}{4pt}
\begin{tabular}{llrlrrr}
\toprule
Model & $K$ & Candidates & Fixed schedule & Tight & Middle & Wide \\
\midrule
\multicolumn{7}{@{}l}{SSIM, margins 0.0075, 0.015, and 0.03} \\
FLUX.1-dev & 29 & 17 & MeanCache & 0.825 [0.805] & 0.917 [0.902] & 0.968 [0.958] \\
FLUX.1-dev & 37 & 17 & MeanCache variant & 0.484 [0.459] & 0.602 [0.577] & 0.773 [0.751] \\
FLUX.1-dev & 41 & 10 & SeaCache most frequent & 0.608 [0.583] & 0.723 [0.700] & 0.866 [0.848] \\
Qwen-Image & 29 & 17 & MeanCache & 0.619 [0.595] & 0.835 [0.816] & 0.964 [0.953] \\
Qwen-Image & 37 & 17 & MeanCache & 0.481 [0.455] & 0.626 [0.601] & 0.821 [0.801] \\
Qwen-Image & 41 & 10 & BudCache & 0.551 [0.526] & 0.689 [0.665] & 0.866 [0.848] \\
HunyuanVideo & 29 & 11 & MeanCache & 0.730 [0.685] & 0.847 [0.808] & 0.940 [0.912] \\
HunyuanVideo & 37 & 10 & MeanCache & 0.523 [0.474] & 0.633 [0.585] & 0.750 [0.705] \\
HunyuanVideo & 41 & 10 & BudCache & 0.503 [0.454] & 0.597 [0.548] & 0.723 [0.678] \\
Wan2.1 & 29 & 11 & MeanCache & 0.567 [0.518] & 0.680 [0.633] & 0.823 [0.783] \\
Wan2.1 & 37 & 10 & BudCache & 0.613 [0.565] & 0.690 [0.643] & 0.827 [0.787] \\
Wan2.1 & 41 & 9 & BudCache & 0.760 [0.716] & 0.863 [0.826] & 0.950 [0.924] \\
\midrule
\multicolumn{7}{@{}l}{LPIPS, margins 0.01, 0.02, and 0.04} \\
FLUX.1-dev & 29 & 17 & MeanCache & 0.851 [0.832] & 0.936 [0.922] & 0.983 [0.976] \\
FLUX.1-dev & 37 & 17 & MeanCache variant & 0.480 [0.454] & 0.623 [0.598] & 0.791 [0.770] \\
FLUX.1-dev & 41 & 10 & SeaCache most frequent & 0.660 [0.636] & 0.761 [0.739] & 0.870 [0.852] \\
Qwen-Image & 29 & 17 & BudCache & 0.649 [0.624] & 0.846 [0.826] & 0.955 [0.943] \\
Qwen-Image & 37 & 17 & MeanCache & 0.463 [0.438] & 0.592 [0.567] & 0.798 [0.777] \\
Qwen-Image & 41 & 10 & BudCache & 0.545 [0.520] & 0.675 [0.650] & 0.853 [0.834] \\
HunyuanVideo & 29 & 11 & MeanCache & 0.727 [0.681] & 0.880 [0.845] & 0.967 [0.944] \\
HunyuanVideo & 37 & 10 & SeaCache most frequent & 0.493 [0.444] & 0.553 [0.504] & 0.647 [0.599] \\
HunyuanVideo & 41 & 10 & SenCache most frequent & 0.487 [0.438] & 0.557 [0.508] & 0.690 [0.643] \\
Wan2.1 & 29 & 11 & MeanCache & 0.690 [0.643] & 0.840 [0.801] & 0.960 [0.936] \\
Wan2.1 & 37 & 10 & MeanCache & 0.617 [0.568] & 0.790 [0.748] & 0.947 [0.920] \\
Wan2.1 & 41 & 9 & MeanCache & 0.690 [0.643] & 0.857 [0.819] & 0.970 [0.948] \\
\bottomrule
\end{tabular}
\end{table}

After the same candidate-count correction, all 12 SSIM bounds exceed 0.5 at
the middle margin, the lowest being 0.521, and seven of the twelve do so at
the tightest margin.  Under LPIPS, 10 of the 12 corrected bounds exceed 0.5
at the middle margin, and eight do so at the tightest margin.
At the middle margin, HunyuanVideo has corrected bounds of
0.477 at $K=37$ and 0.481 at $K=41$.
At the tightest margins, the best random schedules have higher coverage
under SSIM and LPIPS than under PSNR.
The best random schedule under SSIM covers
0.118 of the image prompts and 0.117 of the video prompts.  Under LPIPS the two
shares are 0.132 and 0.087.  Under PSNR they are 0.057 and 0.020.

\subsection{Exhaustive high-ratio schedule search}
\label{app:k41-search}

\subsubsection{Search space and schedule selection}
\label{app:k41-selection}

This experiment uses FLUX.1-dev with 50 FlowMatchEuler steps, BF16 inference,
$1024\times1024$ resolution, guidance scale 3.5, and residual reuse at every
cached step.  Steps 0, 1, 2, and 49 always run the full transformer.  Five
additional full steps are chosen from steps 3 through 48, giving
$\binom{46}{5}=1{,}370{,}754$ schedules with 41 cached steps.

Every schedule is evaluated on four PartiPrompts calibration examples, each
containing one prompt and one seed.
We sort the indices of the 544-prompt selection subset described in
Appendix~\ref{app:calibration-protocol} in ascending order and take its first four prompts.
The generation seed for prompt index $i$ is $42+i$.
The four inputs are:
\begin{itemize}[nosep,leftmargin=*]
\item Index 5, seed 47: ``an illustration of a baby daikon radish in a tutu walking a dog''.
\item Index 8, seed 50: ``an avocado''.
\item Index 9, seed 51: ``a young badger delicately sniffing a yellow rose, richly textured oil painting''.
\item Index 15, seed 57: ``a store front that has the word `openai' written on it.''
\end{itemize}
This search produces 5,483,016 cached images.  Two schedules are selected from
these four PSNR values before the larger evaluation.  One maximizes their mean.
The other maximizes their minimum.  Table~\ref{tab:k41-selected-paths} gives
both schedules and two existing fixed-schedule controls.

\begin{figure}[!htbp]
  \centering
  \includegraphics[width=2.6in]{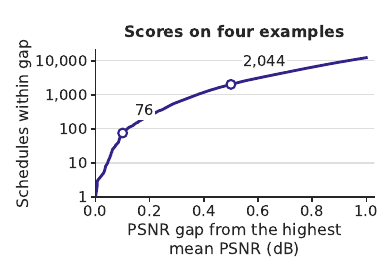}
  \caption{\textbf{On the four selection examples, 76 schedules are within
  0.1 dB of the highest mean PSNR.}
  The curve shows the cumulative number of schedules within each PSNR gap
  after evaluating all 1,370,754 schedules for FLUX.1-dev at cache ratio 0.82.
  The vertical axis is logarithmic.
  Circles mark the counts at gaps of 0.1 and 0.5 dB.}
  \label{fig:exhaustive-selection-counts}
\end{figure}

Figure~\ref{fig:exhaustive-selection-counts} shows how many schedules approach
the highest mean PSNR on the four selection examples.

\begin{table}[!htbp]
\caption{Schedules selected from four calibration examples.  The four boundary
steps are required for every schedule.  PSNR values are in dB.}
\label{tab:k41-selected-paths}
\centering
\small
\begin{tabularx}{\linewidth}{@{}lXrr@{}}
\toprule
Selection rule & Full-compute steps & Mean PSNR & Minimum PSNR \\
\midrule
Highest mean & 0, 1, 2, 4, 6, 11, 24, 41, 49 & 22.746 & 20.496 \\
Highest minimum & 0, 1, 2, 4, 6, 13, 23, 40, 49 & 22.672 & 20.739 \\
MeanCache schedule & 0, 1, 2, 3, 4, 9, 19, 34, 49 & 21.619 & 18.041 \\
BudCache schedule & 0, 1, 2, 4, 7, 13, 20, 39, 49 & 22.235 & 19.190 \\
\bottomrule
\end{tabularx}
\end{table}

The schedule selected by four-run mean PSNR is on average 0.412 dB below
the four runs' individual best schedules.
Table~\ref{tab:k41-per-run-optima} compares its PSNR with the highest PSNR
found for each run across all 1,370,754 schedules in the space defined above.

We also count schedules that are close to all four individual maxima.
For each schedule, we subtract its PSNR from the best PSNR for each run
and retain it only if all four gaps are at most 1.0 dB.
Exactly 42 schedules satisfy this condition in the same fully enumerated space.

\subsubsection{Evaluation data and fixed-schedule results}

For the larger evaluation, we select the top 64 schedules by four-run mean PSNR, the top 64 by
minimum PSNR, and the top 16 by mean PSNR for every subset of one, two,
or three scoring runs.
We also sample 128 distinct schedules uniformly from the full search space
using seed 20270826 and include 11 existing or constructed controls.
Combining these candidates and removing duplicates yields 337 distinct schedules.
The 11 controls include the BudCache, MeanCache, and DPCache schedules,
DiCache's most frequent schedule with 41 cached steps, and a schedule with
additional full steps at 3, 14, 25, 37, and 48.
The other six are two MeanCache variants formed by exchanging one or two
cached steps with full steps, two previously sampled random schedules,
and two schedules constructed from full-compute trajectory measurements.
The last two have full steps at 0, 1, 2, 4, 17, 31, 41, 46, 49 and at
0, 1, 2, 17, 28, 36, 42, 46, 49, respectively.

The two selected schedules, the MeanCache and BudCache schedules, and the adaptive methods
are evaluated on the data in Table~\ref{tab:k41-evaluation-scope}.
The quality and ranking summaries omit the four prompt--seed runs used
for selection, leaving 7,151 runs across the first three datasets.
All 337 candidate schedules are evaluated on these runs.
The two selected schedules and the method comparisons are also run on
30,000 DiffusionDB runs.
The coverage analysis in Figure~\ref{fig:exhaustive-existence} instead excludes
the four selection prompts under all three seeds.
It averages each remaining prompt's scores over the three seeds and uses
2,381 new prompts across the first three datasets.

\begin{table}[!htbp]
\caption{Evaluation data for the high-ratio search.  Each dataset uses three
base seeds.}
\label{tab:k41-evaluation-scope}
\centering
\small
\begin{tabular}{lrrl}
\toprule
Dataset & Prompts & Runs & Schedules evaluated \\
\midrule
DrawBench & 200 & 600 & 337 candidates and method comparisons \\
GenEval-style & 553 & 1,659 & 337 candidates and method comparisons \\
PartiPrompts & 1,632 & 4,892 & 337 candidates and method comparisons \\
DiffusionDB-clean10k & 10,000 & 30,000 & selected schedules and methods \\
\midrule
Total & & 37,151 & \\
\bottomrule
\end{tabular}
\end{table}

Table~\ref{tab:k41-main-results} compares the selected schedules with MeanCache
and BudCache schedules and the adaptive SeaCache method, all using residual reuse.
For DrawBench, GenEval-style, and PartiPrompts, we average each metric over
prompts within each dataset and seed, then give the nine means equal weight.
The schedule selected by mean PSNR has the highest mean PSNR and SSIM and
the lowest mean LPIPS among the results reported for these three datasets.
On DiffusionDB, it has higher PSNR than SeaCache, while SeaCache has higher
SSIM and lower LPIPS.

\begin{table}[!htbp]
\caption{The schedule selected by mean PSNR transfers beyond its four scoring examples.
All fixed schedules use residual reuse.
SeaCache uses its original adaptive schedule and residual-reuse rule.  The
three-dataset results exclude the four selection runs.
MeanCache's SSIM and LPIPS are reported on the full 7,155-run scope in
Appendix~\ref{app:k41-payloads}. Its DiffusionDB row uses all 30,000 runs.
PSNR is in dB.  Bold marks the best value within each scope.}
\label{tab:k41-main-results}
\centering
\small
\begin{tabular}{llrrr}
\toprule
Scope & Method or schedule & PSNR $\uparrow$ & SSIM $\uparrow$ & LPIPS $\downarrow$ \\
\midrule
Three datasets & selected by four-run mean & \textbf{20.944} & \textbf{0.7638} & \textbf{0.2880} \\
Three datasets & selected by four-run minimum & 20.782 & 0.7608 & 0.2890 \\
Three datasets & MeanCache schedule & 20.766 & & \\
Three datasets & BudCache schedule & 20.742 & 0.7583 & 0.2922 \\
Three datasets & SeaCache & 19.916 & 0.7551 & 0.2914 \\
\midrule
DiffusionDB & selected by four-run mean & \textbf{20.485} & 0.6981 & 0.3963 \\
DiffusionDB & selected by four-run minimum & 20.370 & 0.6961 & 0.3952 \\
DiffusionDB & MeanCache schedule & 20.091 & 0.6732 & 0.4144 \\
DiffusionDB & BudCache schedule & 20.415 & 0.6939 & 0.3986 \\
DiffusionDB & SeaCache & 20.061 & \textbf{0.7024} & \textbf{0.3687} \\
\bottomrule
\end{tabular}
\end{table}

Under residual reuse, the schedule selected by mean PSNR also exceeds the
MeanCache schedule by 0.178 dB in mean PSNR, giving the first three datasets
equal weight.
Within each dataset--seed combination, we compute the fraction of prompts
for which the schedule selected by mean PSNR outperforms SeaCache.
Averaging these fractions equally across combinations gives 72.3\% for PSNR,
56.2\% for SSIM, and 45.9\% for LPIPS.  The corresponding DiffusionDB
fractions are 61.2\%, 37.6\%, and 21.8\%.

\subsubsection{Schedule rankings across datasets}

Across the nine dataset--seed combinations, Spearman correlation between the
ranking by four-run mean PSNR and the larger evaluation ranking ranges from 0.896 to
0.946.  In each dataset--seed combination, 34 to 42 of the schedules ranked
in the top 64 by the four-run mean remain in the top 64 of the larger evaluation.
The schedule selected by four-run mean PSNR ranks between 30th
and 42nd in the individual combinations. It ranks 36th when schedules are
ranked by their mean PSNR, with equal weight for the nine dataset--seed combinations.

\begin{figure}[!htbp]
  \centering
  \includegraphics[width=\linewidth]{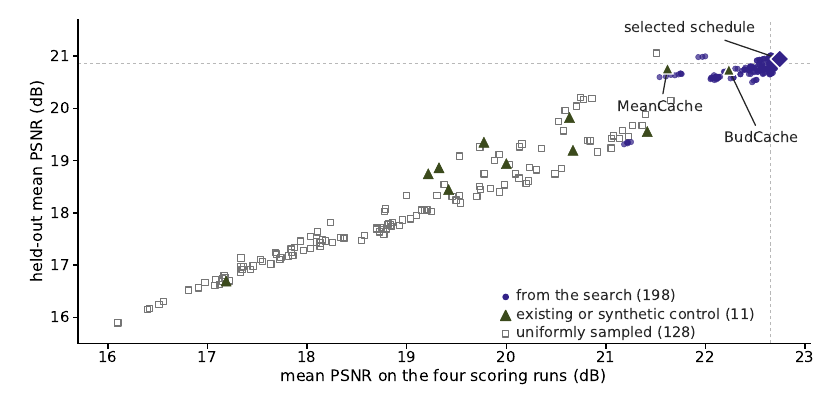}
  \caption{\textbf{Calibration and evaluation scores give similar schedule rankings.}
  Each point is one of the 337 schedules in the set.  Its
  horizontal position is its mean PSNR on the four prompt--seed runs used
  for selection.  Its vertical position is its mean PSNR on the 7,151
  held-out prompt--seed runs, with the nine dataset--seed combinations weighted equally.  We
  obtain a Spearman rank correlation of 0.926 across the set.
  Indigo circles show search-derived schedules, dark olive triangles show
  existing or synthetic controls, and open gray squares show the 128
  uniformly sampled schedules.  The dashed lines mark the
  best 64 schedules on each axis. 39
  schedules are in the best 64 on both. The diamond is the schedule selected
  by four-run mean PSNR. It ranks 36th of 337 by mean PSNR on the held-out runs.
  Its mean PSNR is 0.18 dB above the MeanCache schedule's and 0.20 dB above
  the BudCache schedule's. The best held-out schedule in the
  set is one of the uniform draws.
  }
  \label{fig:k41-transfer}
\end{figure}

Figure~\ref{fig:k41-transfer} plots both scores against each other for the
whole set.

\subsubsection{Five approximation policies}
\label{app:k41-payloads}

We evaluate each of four schedules with each of five approximation policies. The schedules
are those selected by the mean and minimum PSNR of the four scoring runs,
the schedule with the highest mean PSNR on the 337-candidate evaluation,
and the MeanCache schedule.
The approximation policies are residual reuse, first-order Taylor
prediction, second-order Hermite prediction, interval-average velocity, and
two-anchor prediction guided by the first transformer block. Each of the
20 combinations is evaluated on 37,155 prompt--seed runs.  This count includes
the four runs used to select the first two schedules. The 37,151-run transfer
evaluation in Table~\ref{tab:k41-main-results} leaves those four out.
Table~\ref{tab:k41-payload-results} contrasts residual reuse with the
two-anchor rule on the schedule selected by four-run mean.  Its
residual-reuse PSNR is 20.945 dB rather than the 20.944 dB of
Table~\ref{tab:k41-main-results} for that reason.

\begin{table}[!htbp]
\caption{Two-anchor prediction improves all three quality metrics on the schedule
selected by four-run mean.  Three-dataset values equally weight the nine dataset--seed
means.  DiffusionDB values average its three base-seed means.  The four
selection runs are included.  PSNR is in dB.  Bold marks the better approximation policy
within each scope.}
\label{tab:k41-payload-results}
\centering
\small
\begin{tabular}{llrrr}
\toprule
Scope & Approximation policy & PSNR $\uparrow$ & SSIM $\uparrow$ & LPIPS $\downarrow$ \\
\midrule
Three datasets & residual reuse & 20.945 & 0.7638 & 0.2880 \\
Three datasets & two-anchor & \textbf{21.193} & \textbf{0.7865} & \textbf{0.2392} \\
\midrule
DiffusionDB & residual reuse & 20.485 & 0.6981 & 0.3963 \\
DiffusionDB & two-anchor & \textbf{20.929} & \textbf{0.7316} & \textbf{0.3192} \\
\bottomrule
\end{tabular}
\end{table}

Under residual reuse, MeanCache has mean SSIM 0.7459 and mean LPIPS 0.3010
on the full 7,155-run three-dataset evaluation, with equal weight for each
dataset--seed mean.

Two-anchor prediction evaluates the first transformer block at each
cached step.  Its quality gains therefore come with more computation than
residual reuse. For the schedule selected by four-run mean PSNR, the difference
between the highest and lowest mean PSNR across the five approximation policies
is 1.262 dB on the three smaller datasets and 0.899 dB on DiffusionDB.
The schedule with the highest residual-reuse PSNR in the 337-schedule
evaluation set uses full computation at steps 0, 1, 2, 3, 5, 7, 13, 20, and
49. The corresponding differences across approximation policies are
3.963 dB on the three smaller datasets and 3.508 dB on DiffusionDB.
With no full step between steps 20 and 49, its
Taylor and Hermite predictions deteriorate over the long extrapolation
interval. The schedule selected by four-run mean PSNR has later full steps
at 24 and 41.

\section{Additional Full-Trajectory Results}
\label{app:trajectory-details}

Appendix~\ref{app:sampling-conditions} gives the sampling configurations.

\subsection{Step-by-step profiles for the other models}
\label{app:other-model-trajectories}

\begin{figure}[!htbp]
  \centering
  \includegraphics[width=\linewidth]{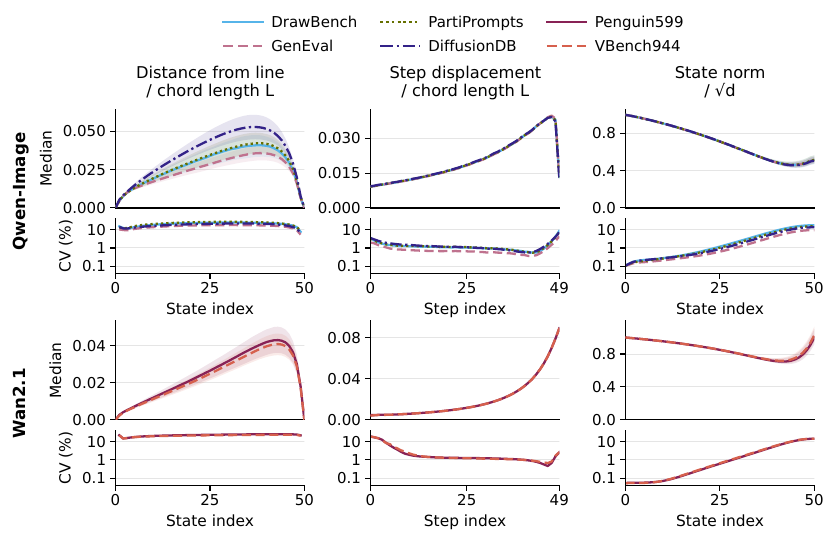}
  \caption{\textbf{Full-compute trajectories show similar patterns across
  datasets in Qwen-Image and Wan2.1.}
  Model groups show Qwen-Image and Wan2.1 from top to bottom.
  Colors and line styles identify datasets.
  Columns show distance from the line joining the initial and final states,
  step displacement, and state norm.
  The first two quantities are divided by each trajectory's chord length.
  State norm is divided by the square root of the latent dimension.
  For each model, upper curves show medians across prompt--seed runs.
  Shaded bands span the middle 50\% of values at each index.
  Lower curves show the coefficient of variation: standard deviation divided
  by the mean across runs at the same index.
  Upper axes use different vertical ranges. Lower axes share a logarithmic
  percentage scale.
  Distance CV omits states 0 and 50, where it is undefined.
  Qwen-Image uses 37,155 runs from four datasets, and Wan2.1 uses
  4,629 runs from two datasets. Every dataset uses three seeds per prompt.
  GenEval and DiffusionDB denote GenEval-style and DiffusionDB-clean10k.
  State $n$ follows $n$ updates, while step $n$ updates state $n$ to $n+1$.}
  \label{fig:full-trajectory-summary-other-models}
\end{figure}

Figure~\ref{fig:full-trajectory-summary-other-models} extends the measurements
in Figure~\ref{fig:full-trajectory-summary} to Qwen-Image and Wan2.1.
Each model group compares all evaluated datasets within that model.

Qwen-Image has a distinct last-step drop in displacement.
Its sampler uses a smaller noise-level change at the final step.

Across datasets, the median CV of normalized step displacement over the
50 steps ranges from 0.67\% to 1.17\% for Qwen-Image and
1.25\% to 1.28\% for Wan2.1.
The CV of normalized step displacement is larger at the start, reaching
19.27\% in Wan2.1.

\subsection{Distance from the line joining the endpoints}

We use the normalized off-chord distance defined in
Section~\ref{sec:trajectory-profiles}.

The normalized off-chord distance rises slowly and returns rapidly.  It
begins at zero, reaches its maximum late in denoising, and returns to zero at
the final latent.  We first take the median normalized distance across all
prompt--seed runs at each state within each dataset.  The peak is the maximum
of this dataset median profile.  Across the four datasets, its height is
0.0925--0.1079 of the chord length for FLUX and 0.0358--0.0529 for
Qwen-Image.  The peak state indices are 40--41 and 37--39, respectively.

\subsection{Numerical precision of turning angles}
\label{app:turning-precision}

The image turning profiles use the first 120 prompts from each dataset
and seed, giving 2,880 trajectories across the two image models.
The video profiles use all trajectories described in
Appendix~\ref{app:sampling-conditions}.

The stored image trajectory states use BF16.  To estimate the rounding floor,
we replace each three-state window's middle state by the exact midpoint of
its endpoints in FP64, round it to BF16, and measure the resulting angle.
The signal-to-floor ratio divides the median measured angle by this median
rounding-induced angle at the same step.  A one-step angle is too small to
separate reliably from rounding, so
Equation~\ref{eq:turn-angle} uses a multi-step window.  For FLUX, $w=5$
produces a signal-to-rounding-floor ratio of 3.42--25.5 over the complete
trajectory.  Qwen-Image is straighter in the middle. Its minimum ratio at
$w=5$ is 2.61. On a subset of 30 Qwen-Image trajectories, increasing the window
gives minimum ratios 3.76, 5.13, and 6.54 for $w=7,9,11$.
The profiles across datasets use $w=5$.
The measurements at $w\geq7$ assess numerical precision on the
30-trajectory subset.

\subsection{Samples for the spatial-direction measurements}
\label{app:plane-samples}

Figure~\ref{fig:trajectory-geometry}(a) uses the first three DrawBench
prompts from the stored FLUX.1-dev full-compute trajectories.
Their zero-based prompt indices are 0, 1, and 2.
The base seed is 41, giving actual seeds 41, 42, and 43.
Each example contains all 51 latent states and uses its own chord length
and fitted principal components.
The PCA fit uses the complete sequence.
The curve through states 0--40 is emphasized, and the segment from state
40 through state 50 remains visible as a pale dashed line.
The display applies a rotation and translation to each plane and one common
scale to both coordinates of all three curves.
The endpoint states have identical off-chord coordinates.
The first two components explain 95.02\%, 95.17\%, and 94.56\% of centered
off-chord variance in these three examples.

The plane-angle measurements in Figure~\ref{fig:trajectory-geometry}(c)
use the first 120 prompts from each of four datasets and three seeds,
giving 1,440 trajectories per image model.
Pairs with identical initial noise contain different prompts.
Pairs with the same prompt use different initial noise.
For FLUX.1-dev, these groups contain 7,856 and 1,440 pairs.
For Qwen-Image, they contain 2,160 and 1,440 pairs.
Each model also uses 20,000 pairs sharing neither prompt nor noise and
500 random-plane comparisons at the same latent dimension.
The same-noise pairs come from 122 distinct initial-noise tensors for FLUX
and 360 for Qwen-Image.

\input{\arxivroot mechanism_appendix}

\section{Schedule Search}
\label{app:search}

\subsection{Search space, scoring examples, and evaluation}

The search of Section~\ref{sec:search} runs on FLUX.1-dev and Qwen-Image at
$K=29$, $37$, and $41$, with 50 FlowMatchEuler steps and residual reuse at every
cached step. Steps 0, 1, 2, and 49 always run the full model. The remaining
full steps are chosen among steps 3 to 48. The spaces contain $1{,}749{,}695{,}026{,}860$
schedules at $K=29$, $1{,}101{,}716{,}330$ at $K=37$, and $1{,}370{,}754$ at
$K=41$.

To evaluate a schedule, we generate images from the eight prompt--seed
examples in Table~\ref{tab:search-calibration}. We compare each image with
the full-compute output from the same prompt and seed, then average the eight
PSNR values. The eight examples were fixed
before any search ran.  None of the eight prompts appears in DrawBench,
GenEval-style, PartiPrompts, or DiffusionDB-clean10k.  Repeat visits to an
already scored schedule reuse its stored score and consume no additional
evaluations.  Initial schedules include recurring schedules selected using
544 PartiPrompts. The final PartiPrompts results include that selection set.
DrawBench, GenEval-style, and DiffusionDB evaluation prompts are separate from
both initialization data and candidate-scoring examples.

\begin{table}[!htbp]
\caption{The eight scoring examples.  Base seeds 50042 and 60042 differ from every
evaluation base seed.}
\label{tab:search-calibration}
\centering
\small
\begin{tabular}{llr}
\toprule
Prompt category & Source & Base seed \\
\midrule
person & COCO 2014 val & 50042 \\
animal & COCO 2014 val & 50042 \\
indoor object & COCO 2014 val & 50042 \\
outdoor scene & COCO 2014 val & 50042 \\
stylized A & DiffusionDB calibration prompt set & 60042 \\
stylized B & DiffusionDB calibration prompt set & 60042 \\
counting & COCO 2014 val & 60042 \\
rendered text & fixed template & 60042 \\
\bottomrule
\end{tabular}
\end{table}

\subsection{Search parameters}

A swap exchanges one cached step with one full step whose position is not
required by the search-space constraints. It preserves the cache count and
the required full steps.
Hill climbing examines the one-swap neighbors in random order and accepts
the first improvement. Greedy coordinate ascent optimizes the movable
full-step positions one at a time.

Each procedure receives the same evaluation budget within a model and cache
ratio: 1,400 at $K=29$, 700 at $K=37$, and 400 at $K=41$.
Before search, we evaluate 25 pairs, each containing a uniformly sampled
schedule and a one-swap neighbor.
These 50 evaluations measure the scale of PSNR changes and count toward the budget.
For annealing, we set $t_{\max}$ to ten times the median absolute change
across these pairs and $t_{\min}$ to that median divided by 100.
The temperatures are rounded to two significant figures.
Table~\ref{tab:search-probe} gives the measured changes and temperatures.

The schedules used to start optimization are separate from these 50 schedules.
Hill climbing and annealing start from a shuffled list of six to eight
schedules from BudCache, MeanCache, DPCache, uniform spacing, and the most
frequent adaptive schedules that satisfy the search-space constraints.
Each procedure optimizes from these schedules in turn, then uses random
restarts if its budget remains. Greedy coordinate ascent starts from a random schedule.

An annealing proposal first selects a cached step uniformly.
With probability 0.7, it exchanges that step with a movable full step at
most three positions away, if one exists.
Otherwise, it selects the full step uniformly from all movable full steps.
Each annealing chain uses 680, 360, or 200 proposals at $K=29,37,41$,
followed by up to 20 rounds of steepest improvement using swaps within
three step positions. All stages are subject to the evaluation budget.

Hill climbing also stops after two consecutive completed climbs that each
improve the best mean score by less than twice its standard error.
Annealing applies this rule to completed chains followed by local improvement.
Greedy coordinate ascent applies it to completed coordinate sweeps.
This standard error is computed from the current best schedule's eight scores.
Random search uses only the evaluation limit.
All PSNR searches exhausted their evaluation budgets.

\begin{table}[!htbp]
\caption{Initial score differences determine the annealing temperatures.
The PSNR range and standard deviation are over the 50 initial schedules.
The swap delta is the absolute change in mean PSNR on the eight scoring runs
between schedules that differ by one swap.  Median SE is the median standard
error across the 50 schedules, with each standard error computed from that
schedule's eight scores.  All values are in dB.}
\label{tab:search-probe}
\centering
\small
\begin{tabular}{llrrrrrr}
\toprule
Model & $K$ & PSNR range & PSNR sd & Median swap delta & Median SE & $t_{\max}$ & $t_{\min}$ \\
\midrule
FLUX.1-dev & 29 & 15.45--22.37 & 1.430 & 0.0236 & 0.625 & 0.24 & 0.00024 \\
FLUX.1-dev & 37 & 14.79--18.92 & 1.103 & 0.0433 & 0.589 & 0.43 & 0.00043 \\
FLUX.1-dev & 41 & 14.64--18.38 & 0.921 & 0.0536 & 0.537 & 0.54 & 0.00054 \\
Qwen-Image & 29 & 14.07--21.30 & 1.603 & 0.0174 & 0.693 & 0.17 & 0.00017 \\
Qwen-Image & 37 & 12.93--19.48 & 1.552 & 0.0347 & 0.542 & 0.35 & 0.00035 \\
Qwen-Image & 41 & 11.73--17.23 & 1.427 & 0.0766 & 0.585 & 0.77 & 0.00077 \\
\bottomrule
\end{tabular}
\end{table}

\subsection{Candidate validation and selection}

Random sampling, hill climbing, simulated annealing, and greedy coordinate
ascent each retain their best three candidates that differ from one another
at four or more step positions. At each such position, one schedule uses a
full step and the other uses a cached step. All candidates for a model--ratio
combination are re-scored on 50 separate validation prompts from COCO 2014
captions at seed 50042.
To select these prompts, we order the COCO 2014 captions by annotation ID
and take the first 50 after the largest
ID used by the scoring examples. Each procedure
selects the candidate with the highest validation mean.  Identical schedules
are merged across procedures.  For each model--ratio combination,
Section~\ref{sec:search} reports the candidate with the highest validation
mean across all four procedures. Table~\ref{tab:search-delivered} lists the six selected schedules.
Table~\ref{tab:search-arbitration} shows the eight procedure--model--ratio
combinations where validation changes the candidate selected by the eight-example mean.

\begin{table}[!htbp]
\caption{The selected schedule for each model--ratio combination.  Full steps are the positions
that run the full model. Every other step of the 50 is cached.  Validation
PSNR is the mean over the 50 validation prompts.}
\label{tab:search-delivered}
\centering
\small
\begin{tabularx}{\linewidth}{@{}llrX@{}}
\toprule
Model & $K$ & Validation PSNR (dB) & Full steps \\
\midrule
FLUX.1-dev & 29 & 30.705 & 0, 1, 2, 3, 4, 5, 6, 7, 8, 9, 10, 11, 13, 15, 17, 19, 24, 30, 38, 46, 49 \\
FLUX.1-dev & 37 & 24.832 & 0, 1, 2, 3, 4, 6, 8, 10, 14, 19, 30, 43, 49 \\
FLUX.1-dev & 41 & 21.552 & 0, 1, 2, 3, 5, 8, 13, 27, 49 \\
Qwen-Image & 29 & 30.381 & 0, 1, 2, 3, 4, 5, 6, 7, 8, 9, 10, 12, 14, 16, 19, 22, 26, 33, 40, 46, 49 \\
Qwen-Image & 37 & 24.499 & 0, 1, 2, 3, 4, 5, 6, 8, 11, 14, 22, 35, 49 \\
Qwen-Image & 41 & 20.766 & 0, 1, 2, 3, 5, 9, 17, 31, 49 \\
\bottomrule
\end{tabularx}
\end{table}

\input{\arxivroot tables/search_arbitration}

\Needspace{14\baselineskip}
\subsection{Results on five quality measures}
\label{app:search-metrics}

\input{\arxivroot tables/search_paired_full}

\subsection{Schedules selected by different objectives}
\label{app:search-objectives}

Section~\ref{sec:search} compares schedules selected using PSNR and LPIPS.
We also repeat hill climbing and annealing over the same eight scoring
examples with the sum of standardized mean PSNR and standardized
negative mean LPIPS.  Standardization uses the mean and standard deviation of
each metric's eight-example average across the 50 initial schedules in that
model--ratio combination.  Thus the joint objective is
$({\rm PSNR}-\mu_P)/\sigma_P-({\rm LPIPS}-\mu_L)/\sigma_L$.
Each objective uses the same 50 validation prompts and four evaluation
datasets, including the full PartiPrompts set described above.

Compared with PSNR-selected schedules, schedules selected by the standardized
PSNR--LPIPS objective also have higher SSIM and lower LPIPS in all 48 tests
and higher PSNR in 26 of them. Changing the objective therefore changes the
quality tradeoff among schedules in the same search space.

\subsection{Changing the approximation policy}
\label{app:search-policies}

We also hold each of the six PSNR-selected schedules fixed and replace residual
reuse with first-order Taylor prediction, second-order Hermite prediction,
interval-average velocity, or two-anchor extrapolation.  Each combination is
evaluated on the same four datasets and three base seeds.  Interval-average
velocity uses MeanCache's average-velocity predictor with its trajectory-based
correction. Two-anchor prediction uses the first transformer block to estimate an
extrapolation coefficient from the two most recent full-step residuals.
Two-anchor
extrapolation gives the highest equal-weight mean PSNR in all six model--ratio combinations.  Its
gain over residual reuse is 1.672, 0.989, and 0.071 dB for FLUX.1-dev, and
1.433, 0.454, and 0.157 dB for Qwen-Image, at cache ratios 0.58, 0.74, and 0.82.

\input{\arxivroot tables/search_payloads}

\subsection{Comparing search procedures using exhaustive scores}

We compare ten search algorithms using the scores from the exhaustive
FLUX.1-dev experiment at $K=41$. Looking up a schedule's stored score counts
as one evaluation.
We use these comparisons to choose the four search procedures and their
hyperparameters for Section~\ref{sec:search}.
At each evaluation count, we subtract the best score found so far from
the highest exhaustive score of 22.746 dB and average this gap over 50 searches.
Table~\ref{tab:search-bench} reports the results.
Unseeded methods use independent random starts. Seeded
variants first evaluate the fixed schedules specified by their method and then
use random restarts.

\begin{table}[!htbp]
\caption{Ten search algorithms evaluated using the stored scores of all
1,370,754 FLUX.1-dev schedules that cache 41 of 50 steps.
The columns 50 to 5,000 give the number of candidate evaluations.
Gaps are PSNR differences in dB below the best exhaustive score.
Each gap is averaged over 50 searches.
0.000 means the algorithm found that score exactly.
``Exact by 1,000'' counts how many of the 50 searches found the best
exhaustive score within 1,000 candidate evaluations. The last column is the
median evaluation count among repetitions that reached a gap of 0.05 dB.
A dash indicates that fewer than 25 of the 50 repetitions reached this gap
within 5,000 evaluations.
The annealing implementation follows BudCache~\citep{budcache2026}, with
temperatures scaled to the observed score differences.}
\label{tab:search-bench}
\centering
\small
\setlength{\tabcolsep}{4pt}
\begin{tabular}{lrrrrrr}
\toprule
Search algorithm & 50 & 200 & 1,000 & 5,000 & Exact by 1,000 & To 0.05 dB \\
\midrule
Random sampling & 1.19 & 0.74 & 0.38 & 0.18 & 0/50 & -- \\
Hill climb, first improvement & 0.75 & 0.18 & 0.010 & 0.000 & 21/50 & 324 \\
Hill climb, steepest ascent & 1.82 & 1.49 & 0.055 & 0.0003 & 9/50 & 802 \\
Hill climb, seeded & 0.57 & 0.16 & 0.005 & 0.000 & 25/50 & 308 \\
Annealing & 1.02 & 0.08 & 0.000 & 0.000 & 50/50 & 171 \\
Annealing, seeded & 0.65 & 0.05 & 0.000 & 0.000 & 50/50 & 172 \\
Greedy coordinate ascent & 1.81 & 0.26 & 0.003 & 0.000 & 32/50 & 550 \\
Learned additive cost & 1.06 & 0.58 & 0.19 & 0.14 & 3/50 & -- \\
Learned pairwise cost & 1.08 & 0.27 & 0.14 & 0.083 & 9/50 & 723 \\
Bayesian optimization & 1.18 & 0.34 & 0.035 & 0.0014 & 18/50 & 642 \\
\bottomrule
\end{tabular}
\end{table}

The exhaustive scores also show how many schedules approach the highest mean PSNR.
Three schedules are within 0.01 dB of that score, 12 within 0.05 dB,
76 within 0.10 dB, and 247 within 0.20 dB.
Steepest ascent from 2,000 random starts under the one-swap neighborhood
converges to one of three schedules: the exhaustive best in 54.6\% of climbs, a
second optimum 0.0057 dB below it in 39.0\%, and a third 0.26 dB below it in
6.3\%.

We also measure how reducing the number of scoring runs affects schedule selection.
For every subset of $n$ of the four scoring runs, we select the schedule
with the highest $n$-run mean PSNR over the whole space.
We then evaluate that schedule by its four-run mean PSNR.
The mean loss relative to the best four-run mean is
0.315 dB over the four single-run subsets, 0.284 dB over the six two-run
subsets, and 0.219 dB over the four three-run subsets.  The individual subset
losses run from 0.000 to 1.017 dB, showing that the choice of scoring
examples affects which schedule is selected.

%% file: tables/full_results_flux.tex
\begingroup
\fontsize{9.5}{11}\selectfont
\setlength{\tabcolsep}{1pt}
\setlength{\aboverulesep}{0.5pt}
\setlength{\belowrulesep}{0.5pt}
\setlength{\LTpre}{2pt}
\renewcommand{\arraystretch}{0.95}
\setlength{\LTcapwidth}{\linewidth}
\begin{longtable}{@{}ll*{3}{rrrrr}@{}}
\caption{Complete FLUX.1-dev results, averaged over three base seeds.
P is PSNR in dB, S is SSIM, and L is LPIPS. C is CLIP score and IR is ImageReward.
$\rho=K/50$ is the target cache ratio. Bold marks the best unrounded mean
within each dataset, ratio, and metric.}
\label{tab:flux-full-results}\\
\toprule
& & \multicolumn{5}{c}{$\rho=.58$ ($K=29$)} & \multicolumn{5}{c}{$\rho=.74$ ($K=37$)} & \multicolumn{5}{c}{$\rho=.82$ ($K=41$)} \\
\cmidrule(lr){3-7}\cmidrule(lr){8-12}\cmidrule(l){13-17}
Data & Method & P$\uparrow$ & S$\uparrow$ & L$\downarrow$ & C$\uparrow$ & IR$\uparrow$ & P$\uparrow$ & S$\uparrow$ & L$\downarrow$ & C$\uparrow$ & IR$\uparrow$ & P$\uparrow$ & S$\uparrow$ & L$\downarrow$ & C$\uparrow$ & IR$\uparrow$ \\
\midrule
\endfirsthead
\caption[]{Continued.}\\
\toprule
& & \multicolumn{5}{c}{$\rho=.58$ ($K=29$)} & \multicolumn{5}{c}{$\rho=.74$ ($K=37$)} & \multicolumn{5}{c}{$\rho=.82$ ($K=41$)} \\
\cmidrule(lr){3-7}\cmidrule(lr){8-12}\cmidrule(l){13-17}
Data & Method & P$\uparrow$ & S$\uparrow$ & L$\downarrow$ & C$\uparrow$ & IR$\uparrow$ & P$\uparrow$ & S$\uparrow$ & L$\downarrow$ & C$\uparrow$ & IR$\uparrow$ & P$\uparrow$ & S$\uparrow$ & L$\downarrow$ & C$\uparrow$ & IR$\uparrow$ \\
\midrule
\endhead
\bottomrule
\endfoot
DB & Sea & 27.208 & .9066 & .0845 & 27.619 & .9983 & 21.435 & .8006 & .2114 & 27.627 & .9897 & 19.407 & \textbf{.7365} & \textbf{.3078} & 27.621 & .9537 \\
DB & Tea & 18.497 & .7463 & .2759 & 27.604 & 1.0032 & 16.868 & .6880 & .3628 & 27.657 & .9765 & 15.458 & .6307 & .4590 & 27.531 & .8569 \\
DB & Sen & 27.769 & .9083 & .0834 & 27.568 & .9943 & 22.437 & .7711 & .2740 & \textbf{27.723} & .8880 & 19.826 & .6927 & .3876 & 27.700 & .8271 \\
DB & Di & 28.200 & .9172 & .0721 & 27.590 & .9964 & 21.397 & .7157 & .3861 & 27.652 & .7174 & 19.070 & .6981 & .4010 & 27.579 & .7091 \\
DB & Tay O1 & 21.442 & .8250 & .1713 & 27.521 & .9812 & 17.571 & .7194 & .3017 & 27.516 & .9893 & 14.275 & .5709 & .5221 & 27.204 & .7305 \\
DB & Hi O2 & 21.981 & .8322 & .1622 & 27.636 & .9981 & 18.071 & .7301 & .2885 & 27.449 & .9925 & 14.275 & .5689 & .5116 & 27.348 & .7608 \\
DB & L2P & 22.742 & .8460 & .1481 & 27.662 & \textbf{1.0093} & 16.467 & .6055 & .4284 & 27.339 & .7865 & 16.888 & .5415 & .5230 & 27.448 & .6667 \\
DB & DP & 24.396 & .8797 & .1120 & 27.555 & .9959 & 20.593 & .7922 & .2067 & 27.574 & \textbf{1.0199} & 17.025 & .6815 & .3500 & 27.755 & \textbf{.9887} \\
DB & Bud & 23.655 & .8551 & .1390 & \textbf{27.681} & 1.0027 & 22.553 & .8165 & .1934 & 27.657 & .9879 & \textbf{20.099} & .7352 & .3153 & 27.607 & .9082 \\
DB & Mean & \textbf{29.576} & \textbf{.9307} & \textbf{.0590} & 27.631 & 1.0034 & \textbf{23.930} & \textbf{.8449} & \textbf{.1507} & 27.653 & .9968 & 19.939 & .7058 & .3364 & \textbf{27.947} & .9198 \\
\midrule
PP & Sea & 27.608 & .9091 & .0836 & 27.459 & 1.1947 & 22.191 & .8127 & .2042 & 27.545 & 1.1816 & 20.065 & \textbf{.7444} & \textbf{.3047} & 27.681 & 1.1629 \\
PP & Tea & 19.206 & .7632 & .2610 & 27.439 & 1.2017 & 17.345 & .7026 & .3502 & 27.534 & 1.1822 & 15.745 & .6411 & .4522 & 27.533 & 1.1225 \\
PP & Sen & 27.878 & .9092 & .0849 & 27.459 & 1.1942 & 22.576 & .7680 & .2863 & \textbf{27.738} & 1.0673 & 20.201 & .6927 & .3926 & 27.600 & 1.0222 \\
PP & Di & 28.502 & .9212 & .0695 & 27.396 & 1.1986 & 21.480 & .7105 & .3978 & 27.542 & .8851 & 19.461 & .7048 & .3998 & 27.754 & .9303 \\
PP & Tay O1 & 22.395 & .8422 & .1556 & 27.402 & 1.2018 & 18.098 & .7325 & .2916 & 27.424 & 1.1939 & 14.382 & .5688 & .5258 & 27.616 & 1.0275 \\
PP & Hi O2 & 23.088 & .8515 & .1433 & \textbf{27.459} & 1.1995 & 18.705 & .7473 & .2724 & 27.487 & 1.1906 & 14.405 & .5732 & .5069 & 27.437 & 1.0118 \\
PP & L2P & 23.970 & .8655 & .1281 & 27.408 & \textbf{1.2022} & 17.225 & .6260 & .4123 & 27.488 & 1.0070 & 17.417 & .5520 & .5155 & 27.784 & .9394 \\
PP & DP & 25.312 & .8915 & .1002 & 27.407 & 1.1963 & 21.370 & .8091 & .1916 & 27.428 & \textbf{1.2032} & 17.442 & .6946 & .3401 & 27.585 & \textbf{1.1889} \\
PP & Bud & 24.557 & .8687 & .1271 & 27.453 & 1.1928 & 23.132 & .8238 & .1913 & 27.558 & 1.1797 & \textbf{20.747} & .7433 & .3139 & 27.712 & 1.1079 \\
PP & Mean & \textbf{29.760} & \textbf{.9330} & \textbf{.0572} & 27.421 & 1.1990 & \textbf{24.482} & \textbf{.8507} & \textbf{.1475} & 27.470 & 1.1928 & 20.482 & .7125 & .3335 & \textbf{27.991} & 1.1287 \\
\midrule
GE & Sea & 29.312 & .9357 & .0598 & 28.172 & 1.0007 & 22.606 & .8424 & .1749 & 28.204 & .9935 & 20.275 & .7843 & .2617 & 28.306 & .9843 \\
GE & Tea & 18.988 & .7777 & .2633 & 28.070 & .9993 & 16.990 & .7254 & .3490 & 28.133 & .9628 & 15.545 & .6817 & .4363 & 28.060 & .8710 \\
GE & Sen & 29.885 & .9410 & .0549 & 28.177 & 1.0019 & 24.173 & .8372 & .1954 & 28.457 & .9503 & 21.256 & .7551 & .3221 & 28.291 & .9048 \\
GE & Di & 30.404 & .9469 & .0476 & 28.092 & 1.0031 & 23.006 & .7765 & .3251 & 28.540 & .8533 & 20.296 & .7663 & .3143 & 28.489 & .7990 \\
GE & Tay O1 & 22.190 & .8499 & .1575 & 28.072 & 1.0075 & 17.892 & .7538 & .2831 & 27.978 & .9919 & 14.208 & .6031 & .5080 & 27.841 & .7577 \\
GE & Hi O2 & 22.969 & .8606 & .1434 & \textbf{28.177} & 1.0163 & 18.507 & .7658 & .2671 & 28.126 & 1.0019 & 14.370 & .6056 & .5004 & 27.687 & .7890 \\
GE & L2P & 23.906 & .8738 & .1289 & 28.139 & \textbf{1.0168} & 17.337 & .6491 & .4217 & \textbf{28.572} & .8499 & 17.630 & .5527 & .5426 & 28.495 & .7881 \\
GE & DP & 25.133 & .8949 & .1050 & 28.077 & 1.0023 & 21.392 & .8234 & .1893 & 28.043 & \textbf{1.0136} & 17.642 & .7217 & .3312 & 27.976 & \textbf{.9866} \\
GE & Bud & 25.109 & .8895 & .1109 & 28.161 & 1.0051 & 24.122 & .8628 & .1480 & 28.227 & .9952 & \textbf{21.381} & \textbf{.7965} & \textbf{.2472} & 28.360 & .9598 \\
GE & Mean & \textbf{31.808} & \textbf{.9559} & \textbf{.0382} & 28.135 & 1.0039 & \textbf{25.620} & \textbf{.8853} & \textbf{.1156} & 28.129 & .9995 & 21.318 & .7629 & .2985 & \textbf{28.664} & .9558 \\
\midrule
DDB & Sea & 25.926 & .8729 & .1203 & 27.798 & .9382 & 21.814 & .7737 & .2540 & 27.890 & .9182 & 20.061 & \textbf{.7024} & \textbf{.3687} & \textbf{28.003} & .8760 \\
DDB & Tea & 19.549 & .7378 & .2865 & \textbf{27.878} & .9589 & 17.575 & .6685 & .3867 & 27.918 & .9276 & 15.904 & .6016 & .4958 & 27.892 & .8562 \\
DDB & Sen & 25.938 & .8673 & .1286 & 27.795 & .9371 & 21.472 & .7067 & .3903 & 27.638 & .6980 & 19.679 & .6539 & .4744 & 27.575 & .6636 \\
DDB & Di & 26.351 & .8813 & .1092 & 27.784 & .9520 & 20.452 & .6541 & .5161 & 26.548 & .2812 & 18.948 & .6550 & .4928 & 27.534 & .5655 \\
DDB & Tay O1 & 22.372 & .8180 & .1805 & 27.797 & \textbf{.9644} & 18.385 & .7056 & .3215 & 27.895 & \textbf{.9743} & 14.985 & .5656 & .5262 & 27.687 & .8925 \\
DDB & Hi O2 & 22.851 & .8243 & .1719 & 27.839 & .9589 & 19.007 & .7202 & .3032 & \textbf{27.927} & .9611 & 15.051 & .5782 & .4997 & 27.789 & .8787 \\
DDB & L2P & 23.611 & .8396 & .1551 & 27.822 & .9607 & 17.253 & .6039 & .4143 & 27.074 & .7824 & 17.501 & .5650 & .4964 & 27.373 & .7281 \\
DDB & DP & 25.057 & .8716 & .1190 & 27.757 & .9578 & 21.334 & .7798 & .2214 & 27.848 & .9711 & 17.491 & .6618 & .3742 & 27.961 & \textbf{.9577} \\
DDB & Bud & 23.780 & .8337 & .1652 & 27.799 & .9370 & 22.246 & .7767 & .2538 & 27.883 & .9074 & \textbf{20.415} & .6939 & .3986 & 27.862 & .7943 \\
DDB & Mean & \textbf{27.595} & \textbf{.8994} & \textbf{.0894} & 27.776 & .9502 & \textbf{23.331} & \textbf{.8055} & \textbf{.1991} & 27.846 & .9427 & 19.894 & .6630 & .3901 & 27.897 & .8678 \\
\end{longtable}
\endgroup

%% file: tables/full_results_qwen.tex
\begingroup
\fontsize{9.5}{11}\selectfont
\setlength{\tabcolsep}{1pt}
\setlength{\aboverulesep}{0.5pt}
\setlength{\belowrulesep}{0.5pt}
\setlength{\LTpre}{2pt}
\renewcommand{\arraystretch}{0.95}
\setlength{\LTcapwidth}{\linewidth}
\begin{longtable}{@{}ll*{3}{rrrrr}@{}}
\caption{Complete Qwen-Image results, averaged over three base seeds.
P is PSNR in dB, S is SSIM, and L is LPIPS. C is CLIP score and IR is ImageReward.
$\rho=K/50$ is the target cache ratio. Bold marks the best unrounded mean
within each dataset, ratio, and metric.}
\label{tab:qwen-full-results}\\
\toprule
& & \multicolumn{5}{c}{$\rho=.58$ ($K=29$)} & \multicolumn{5}{c}{$\rho=.74$ ($K=37$)} & \multicolumn{5}{c}{$\rho=.82$ ($K=41$)} \\
\cmidrule(lr){3-7}\cmidrule(lr){8-12}\cmidrule(l){13-17}
Data & Method & P$\uparrow$ & S$\uparrow$ & L$\downarrow$ & C$\uparrow$ & IR$\uparrow$ & P$\uparrow$ & S$\uparrow$ & L$\downarrow$ & C$\uparrow$ & IR$\uparrow$ & P$\uparrow$ & S$\uparrow$ & L$\downarrow$ & C$\uparrow$ & IR$\uparrow$ \\
\midrule
\endfirsthead
\caption[]{Continued.}\\
\toprule
& & \multicolumn{5}{c}{$\rho=.58$ ($K=29$)} & \multicolumn{5}{c}{$\rho=.74$ ($K=37$)} & \multicolumn{5}{c}{$\rho=.82$ ($K=41$)} \\
\cmidrule(lr){3-7}\cmidrule(lr){8-12}\cmidrule(l){13-17}
Data & Method & P$\uparrow$ & S$\uparrow$ & L$\downarrow$ & C$\uparrow$ & IR$\uparrow$ & P$\uparrow$ & S$\uparrow$ & L$\downarrow$ & C$\uparrow$ & IR$\uparrow$ & P$\uparrow$ & S$\uparrow$ & L$\downarrow$ & C$\uparrow$ & IR$\uparrow$ \\
\midrule
\endhead
\bottomrule
\endfoot
DB & Sea & 27.437 & .9192 & .0765 & 29.226 & 1.2162 & 21.385 & .8267 & .1873 & 29.113 & 1.1725 & 16.778 & .7007 & .3526 & 29.002 & 1.1046 \\
DB & Tea & 20.363 & .8218 & .1796 & 29.217 & 1.2135 & 17.022 & .7146 & .3163 & 29.073 & 1.1344 & 14.457 & .6162 & .4615 & 28.793 & .9185 \\
DB & Sen & 28.182 & .9252 & .0707 & 29.181 & 1.2186 & 21.825 & .7634 & .2641 & 28.768 & 1.0396 & 18.904 & .7092 & .3339 & 28.883 & .9792 \\
DB & Di & 30.210 & .9463 & .0458 & \textbf{29.242} & 1.2233 & 20.468 & .7283 & .3225 & 28.722 & .8573 & 17.205 & .6562 & .4084 & 28.381 & .6534 \\
DB & Tay O1 & 20.834 & .8368 & .1613 & 29.142 & 1.2188 & 13.984 & .6163 & .4399 & 28.220 & .9413 & 9.680 & .4308 & .6760 & 26.485 & .1029 \\
DB & Hi O2 & 21.715 & .8492 & .1452 & 29.217 & 1.2171 & 13.804 & .6039 & .4531 & 28.229 & .9149 & 8.099 & .3894 & .7636 & 23.437 & -1.0827 \\
DB & L2P & 22.670 & .8663 & .1280 & 29.231 & 1.2178 & 18.737 & .7491 & .2724 & 29.049 & 1.1697 & 14.421 & .5387 & .5467 & 27.859 & .6882 \\
DB & DP & 21.573 & .8478 & .1501 & 29.157 & \textbf{1.2300} & 17.533 & .7366 & .2776 & 29.043 & 1.1830 & 10.693 & .4795 & .6413 & 26.605 & .3402 \\
DB & Bud & 29.627 & .9343 & .0651 & 29.211 & 1.2218 & 24.417 & .8464 & .1719 & \textbf{29.196} & 1.1759 & 19.320 & \textbf{.7564} & \textbf{.2829} & 29.129 & \textbf{1.1238} \\
DB & Mean & \textbf{32.051} & \textbf{.9530} & \textbf{.0438} & 29.206 & 1.2206 & \textbf{24.911} & \textbf{.8608} & \textbf{.1295} & 29.067 & \textbf{1.1967} & \textbf{19.869} & .7234 & .2873 & \textbf{29.292} & 1.1085 \\
\midrule
PP & Sea & 28.278 & .9304 & .0667 & 28.546 & 1.3633 & 22.109 & .8431 & .1715 & \textbf{28.641} & 1.3441 & 17.284 & .7148 & .3315 & 28.669 & 1.2760 \\
PP & Tea & 21.210 & .8419 & .1600 & 28.563 & 1.3607 & 17.555 & .7273 & .2968 & 28.597 & 1.3015 & 14.632 & .6177 & .4547 & 28.453 & 1.0761 \\
PP & Sen & 28.569 & .9320 & .0657 & 28.537 & 1.3600 & 21.894 & .7631 & .2717 & 28.538 & 1.2045 & 19.140 & .7147 & .3370 & 28.511 & 1.1438 \\
PP & Di & 30.705 & .9527 & \textbf{.0399} & 28.510 & 1.3694 & 20.452 & .7211 & .3415 & 28.322 & .9587 & 17.249 & .6566 & .4181 & 28.148 & .7717 \\
PP & Tay O1 & 21.804 & .8576 & .1404 & 28.525 & 1.3736 & 14.300 & .6318 & .4267 & 27.982 & 1.1453 & 9.614 & .4303 & .6730 & 26.636 & .3364 \\
PP & Hi O2 & 22.787 & .8697 & .1247 & \textbf{28.572} & 1.3718 & 14.161 & .6168 & .4358 & 28.019 & 1.0993 & 8.312 & .3945 & .7520 & 23.397 & -1.0213 \\
PP & L2P & 23.639 & .8839 & .1105 & 28.503 & \textbf{1.3745} & 19.386 & .7685 & .2568 & 28.605 & 1.3477 & 14.588 & .5532 & .5376 & 28.123 & .8367 \\
PP & DP & 22.210 & .8625 & .1358 & 28.483 & 1.3733 & 17.837 & .7478 & .2673 & 28.403 & \textbf{1.3610} & 10.685 & .4830 & .6348 & 26.687 & .5402 \\
PP & Bud & 30.127 & .9407 & .0603 & 28.560 & 1.3596 & 24.852 & .8529 & .1709 & 28.598 & 1.3197 & 19.966 & \textbf{.7693} & \textbf{.2721} & 28.635 & 1.2860 \\
PP & Mean & \textbf{32.522} & \textbf{.9573} & .0408 & 28.537 & 1.3650 & \textbf{25.552} & \textbf{.8694} & \textbf{.1255} & 28.502 & 1.3497 & \textbf{20.330} & .7317 & .2824 & \textbf{28.959} & \textbf{1.2983} \\
\midrule
GE & Sea & 30.394 & .9426 & .0454 & 30.661 & 1.3485 & 23.880 & .8663 & .1273 & \textbf{30.716} & 1.3440 & 19.674 & .7783 & .2360 & 30.703 & \textbf{1.3316} \\
GE & Tea & 22.771 & .8645 & .1231 & 30.673 & 1.3530 & 19.159 & .7685 & .2219 & 30.654 & 1.3425 & 16.106 & .6762 & .3635 & 30.318 & 1.2631 \\
GE & Sen & 30.704 & .9454 & .0430 & \textbf{30.679} & 1.3481 & 23.889 & .8014 & .1878 & 30.593 & 1.2997 & 20.832 & .7569 & .2421 & 30.582 & 1.2861 \\
GE & Di & 32.853 & \textbf{.9643} & \textbf{.0238} & 30.626 & 1.3521 & 22.341 & .7669 & .2347 & 30.517 & 1.2335 & 18.473 & .7086 & .3183 & 30.256 & 1.1117 \\
GE & Tay O1 & 23.087 & .8772 & .1127 & 30.564 & 1.3569 & 14.594 & .6565 & .4135 & 29.430 & 1.0941 & 9.477 & .4527 & .6678 & 27.930 & .3777 \\
GE & Hi O2 & 24.198 & .8881 & .0963 & 30.675 & 1.3568 & 14.876 & .6642 & .4027 & 29.622 & 1.0983 & 8.615 & .4372 & .7350 & 25.432 & -.7869 \\
GE & L2P & 25.272 & .9040 & .0805 & 30.602 & 1.3558 & 20.597 & .7904 & .2241 & 30.270 & 1.3399 & 14.596 & .5441 & .5693 & 29.439 & .8622 \\
GE & DP & 24.324 & .8913 & .0954 & 30.520 & \textbf{1.3590} & 19.895 & .7951 & .1994 & 30.269 & \textbf{1.3605} & 11.678 & .5500 & .5842 & 28.647 & .8084 \\
GE & Bud & 32.154 & .9501 & .0400 & 30.664 & 1.3449 & 26.887 & .8724 & .1177 & 30.711 & 1.3334 & 21.265 & \textbf{.7930} & \textbf{.2034} & 30.724 & 1.3289 \\
GE & Mean & \textbf{34.465} & .9635 & .0267 & 30.624 & 1.3481 & \textbf{27.713} & \textbf{.8888} & \textbf{.0854} & 30.504 & 1.3437 & \textbf{21.920} & .7589 & .2414 & \textbf{30.828} & 1.2910 \\
\midrule
DDB & Sea & 26.216 & .9127 & .0945 & 29.805 & 1.3186 & 21.109 & .8197 & .2183 & \textbf{29.782} & 1.2883 & 16.290 & .6615 & .4168 & 29.463 & 1.1681 \\
DDB & Tea & 20.502 & .8193 & .1988 & 29.806 & 1.3158 & 16.873 & .6920 & .3553 & 29.592 & 1.2228 & 12.790 & .5344 & .5877 & 28.185 & .6855 \\
DDB & Sen & 26.272 & .9118 & .0979 & 29.792 & 1.3128 & 20.280 & .7299 & .3582 & 29.117 & 1.0921 & 17.777 & .6747 & .4391 & 28.982 & .9865 \\
DDB & Di & 27.706 & .9315 & .0688 & 29.782 & 1.3252 & 18.663 & .6685 & .4783 & 28.074 & .6158 & 15.893 & .5974 & .5480 & 27.717 & .4255 \\
DDB & Tay O1 & 20.865 & .8301 & .1806 & 29.816 & \textbf{1.3310} & 13.746 & .5959 & .4563 & 29.231 & 1.1502 & 9.174 & .3992 & .7005 & 26.439 & .2964 \\
DDB & Hi O2 & 21.745 & .8465 & .1615 & \textbf{29.846} & 1.3281 & 13.127 & .5651 & .4882 & 29.025 & 1.0891 & 7.455 & .3496 & .8208 & 20.746 & -1.2761 \\
DDB & L2P & 21.738 & .8402 & .1690 & 29.770 & 1.3264 & 18.522 & .7354 & .2989 & 29.438 & 1.2569 & 13.777 & .5099 & .5481 & 27.087 & .5743 \\
DDB & DP & 20.985 & .8309 & .1809 & 29.810 & 1.3302 & 16.886 & .6965 & .3332 & 29.688 & 1.2888 & 9.942 & .4216 & .6860 & 26.219 & .2505 \\
DDB & Bud & 27.898 & .9265 & .0850 & 29.809 & 1.3140 & 23.236 & .8303 & .2279 & 29.729 & 1.2566 & \textbf{19.197} & \textbf{.7456} & \textbf{.3406} & \textbf{29.680} & \textbf{1.2146} \\
DDB & Mean & \textbf{29.900} & \textbf{.9456} & \textbf{.0603} & 29.786 & 1.3191 & \textbf{23.606} & \textbf{.8443} & \textbf{.1724} & 29.605 & \textbf{1.2913} & 18.776 & .6865 & .3482 & 29.165 & 1.1818 \\
\end{longtable}
\endgroup

%% file: tables/full_results_hunyuan_video.tex
\begingroup
\fontsize{9.5}{11}\selectfont
\setlength{\tabcolsep}{1pt}
\setlength{\aboverulesep}{0.5pt}
\setlength{\belowrulesep}{0.5pt}
\setlength{\LTpre}{2pt}
\renewcommand{\arraystretch}{0.95}
\setlength{\LTcapwidth}{\linewidth}
\begin{longtable}{@{}ll*{3}{rrrrr}@{}}
\caption{Complete HunyuanVideo results, averaged over three base seeds.
P is PSNR in dB, S is SSIM, and L is LPIPS. $\Delta_t$ is temporal LPIPS for the
cached video minus that for the full-compute video. Values closer to zero are
better. VB is the VBench score, reported only on VB944.
$\rho=K/50$ is the target cache ratio. Bold marks the best unrounded mean
within each dataset, ratio, and metric.}
\label{tab:hunyuan_video-full-results}\\
\toprule
& & \multicolumn{5}{c}{$\rho=.58$ ($K=29$)} & \multicolumn{5}{c}{$\rho=.74$ ($K=37$)} & \multicolumn{5}{c}{$\rho=.82$ ($K=41$)} \\
\cmidrule(lr){3-7}\cmidrule(lr){8-12}\cmidrule(l){13-17}
Data & Method & P$\uparrow$ & S$\uparrow$ & L$\downarrow$ & $\Delta_t$ & VB$\uparrow$ & P$\uparrow$ & S$\uparrow$ & L$\downarrow$ & $\Delta_t$ & VB$\uparrow$ & P$\uparrow$ & S$\uparrow$ & L$\downarrow$ & $\Delta_t$ & VB$\uparrow$ \\
\midrule
\endfirsthead
\caption[]{Continued.}\\
\toprule
& & \multicolumn{5}{c}{$\rho=.58$ ($K=29$)} & \multicolumn{5}{c}{$\rho=.74$ ($K=37$)} & \multicolumn{5}{c}{$\rho=.82$ ($K=41$)} \\
\cmidrule(lr){3-7}\cmidrule(lr){8-12}\cmidrule(l){13-17}
Data & Method & P$\uparrow$ & S$\uparrow$ & L$\downarrow$ & $\Delta_t$ & VB$\uparrow$ & P$\uparrow$ & S$\uparrow$ & L$\downarrow$ & $\Delta_t$ & VB$\uparrow$ & P$\uparrow$ & S$\uparrow$ & L$\downarrow$ & $\Delta_t$ & VB$\uparrow$ \\
\midrule
\endhead
\bottomrule
\endfoot
P599 & Sea & 29.36 & .8954 & .0725 & -.00141 & -- & 23.26 & .7756 & .1891 & -.00101 & -- & 19.40 & .6771 & .3171 & -.00495 & -- \\
P599 & Tea & 23.18 & .7800 & .1726 & +.00190 & -- & 19.18 & .6713 & .2952 & +.00531 & -- & 17.80 & .6217 & .3760 & +.00732 & -- \\
P599 & Sen & 30.60 & .9093 & .0630 & -.00208 & -- & 24.91 & .7908 & .1919 & -.00335 & -- & 21.65 & .7058 & .2877 & \textbf{+.00027} & -- \\
P599 & Di & 31.82 & .9241 & .0471 & -.00042 & -- & 22.24 & .6910 & .3632 & -.00650 & -- & 20.19 & .6631 & .3947 & -.00340 & -- \\
P599 & Bud & 30.92 & .9123 & .0658 & -.00281 & -- & 25.07 & .7837 & .2133 & -.00396 & -- & 22.13 & .7052 & .3253 & -.00386 & -- \\
P599 & Mean & \textbf{33.47} & \textbf{.9382} & \textbf{.0401} & -.00225 & -- & \textbf{27.58} & \textbf{.8544} & \textbf{.1421} & -.00574 & -- & \textbf{23.12} & \textbf{.7600} & \textbf{.2569} & -.00666 & -- \\
P599 & Tay O1 & 23.24 & .7880 & .1641 & \textbf{-.00019} & -- & 18.40 & .6611 & .2943 & \textbf{-.00099} & -- & 15.11 & .5508 & .4458 & -.00336 & -- \\
P599 & Hi O2 & 23.81 & .7947 & .1571 & +.00084 & -- & 20.05 & .6974 & .2566 & +.00227 & -- & 17.43 & .6129 & .3627 & +.00396 & -- \\
P599 & L2P & 23.95 & .7983 & .1567 & +.00107 & -- & 20.91 & .7220 & .2380 & +.00247 & -- & 19.33 & .6711 & .3081 & +.00455 & -- \\
\midrule
VB944 & Sea & 29.74 & .9012 & .0718 & -.00100 & .8077 & 23.17 & .7790 & .1926 & \textbf{-.00066} & .8051 & 19.52 & .6833 & .3178 & -.00264 & .7929 \\
VB944 & Tea & 22.84 & .7754 & .1874 & +.00131 & .8106 & 19.10 & .6689 & .3097 & +.00419 & .8064 & 17.82 & .6250 & .3884 & +.00583 & .7885 \\
VB944 & Sen & 31.25 & .9187 & .0585 & -.00139 & .8071 & 25.37 & .8093 & .1790 & -.00197 & .7929 & 21.64 & .7166 & .2889 & \textbf{+.00041} & .7830 \\
VB944 & Di & 32.54 & .9318 & .0445 & \textbf{-.00034} & .8084 & 22.63 & .7104 & .3581 & -.00318 & .7254 & 20.36 & .6780 & .4013 & -.00235 & .7158 \\
VB944 & Bud & 31.79 & .9238 & .0584 & -.00192 & .8078 & 25.55 & .8029 & .2011 & -.00228 & .7872 & 22.31 & .7216 & .3243 & -.00221 & .7542 \\
VB944 & Mean & \textbf{34.34} & \textbf{.9457} & \textbf{.0365} & -.00151 & .8069 & \textbf{28.09} & \textbf{.8669} & \textbf{.1336} & -.00364 & .7975 & \textbf{23.16} & \textbf{.7699} & \textbf{.2584} & -.00439 & .7803 \\
VB944 & Tay O1 & 23.09 & .7872 & .1741 & -.00050 & .8100 & 18.74 & .6704 & .2955 & -.00189 & .8018 & 16.02 & .5759 & .4192 & -.00402 & .7820 \\
VB944 & Hi O2 & 23.55 & .7927 & .1688 & +.00043 & \textbf{.8108} & 20.07 & .7017 & .2653 & +.00099 & .8032 & 17.91 & .6301 & .3594 & +.00108 & .7908 \\
VB944 & L2P & 23.59 & .7934 & .1710 & +.00085 & .8105 & 20.56 & .7163 & .2569 & +.00213 & \textbf{.8069} & 19.17 & .6701 & .3236 & +.00373 & \textbf{.7991} \\
\end{longtable}
\endgroup

%% file: tables/full_results_wan21.tex
\begingroup
\fontsize{9.5}{11}\selectfont
\setlength{\tabcolsep}{1pt}
\setlength{\aboverulesep}{0.5pt}
\setlength{\belowrulesep}{0.5pt}
\setlength{\LTpre}{2pt}
\renewcommand{\arraystretch}{0.95}
\setlength{\LTcapwidth}{\linewidth}
\begin{longtable}{@{}ll*{3}{rrrrr}@{}}
\caption{Complete Wan2.1 results, averaged over three base seeds.
P is PSNR in dB, S is SSIM, and L is LPIPS. $\Delta_t$ is temporal LPIPS for the
cached video minus that for the full-compute video. Values closer to zero are
better. VB is the VBench score, reported only on VB944.
$\rho=K/50$ is the target cache ratio. Bold marks the best unrounded mean
within each dataset, ratio, and metric.}
\label{tab:wan21-full-results}\\
\toprule
& & \multicolumn{5}{c}{$\rho=.58$ ($K=29$)} & \multicolumn{5}{c}{$\rho=.74$ ($K=37$)} & \multicolumn{5}{c}{$\rho=.82$ ($K=41$)} \\
\cmidrule(lr){3-7}\cmidrule(lr){8-12}\cmidrule(l){13-17}
Data & Method & P$\uparrow$ & S$\uparrow$ & L$\downarrow$ & $\Delta_t$ & VB$\uparrow$ & P$\uparrow$ & S$\uparrow$ & L$\downarrow$ & $\Delta_t$ & VB$\uparrow$ & P$\uparrow$ & S$\uparrow$ & L$\downarrow$ & $\Delta_t$ & VB$\uparrow$ \\
\midrule
\endfirsthead
\caption[]{Continued.}\\
\toprule
& & \multicolumn{5}{c}{$\rho=.58$ ($K=29$)} & \multicolumn{5}{c}{$\rho=.74$ ($K=37$)} & \multicolumn{5}{c}{$\rho=.82$ ($K=41$)} \\
\cmidrule(lr){3-7}\cmidrule(lr){8-12}\cmidrule(l){13-17}
Data & Method & P$\uparrow$ & S$\uparrow$ & L$\downarrow$ & $\Delta_t$ & VB$\uparrow$ & P$\uparrow$ & S$\uparrow$ & L$\downarrow$ & $\Delta_t$ & VB$\uparrow$ & P$\uparrow$ & S$\uparrow$ & L$\downarrow$ & $\Delta_t$ & VB$\uparrow$ \\
\midrule
\endhead
\bottomrule
\endfoot
P599 & Sea & 26.96 & .8732 & .0885 & -.00245 & -- & 21.72 & .7452 & .2041 & -.00094 & -- & 18.31 & .6408 & .3235 & +.00401 & -- \\
P599 & Tea & 23.32 & .8080 & .1376 & -.00082 & -- & 21.35 & .7480 & .2046 & -.00191 & -- & 18.76 & .6414 & .3516 & +.00318 & -- \\
P599 & Sen & 26.57 & .8733 & .0889 & -.00156 & -- & 22.42 & .7302 & .2262 & +.00173 & -- & 19.61 & .6382 & .3332 & +.01492 & -- \\
P599 & Di & 26.83 & .8824 & .0744 & -.00078 & -- & 20.43 & .7319 & .2129 & \textbf{-.00057} & -- & 18.44 & .6381 & .3612 & \textbf{+.00029} & -- \\
P599 & Bud & 24.23 & .8282 & .1216 & -.00182 & -- & \textbf{22.86} & .7673 & .1897 & -.00202 & -- & \textbf{20.36} & \textbf{.6742} & \textbf{.2956} & +.00479 & -- \\
P599 & Mean & \textbf{30.68} & \textbf{.9275} & \textbf{.0464} & -.00234 & -- & 22.77 & \textbf{.7887} & \textbf{.1601} & +.00293 & -- & 18.13 & .6391 & .3218 & +.01160 & -- \\
P599 & Tay O1 & 17.92 & .6571 & .2821 & +.00092 & -- & 13.27 & .4916 & .4962 & +.00120 & -- & 10.75 & .4038 & .6442 & +.00608 & -- \\
P599 & Hi O2 & 19.57 & .7039 & .2332 & \textbf{-.00038} & -- & 16.14 & .5835 & .3808 & -.00112 & -- & 13.43 & .4898 & .5353 & -.00306 & -- \\
P599 & L2P & 20.10 & .7213 & .2151 & +.00061 & -- & 17.81 & .6377 & .3024 & +.00927 & -- & 14.00 & .4642 & .5015 & +.06617 & -- \\
\midrule
VB944 & Sea & 27.19 & .8872 & .0745 & -.00085 & .7996 & 21.18 & .7511 & .1927 & +.00099 & .7952 & 17.53 & .6383 & .3221 & +.00595 & \textbf{.7816} \\
VB944 & Tea & 22.95 & .8201 & .1268 & -.00015 & .8042 & 20.94 & .7599 & .1906 & -.00038 & .7976 & 18.35 & .6508 & .3340 & +.00597 & .7533 \\
VB944 & Sen & 26.97 & .8922 & .0720 & -.00063 & .8007 & 22.42 & .7455 & .1981 & +.00432 & .7826 & 19.32 & .6506 & .3064 & +.01514 & .7484 \\
VB944 & Di & 27.24 & .9013 & .0592 & -.00042 & .8040 & 20.09 & .7483 & .1962 & \textbf{-.00001} & \textbf{.7986} & 18.07 & .6511 & .3370 & +.00297 & .7345 \\
VB944 & Bud & 23.90 & .8390 & .1112 & -.00066 & .8025 & 22.63 & .7796 & .1703 & +.00055 & .7949 & \textbf{19.97} & \textbf{.6836} & \textbf{.2736} & +.00751 & .7699 \\
VB944 & Mean & \textbf{31.33} & \textbf{.9399} & \textbf{.0362} & -.00136 & .8052 & \textbf{22.87} & \textbf{.8100} & \textbf{.1389} & +.00294 & .7869 & 17.82 & .6545 & .3033 & +.01056 & .7374 \\
VB944 & Tay O1 & 17.46 & .6708 & .2729 & +.00059 & .8051 & 13.41 & .5192 & .4744 & +.00309 & .7784 & 11.07 & .4300 & .6208 & +.01202 & .6899 \\
VB944 & Hi O2 & 18.84 & .7089 & .2311 & \textbf{-.00002} & \textbf{.8083} & 15.90 & .6020 & .3673 & +.00009 & .7976 & 13.57 & .5149 & .5135 & \textbf{+.00121} & .7474 \\
VB944 & L2P & 19.28 & .7218 & .2173 & +.00079 & .8068 & 17.25 & .6461 & .3002 & +.00741 & .7848 & 13.89 & .4849 & .4862 & +.05478 & .6661 \\
\end{longtable}
\endgroup

%% file: tables/actual_k_video.tex
\begin{table}[!htbp]
\caption{HunyuanVideo mean realized cache counts for the four adaptive methods.}
\label{tab:actual-k-hunyuan_video}
\centering
\small
\begin{tabular}{llrrr}
\toprule
Dataset & Method & $K_{\rm target}=29$ & $K_{\rm target}=37$ & $K_{\rm target}=41$ \\
\midrule
Penguin599 & SeaCache & 29.042 & 37.000 & 41.000 \\
Penguin599 & TeaCache & 28.053 & 36.534 & 41.019 \\
Penguin599 & SenCache & 28.959 & 36.000 & 40.853 \\
Penguin599 & DiCache & 28.710 & 37.000 & 41.000 \\
\midrule
VBench944 & SeaCache & 29.019 & 37.000 & 41.000 \\
VBench944 & TeaCache & 28.055 & 36.687 & 41.019 \\
VBench944 & SenCache & 28.965 & 36.000 & 40.884 \\
VBench944 & DiCache & 28.659 & 37.000 & 41.000 \\
\bottomrule
\end{tabular}
\end{table}

\begin{table}[!htbp]
\caption{Wan2.1 mean realized cache counts for the four adaptive methods.}
\label{tab:actual-k-wan21}
\centering
\small
\begin{tabular}{llrrr}
\toprule
Dataset & Method & $K_{\rm target}=29$ & $K_{\rm target}=37$ & $K_{\rm target}=41$ \\
\midrule
Penguin599 & SeaCache & 29.973 & 37.018 & 40.000 \\
Penguin599 & TeaCache & 29.000 & 36.000 & 41.000 \\
Penguin599 & SenCache & 28.975 & 37.000 & 41.010 \\
Penguin599 & DiCache & 28.954 & 36.900 & 41.000 \\
\midrule
VBench944 & SeaCache & 29.956 & 37.007 & 40.000 \\
VBench944 & TeaCache & 29.000 & 36.000 & 41.000 \\
VBench944 & SenCache & 28.871 & 36.999 & 41.004 \\
VBench944 & DiCache & 28.925 & 36.867 & 41.000 \\
\bottomrule
\end{tabular}
\end{table}

%% file: tables/compute_costs.tex
\begin{table}[!htbp]
\caption{Generation is faster with caching on FLUX.1-dev.  Columns specify the target cache count $K$.  Each entry gives the mean time in seconds per image, followed by the speedup relative to full-compute generation.  Full-compute generation takes 10.19 s per image.  We average 12 H100 measurements after excluding four initial samples.}
\label{tab:latency-flux}
\centering
\small
\begin{tabular}{lrrr}
\toprule
Method & $K=29$ & $K=37$ & $K=41$ \\
\midrule
SeaCache & 4.81 / 2.12$\times$ & 3.26 / 3.13$\times$ & 2.51 / 4.06$\times$ \\
TeaCache & 4.73 / 2.15$\times$ & 3.16 / 3.22$\times$ & 2.45 / 4.17$\times$ \\
SenCache & 4.74 / 2.15$\times$ & 3.19 / 3.20$\times$ & 2.45 / 4.15$\times$ \\
DiCache & 4.89 / 2.08$\times$ & 3.34 / 3.05$\times$ & 2.61 / 3.91$\times$ \\
TaylorSeer O1 & 5.42 / 1.88$\times$ & 4.00 / 2.55$\times$ & 3.31 / 3.08$\times$ \\
HiCache O2 & 5.86 / 1.74$\times$ & 4.55 / 2.24$\times$ & 3.88 / 2.63$\times$ \\
L2P & 4.68 / 2.18$\times$ & 3.17 / 3.21$\times$ & 2.40 / 4.24$\times$ \\
DPCache & 4.71 / 2.16$\times$ & 3.16 / 3.22$\times$ & 2.41 / 4.23$\times$ \\
BudCache & 4.69 / 2.17$\times$ & 3.15 / 3.23$\times$ & 2.37 / 4.29$\times$ \\
MeanCache & 4.69 / 2.17$\times$ & 3.15 / 3.23$\times$ & 2.40 / 4.25$\times$ \\
\bottomrule
\end{tabular}
\end{table}

\begin{table}[!htbp]
\caption{Generation is faster with caching on Qwen-Image.  Columns specify the target cache count $K$.  Each entry gives the mean time in seconds per image, followed by the speedup relative to full-compute generation.  Full-compute generation takes 33.45 s per image.  We average 12 H100 measurements after excluding four initial samples.}
\label{tab:latency-qwen}
\centering
\small
\begin{tabular}{lrrr}
\toprule
Method & $K=29$ & $K=37$ & $K=41$ \\
\midrule
SeaCache & 14.79 / 2.26$\times$ & 9.48 / 3.53$\times$ & 6.84 / 4.89$\times$ \\
TeaCache & 14.67 / 2.28$\times$ & 9.30 / 3.60$\times$ & 6.68 / 5.00$\times$ \\
SenCache & 14.58 / 2.29$\times$ & 9.36 / 3.57$\times$ & 6.66 / 5.02$\times$ \\
DiCache & 14.85 / 2.25$\times$ & 9.73 / 3.44$\times$ & 7.11 / 4.71$\times$ \\
TaylorSeer O1 & 17.62 / 1.90$\times$ & 12.95 / 2.58$\times$ & 10.59 / 3.16$\times$ \\
HiCache O2 & 19.76 / 1.69$\times$ & 15.40 / 2.17$\times$ & 13.13 / 2.55$\times$ \\
L2P & 14.48 / 2.31$\times$ & 9.20 / 3.64$\times$ & 6.61 / 5.06$\times$ \\
DPCache & 14.46 / 2.31$\times$ & 9.21 / 3.63$\times$ & 6.59 / 5.08$\times$ \\
BudCache & 14.59 / 2.29$\times$ & 9.33 / 3.58$\times$ & 6.69 / 5.00$\times$ \\
MeanCache & 14.52 / 2.30$\times$ & 9.23 / 3.63$\times$ & 6.63 / 5.05$\times$ \\
\bottomrule
\end{tabular}
\end{table}

\begin{table}[!htbp]
\caption{Generation is faster with caching on HunyuanVideo.  Columns specify the target cache count $K$.  Each entry gives the mean time in seconds per video, followed by the speedup relative to full-compute generation.  Full-compute generation takes 103.49 s per video.  H100 measurements include writing the video file.  We exclude each worker's first video.}
\label{tab:latency-hunyuan_video}
\centering
\small
\begin{tabular}{lrrr}
\toprule
Method & $K=29$ & $K=37$ & $K=41$ \\
\midrule
SeaCache & 47.64 / 2.17$\times$ & 32.24 / 3.21$\times$ & 24.50 / 4.22$\times$ \\
TeaCache & 49.26 / 2.10$\times$ & 32.99 / 3.14$\times$ & 24.20 / 4.28$\times$ \\
SenCache & 47.36 / 2.19$\times$ & 33.78 / 3.06$\times$ & 24.32 / 4.25$\times$ \\
DiCache & 48.97 / 2.11$\times$ & 33.17 / 3.12$\times$ & 25.72 / 4.02$\times$ \\
TaylorSeer O1 & 49.68 / 2.08$\times$ & 34.53 / 3.00$\times$ & 26.98 / 3.84$\times$ \\
HiCache O2 & 50.83 / 2.04$\times$ & 35.72 / 2.90$\times$ & 28.14 / 3.68$\times$ \\
L2P & 47.25 / 2.19$\times$ & 31.78 / 3.26$\times$ & 24.31 / 4.26$\times$ \\
BudCache & 47.26 / 2.19$\times$ & 31.86 / 3.25$\times$ & 24.09 / 4.30$\times$ \\
MeanCache & 47.25 / 2.19$\times$ & 31.75 / 3.26$\times$ & 24.02 / 4.31$\times$ \\
\bottomrule
\end{tabular}
\end{table}

\begin{table}[!htbp]
\caption{Generation is faster with caching on Wan2.1.  Columns specify the target cache count $K$.  Each entry gives the mean time in seconds per video, followed by the speedup relative to full-compute generation.  Full-compute generation takes 74.98 s per video.  H100 measurements include writing the video file.  We exclude each worker's first video.}
\label{tab:latency-wan21}
\centering
\small
\begin{tabular}{lrrr}
\toprule
Method & $K=29$ & $K=37$ & $K=41$ \\
\midrule
SeaCache & 32.66 / 2.30$\times$ & 22.64 / 3.31$\times$ & 18.42 / 4.07$\times$ \\
TeaCache & 33.74 / 2.22$\times$ & 23.77 / 3.15$\times$ & 16.75 / 4.48$\times$ \\
SenCache & 33.76 / 2.22$\times$ & 22.34 / 3.36$\times$ & 16.66 / 4.50$\times$ \\
DiCache & 35.38 / 2.12$\times$ & 24.48 / 3.06$\times$ & 18.91 / 3.96$\times$ \\
TaylorSeer O1 & 37.21 / 2.02$\times$ & 26.29 / 2.85$\times$ & 20.91 / 3.58$\times$ \\
HiCache O2 & 38.81 / 1.93$\times$ & 27.93 / 2.68$\times$ & 22.44 / 3.34$\times$ \\
L2P & 33.69 / 2.23$\times$ & 22.31 / 3.36$\times$ & 16.69 / 4.49$\times$ \\
BudCache & 33.71 / 2.22$\times$ & 22.42 / 3.34$\times$ & 16.75 / 4.48$\times$ \\
MeanCache & 33.67 / 2.23$\times$ & 22.28 / 3.37$\times$ & 16.66 / 4.50$\times$ \\
\bottomrule
\end{tabular}
\end{table}

%% file: tables/fixed_replay_image.tex
\begin{table}[!htbp]
\caption{\textbf{Replacing adaptive schedules with one fixed schedule changes
mean PSNR by $-0.81$ to $+1.56$ dB on new prompts where the schedules differ.}
We select the fixed schedule on 544 prompts and reuse it with the adaptive
method's approximation policy.  $m$ is the fraction of prompt--seed runs on
which the adaptive method already used the fixed schedule.
$n_{\rm off}$ counts runs on which the adaptive method follows another schedule.
For each run, we obtain $\Delta P$ by subtracting the adaptive method's output
PSNR from the fixed schedule's output PSNR on the same prompt and seed.
$\overline{\Delta P}$ is the mean of these differences over the $n_{\rm off}$ runs.
Each model--method--ratio combination uses
1,632 prompts under three base seeds, or 4,896 runs.
The held-out columns use the 1,088 prompts not used for schedule selection.
All PSNR differences are in dB.  Positive differences favor the fixed schedule.
The 95\% interval belongs to
$\overline{\Delta P}^{\rm held}$ and is given to three decimals.  It comes
from resampling prompts and keeping each prompt's runs together, using only
runs where the schedules differ.  Prompt means are weighted by these run counts.
The final column is the median $|\Delta P|$ over the $n_{\rm off}^{\rm held}$ runs.
Table~\ref{tab:gate-paths} in Appendix~\ref{app:schedule-distribution} reports a larger fraction for the same methods because
it averages over four image datasets and counts the most frequent schedule whatever
its cache count, while $m$ counts one schedule that caches exactly $K$ steps on
PartiPrompts.
The lower block replaces the fixed schedule with random schedules of the same
$K$. For each random schedule, we retain its available held-out runs where
the adaptive method's schedule differs from the selected fixed schedule.
Each retained run contributes one paired evaluation for that random schedule.
We subtract the adaptive method's output PSNR from each random schedule's
output PSNR.  The median and minimum-to-maximum range summarize these
differences across random schedules, models, and cache ratios.
}
\label{tab:fixed-replay-image-full}
\centering
\footnotesize
{\setlength{\tabcolsep}{2.5pt}
\begin{tabular}{lrrrrrrrr}
\toprule
Method & $K$ & $m$ (\%) & $n_{\rm off}$ & $\overline{\Delta P}$ &
$n_{\rm off}^{\rm held}$ & $\overline{\Delta P}^{\rm held}$ &
95\% interval & Med. $|\Delta P|^{\rm held}$ \\
\midrule
\multicolumn{9}{l}{\textbf{FLUX.1-dev}} \\
SeaCache & 29 & 85.7 & 699 & -0.07 & 461 & -0.07 & $[-0.104,\,-0.046]$ & 0.06 \\
TeaCache & 29 & 38.5 & 3,009 & +0.26 & 1,992 & +0.23 & $[0.142,\,0.329]$ & 0.02 \\
SenCache & 29 & 11.9 & 4,313 & +0.05 & 2,901 & +0.05 & $[0.031,\,0.067]$ & 0.13 \\
DiCache & 29 & 18.7 & 3,979 & -0.12 & 2,639 & -0.12 & $[-0.146,\,-0.088]$ & 0.13 \\
SeaCache & 37 & 68.7 & 1,534 & +0.01 & 1,057 & +0.01 & $[0.009,\,0.019]$ & 0.03 \\
TeaCache & 37 & 18.9 & 3,973 & +0.96 & 2,644 & +0.92 & $[0.825,\,1.028]$ & 0.07 \\
SenCache & 37 & 21.9 & 3,823 & -0.09 & 2,551 & -0.09 & $[-0.108,\,-0.080]$ & 0.09 \\
DiCache & 37 & 40.6 & 2,907 & -0.15 & 1,958 & -0.16 & $[-0.168,\,-0.144]$ & 0.13 \\
SeaCache & 41 & 40.0 & 2,938 & -0.03 & 1,915 & -0.03 & $[-0.031,\,-0.027]$ & 0.03 \\
TeaCache & 41 & 46.8 & 2,605 & +1.60 & 1,743 & +1.56 & $[1.446,\,1.682]$ & 1.53 \\
SenCache & 41 & 30.9 & 3,383 & -0.13 & 2,291 & -0.13 & $[-0.153,\,-0.108]$ & 0.06 \\
DiCache & 41 & 25.8 & 3,631 & -0.10 & 2,440 & -0.10 & $[-0.111,\,-0.091]$ & 0.08 \\
\midrule
\multicolumn{9}{l}{\textbf{Qwen-Image}} \\
SeaCache & 29 & 75.2 & 1,213 & -0.06 & 779 & -0.06 & $[-0.073,\,-0.051]$ & 0.05 \\
TeaCache & 29 & 80.5 & 954 & 0.00 & 651 & -0.01 & $[-0.007,\,-0.003]$ & 0.00 \\
SenCache & 29 & 83.8 & 795 & +0.08 & 543 & +0.08 & $[0.039,\,0.125]$ & 0.12 \\
DiCache & 29 & 41.2 & 2,877 & -0.01 & 1,937 & -0.02 & $[-0.038,\,0.002]$ & 0.09 \\
SeaCache & 37 & 98.9 & 54 & -0.03 & 34 & -0.03 & $[-0.105,\,0.025]$ & 0.03 \\
TeaCache & 37 & 28.6 & 3,497 & -0.79 & 2,320 & -0.81 & $[-0.858,\,-0.763]$ & 0.55 \\
SenCache & 37 & 85.5 & 711 & -0.26 & 460 & -0.26 & $[-0.287,\,-0.231]$ & 0.18 \\
DiCache & 37 & 47.4 & 2,573 & +0.11 & 1,737 & +0.11 & $[0.087,\,0.127]$ & 0.22 \\
SeaCache & 41 & 77.0 & 1,127 & +0.02 & 739 & +0.02 & $[0.013,\,0.021]$ & 0.02 \\
TeaCache & 41 & 57.8 & 2,064 & -0.07 & 1,365 & -0.08 & $[-0.106,\,-0.049]$ & 0.01 \\
SenCache & 41 & 66.2 & 1,657 & +0.29 & 1,096 & +0.30 & $[0.276,\,0.323]$ & 0.28 \\
DiCache & 41 & 64.7 & 1,726 & +0.18 & 1,173 & +0.18 & $[0.164,\,0.195]$ & 0.19 \\
\bottomrule
\end{tabular}
}

\vspace{0.7em}

{\setlength{\tabcolsep}{4pt}
\begin{tabular}{lrrr}
\toprule
\multicolumn{4}{@{}l}{Random exact-$K$ control on the same evaluation subset} \\
Method & Paired evaluations & Median $\Delta P$, dB & Min--max $\Delta P$, dB \\
\midrule
SeaCache & 12,319 & $-2.605$ & $-26.675$ to $+16.731$ \\
SenCache & 26,194 & $-4.232$ & $-28.694$ to $+9.086$ \\
TeaCache & 29,032 & $+0.632$ & $-14.963$ to $+24.957$ \\
DiCache & 32,036 & $-4.411$ & $-32.094$ to $+13.442$ \\
\bottomrule
\end{tabular}
}
\end{table}

%% file: tables/fixed_replay_image_seeds.tex
\begin{table}[!htbp]
\caption{\textbf{Replacing adaptive schedules with one fixed schedule changes
mean PSNR by $-0.83$ to $+1.60$ dB across base seeds on new prompts.}
We use only runs where the adaptive method chooses a different schedule.
For each run, we subtract the adaptive method's output PSNR from the fixed
schedule's output PSNR.  Each entry gives the number of these runs, the mean
PSNR difference, and the median absolute PSNR difference, in that order.
Both differences are in dB.
Each base-seed assignment uses 1,088 PartiPrompts.
Base seeds A, B, and C are 41, 42, and 43 for FLUX.1-dev and
42, 100042, and 200042 for Qwen-Image.}
\label{tab:fixed-replay-image-seeds}
\centering
\footnotesize
{\setlength{\tabcolsep}{4pt}
\begin{tabular}{llrccc}
\toprule
Model & Method & $K$ & Base seed A & Base seed B & Base seed C \\
\midrule
FLUX & SeaCache & 29 & 157/$-0.08$/0.06 & 157/$-0.10$/0.06 & 147/$-0.04$/0.06 \\
FLUX & TeaCache & 29 & 675/$+0.26$/0.02 & 654/$+0.19$/0.02 & 663/$+0.25$/0.02 \\
FLUX & SenCache & 29 & 967/$+0.05$/0.13 & 975/$+0.04$/0.13 & 959/$+0.06$/0.13 \\
FLUX & DiCache & 29 & 868/$-0.11$/0.12 & 886/$-0.13$/0.14 & 885/$-0.11$/0.13 \\
FLUX & SeaCache & 37 & 353/$+0.01$/0.03 & 356/$+0.02$/0.03 & 348/$+0.01$/0.03 \\
FLUX & TeaCache & 37 & 873/$+0.91$/0.07 & 884/$+0.93$/0.08 & 887/$+0.93$/0.06 \\
FLUX & SenCache & 37 & 845/$-0.10$/0.10 & 856/$-0.09$/0.09 & 850/$-0.09$/0.09 \\
FLUX & DiCache & 37 & 642/$-0.16$/0.13 & 661/$-0.15$/0.13 & 655/$-0.16$/0.13 \\
FLUX & SeaCache & 41 & 647/$-0.03$/0.03 & 622/$-0.03$/0.03 & 646/$-0.03$/0.03 \\
FLUX & TeaCache & 41 & 581/$+1.60$/1.56 & 579/$+1.56$/1.51 & 583/$+1.53$/1.53 \\
FLUX & SenCache & 41 & 754/$-0.13$/0.06 & 765/$-0.13$/0.06 & 772/$-0.13$/0.06 \\
FLUX & DiCache & 41 & 802/$-0.10$/0.08 & 810/$-0.10$/0.08 & 828/$-0.10$/0.08 \\
\midrule
Qwen & SeaCache & 29 & 260/$-0.05$/0.05 & 259/$-0.07$/0.04 & 260/$-0.07$/0.05 \\
Qwen & TeaCache & 29 & 216/$-0.01$/0.00 & 219/$0.00$/0.00 & 216/$0.00$/0.00 \\
Qwen & SenCache & 29 & 183/$+0.07$/0.13 & 183/$+0.04$/0.09 & 177/$+0.13$/0.14 \\
Qwen & DiCache & 29 & 631/$-0.02$/0.09 & 655/$0.00$/0.09 & 651/$-0.03$/0.08 \\
Qwen & SeaCache & 37 & 10/$-0.03$/0.04 & 15/$-0.04$/0.03 & 9/$-0.03$/0.01 \\
Qwen & TeaCache & 37 & 780/$-0.83$/0.56 & 769/$-0.77$/0.49 & 771/$-0.83$/0.59 \\
Qwen & SenCache & 37 & 156/$-0.26$/0.18 & 157/$-0.29$/0.20 & 147/$-0.23$/0.17 \\
Qwen & DiCache & 37 & 578/$+0.11$/0.22 & 568/$+0.10$/0.22 & 591/$+0.11$/0.21 \\
Qwen & SeaCache & 41 & 251/$+0.02$/0.02 & 246/$+0.02$/0.02 & 242/$+0.02$/0.02 \\
Qwen & TeaCache & 41 & 457/$-0.07$/0.01 & 446/$-0.08$/0.01 & 462/$-0.08$/0.01 \\
Qwen & SenCache & 41 & 363/$+0.29$/0.26 & 369/$+0.30$/0.28 & 364/$+0.31$/0.29 \\
Qwen & DiCache & 41 & 383/$+0.18$/0.19 & 376/$+0.17$/0.19 & 414/$+0.19$/0.20 \\
\bottomrule
\end{tabular}
}
\end{table}

%% file: tables/fixed_replay_video.tex
\begin{table}[!htbp]
\caption{\textbf{Reusing a fixed schedule loses at most 0.25 dB in mean PSNR
relative to adaptive scheduling in 17 of 19 comparisons.}
In these comparisons, the fixed schedule has the requested cache count and
differs from the adaptive schedule on at least one prompt. Each row uses 300 prompts,
split evenly between Penguin599 and VBench944.
$m$ is the fraction of prompts on which the adaptive
method already used the fixed schedule.
$n_{\rm off}$ counts prompts where it chooses a different schedule.
For each such prompt, we subtract the adaptive method's output PSNR from the
fixed schedule's output PSNR, using the same seed.
$\overline{\Delta P}_{\rm off}$ is the mean of these differences in dB.
Positive values favor the fixed schedule.
Target $K$ is the requested cached-step count.  Actual $K$ is the count in the
fixed schedule.  The lower block contains schedules whose counts differ from
the target.  Mean differences are not applicable when $n_{\rm off}=0$.}
\label{tab:fixed-replay-video-full}
\centering
\scriptsize
\renewcommand{\arraystretch}{0.90}
\begin{tabular}{llrrrrrc}
\toprule
Model & Method & Target $K$ & Actual $K$ & $m$ (\%) & $n_{\rm off}$ &
$\overline{\Delta P}_{\rm off}$ & \shortstack{Mean loss\\$\leq0.25$ dB} \\
\midrule
HunyuanVideo & SeaCache & 29 & 29 & 55 & 134 & +0.03 & yes \\
HunyuanVideo & TeaCache & 29 & 29 & 2 & 294 & +0.03 & yes \\
HunyuanVideo & SenCache & 29 & 29 & 26 & 223 & -0.54 & no \\
HunyuanVideo & DiCache & 29 & 29 & 36 & 192 & +0.05 & yes \\
HunyuanVideo & SeaCache & 37 & 37 & 76 & 71 & +0.02 & yes \\
HunyuanVideo & TeaCache & 37 & 37 & 51 & 146 & -0.03 & yes \\
HunyuanVideo & DiCache & 37 & 37 & 58 & 125 & +0.21 & yes \\
HunyuanVideo & SeaCache & 41 & 41 & 30 & 211 & +0.01 & yes \\
HunyuanVideo & TeaCache & 41 & 41 & 66 & 101 & +0.15 & yes \\
HunyuanVideo & SenCache & 41 & 41 & 85 & 46 & -0.70 & no \\
HunyuanVideo & DiCache & 41 & 41 & 31 & 206 & -0.07 & yes \\
\midrule
Wan2.1 & SeaCache & 29 & 29 & 4 & 288 & +0.37 & yes \\
Wan2.1 & TeaCache & 29 & 29 & 100 & 0 & n/a & n/a \\
Wan2.1 & SenCache & 29 & 29 & 35 & 196 & -0.20 & yes \\
Wan2.1 & DiCache & 29 & 29 & 79 & 63 & -0.22 & yes \\
Wan2.1 & SeaCache & 37 & 37 & 67 & 98 & +0.08 & yes \\
Wan2.1 & SenCache & 37 & 37 & 36 & 192 & -0.01 & yes \\
Wan2.1 & DiCache & 37 & 37 & 30 & 209 & -0.23 & yes \\
Wan2.1 & TeaCache & 41 & 41 & 100 & 0 & n/a & n/a \\
Wan2.1 & SenCache & 41 & 41 & 51 & 146 & +0.08 & yes \\
Wan2.1 & DiCache & 41 & 41 & 37 & 189 & -0.03 & yes \\
\midrule
HunyuanVideo & SenCache & 37 & 36 & 64 & 107 & +0.10 & yes \\
Wan2.1 & TeaCache & 37 & 36 & 100 & 0 & n/a & n/a \\
Wan2.1 & SeaCache & 41 & 40 & 80 & 61 & +0.04 & yes \\
\bottomrule
\end{tabular}
\end{table}

%% file: tables/fixed_replay_video_datasets.tex
\begin{table}[!htbp]
\caption{\textbf{Replacing adaptive schedules with a fixed schedule causes the
largest mean PSNR losses for SenCache on HunyuanVideo in both video datasets.}
We use only prompts where the schedules differ. For each such prompt, we
subtract the adaptive method's output PSNR from the fixed schedule's output
PSNR, using the same seed.  Each entry gives the number of these prompts,
the mean PSNR difference, and the median absolute PSNR difference, in that order.
Both differences are in dB.  Each dataset uses 150 prompts.
Penguin599 uses base seed 54 and VBench944 uses base seed 42.
Target $K$ is the requested cached-step count.  Actual $K$ is the count in the
fixed schedule.  The lower block contains schedules whose counts differ from
the target.  A zero count means that the adaptive method uses the fixed
schedule on every prompt, so the mean and median are not applicable.}
\label{tab:fixed-replay-video-datasets}
\centering
\scriptsize
\renewcommand{\arraystretch}{0.90}
{\setlength{\tabcolsep}{4pt}
\begin{tabular}{llrrcc}
\toprule
Model & Method & Target $K$ & Actual $K$ & Penguin599 & VBench944 \\
\midrule
HunyuanVideo & SeaCache & 29 & 29 & 64/$+0.02$/0.04 & 70/$+0.04$/0.03 \\
HunyuanVideo & TeaCache & 29 & 29 & 146/$0.00$/0.01 & 148/$+0.06$/0.01 \\
HunyuanVideo & SenCache & 29 & 29 & 113/$-0.53$/0.51 & 110/$-0.55$/0.49 \\
HunyuanVideo & DiCache & 29 & 29 & 89/$+0.05$/0.20 & 103/$+0.05$/0.16 \\
HunyuanVideo & SeaCache & 37 & 37 & 37/$+0.02$/0.02 & 34/$+0.02$/0.02 \\
HunyuanVideo & TeaCache & 37 & 37 & 83/$-0.03$/0.02 & 63/$-0.03$/0.02 \\
HunyuanVideo & DiCache & 37 & 37 & 52/$+0.22$/0.20 & 73/$+0.20$/0.19 \\
HunyuanVideo & SeaCache & 41 & 41 & 103/$+0.01$/0.02 & 108/$+0.01$/0.02 \\
HunyuanVideo & TeaCache & 41 & 41 & 48/$+0.18$/0.15 & 53/$+0.12$/0.08 \\
HunyuanVideo & SenCache & 41 & 41 & 27/$-0.75$/0.71 & 19/$-0.64$/0.50 \\
HunyuanVideo & DiCache & 41 & 41 & 112/$-0.05$/0.08 & 94/$-0.08$/0.07 \\
\midrule
Wan2.1 & SeaCache & 29 & 29 & 144/$+0.37$/0.25 & 144/$+0.37$/0.24 \\
Wan2.1 & TeaCache & 29 & 29 & 0 / n/a / n/a & 0 / n/a / n/a \\
Wan2.1 & SenCache & 29 & 29 & 100/$-0.22$/0.14 & 96/$-0.19$/0.20 \\
Wan2.1 & DiCache & 29 & 29 & 33/$-0.20$/0.08 & 30/$-0.24$/0.14 \\
Wan2.1 & SeaCache & 37 & 37 & 46/$+0.16$/0.01 & 52/$+0.01$/0.01 \\
Wan2.1 & SenCache & 37 & 37 & 100/$-0.01$/0.07 & 92/$-0.00$/0.07 \\
Wan2.1 & DiCache & 37 & 37 & 100/$-0.24$/0.20 & 109/$-0.23$/0.20 \\
Wan2.1 & TeaCache & 41 & 41 & 0 / n/a / n/a & 0 / n/a / n/a \\
Wan2.1 & SenCache & 41 & 41 & 76/$+0.08$/0.18 & 70/$+0.07$/0.15 \\
Wan2.1 & DiCache & 41 & 41 & 92/$-0.03$/0.05 & 97/$-0.03$/0.05 \\
\midrule
HunyuanVideo & SenCache & 37 & 36 & 47/$+0.08$/0.15 & 60/$+0.12$/0.14 \\
Wan2.1 & TeaCache & 37 & 36 & 0 / n/a / n/a & 0 / n/a / n/a \\
Wan2.1 & SeaCache & 41 & 40 & 32/$+0.05$/0.03 & 29/$+0.03$/0.01 \\
\bottomrule
\end{tabular}
}
\end{table}

%% file: tables/coverage_common_seed.tex
\begin{table}[!htbp]
\caption{\textbf{One fixed schedule stays within 0.5 dB of the per-prompt best
on most prompts in every enlarged image candidate set.}
Each candidate set uses the same 1,088 prompts at base seed 42 and residual
reuse for every schedule.  ``Schedules'' gives the number of candidates.
Within each candidate set, we select the schedule with the highest coverage
at 0.25 dB under this common seed and report it at all three margins.
Entries give the fraction of prompts within each PSNR margin of the
per-prompt best.  Brackets give the Bonferroni-adjusted one-sided 95\% lower bound.}
\label{tab:coverage-common-seed}
\centering
\small
\setlength{\tabcolsep}{3pt}
\begin{tabular}{llrlrrr}
\toprule
Model & Cache ratio & Schedules & Fixed schedule & 0.25 dB & 0.5 dB & 1.0 dB \\
\midrule
FLUX.1-dev & 0.58 & 32 & MeanCache & .616 [.571] & .737 [.696] & .865 [.832] \\
FLUX.1-dev & 0.74 & 31 & MeanCache & .500 [.455] & .670 [.627] & .817 [.780] \\
FLUX.1-dev & 0.82 & 15 & BudCache & .556 [.515] & .671 [.631] & .806 [.772] \\
Qwen-Image & 0.58 & 29 & MeanCache & .449 [.405] & .559 [.514] & .703 [.661] \\
Qwen-Image & 0.74 & 30 & MeanCache & .438 [.393] & .549 [.504] & .727 [.686] \\
Qwen-Image & 0.82 & 13 & MeanCache & .589 [.549] & .667 [.628] & .800 [.765] \\
\bottomrule
\end{tabular}
\end{table}

%% file: mechanism_appendix.tex
\section{Additional Cache-Error Analysis}
\label{app:mechanism}

\subsection{State and current contributions}
\label{app:closed-loop}

Let $z_n^F$ and $\widetilde z_n$ be the full-compute and cached latent states
at step $n$.
The full model's velocity prediction is $v_\theta(z,\tau_n,c)$, where $z$
is the latent state, $\tau_n$ is the noise level, and $c$ is the prompt conditioning.
Write the sampler update as $z_{n+1}=\Psi_n(z_n,v_n)$.

Equation~\ref{eq:action-state} follows by adding and subtracting the full model's
velocity prediction at the cached state:
\begin{align}
  \bar v_n-v_n^F
  &=\bar v_n-v_\theta(\widetilde z_n,\tau_n,c)
  +v_\theta(\widetilde z_n,\tau_n,c)
  -v_\theta(z_n^F,\tau_n,c)\\
  &=a_n+g_n.
\end{align}
For a one-step sampler, the latent-state error
$e_n=\widetilde z_n-z_n^F$ evolves as
\begin{equation}
  e_{n+1}=\Psi_n(\widetilde z_n,\bar v_n)
  -\Psi_n(z_n^F,v_n^F).
\end{equation}
For an Euler sampler with step matrix $H_n$,
\begin{equation}
  e_{n+1}=e_n+H_n(a_n+g_n).
  \label{eq:closed-loop-euler}
\end{equation}
At a full step, $a_n=0$ and the update becomes
$e_{n+1}=e_n+H_ng_n$.  The incoming state error remains part of the update.
Its direction relative to $H_ng_n$ determines whether the next step reduces
or enlarges the gap.

We measure $a_n$ and $g_n$ on FLUX.1-dev with SeaCache and TeaCache for 100
PartiPrompts, using seed $42+i$ for prompt index $i$ and 50 Euler steps.
SeaCache thresholds 0.29 and 1.0 yield mean cache ratios 0.580 and 0.820.
TeaCache thresholds 0.38 and 1.2 yield 0.581 and 0.820.
At cached steps, the full model is evaluated at the cached state
for measurement, without feeding that output back into generation.  Define
\begin{equation}
  E_{\mathrm{current}}=\sum_n\norm{a_n}_2,
  \qquad
  E_{\mathrm{state}}=\sum_n\norm{g_n}_2.
\end{equation}
The output error magnitude is $\norm{e_{50}}_2=
\norm{\widetilde z_{50}-z_{50}^F}_2$.
For SeaCache at mean cache ratios 0.580 and 0.820, the Spearman correlations
between output-error magnitude and $E_{\mathrm{state}}$ are 0.988 and 0.991,
compared with 0.748 and 0.742
for $E_{\mathrm{current}}$.
For TeaCache at mean cache ratios 0.581 and 0.820, the correlations are
0.994 and 0.996 for $E_{\mathrm{state}}$, compared with 0.681 and 0.579
for $E_{\mathrm{current}}$.
The accumulated state contribution is more strongly correlated
with output error magnitude than the accumulated current contribution.

\subsection{Complete cached trajectories}
\label{app:cached-trajectory}

We compare differences between cached and full-compute latent states with
final image quality across schedules and approximation policies.
The image analysis uses 256 combinations of model, cache ratio, schedule,
and approximation policy. Each combination uses
49 PartiPrompts with one seed per prompt, $42+i$ for prompt index $i$, and the
matching full-compute output.  Retaining all 51 latent states gives 12,544
paired trajectories.  For state $n$, the normalized latent-state gap is
\begin{equation}
  D_n=\frac{\norm{\widetilde z_n-z_n^F}_2}
  {\norm{z_{50}^F-z_0^F}_2}.
  \label{eq:trajectory-gap}
\end{equation}
The states agree before the first cached step.  The gap then grows along the
cached trajectory. For each combination, we take the median $D_{50}$ over
the 49 prompts with stored trajectories and mean PSNR over all 1,632 PartiPrompts at base seed 42.
We compare these two summaries across combinations within each model and
cache ratio.  The Spearman correlations range from $-0.98$ to $-0.95$
across the six model--ratio combinations.

We next compare early state differences with the position of the first cached
step as predictors of final quality.
For each combination, we use median $D_{10}$ over the 49 prompts with stored trajectories
and the same 1,632-prompt mean PSNR.
We retain the 218 combinations whose first cached step occurs before state 10,
so that caching has begun when the early gap is measured.
Within each model and cache count $K$, we
standardize the first cached step, median $D_{10}$, and mean PSNR to zero mean
and unit variance, then pool the standardized values.  Ordinary least-squares
regression with an intercept gives $R^2=0.268$ for the first cached step alone
and $R^2=0.715$ for median $D_{10}$ alone.  Using both predictors also gives
$R^2=0.715$.
The early gap accounts for more of the variation in PSNR than the first
cached-step position alone.

To measure this association using prompts outside schedule selection, we
recompute median $D_{10}$ over the 32 prompts with stored trajectories outside the
544-prompt selection split.
The other 17 prompts with stored trajectories belong to that split.
We compare the recomputed median with the same 1,632-prompt mean PSNR.
This rank-correlation analysis standardizes values within each model and $K$ over all combinations
before retaining the 218 combinations that begin caching before state 10. Their pooled Spearman
correlation is $-0.82$.

\begin{figure}[!htbp]
  \centering
  \includegraphics[width=\linewidth]{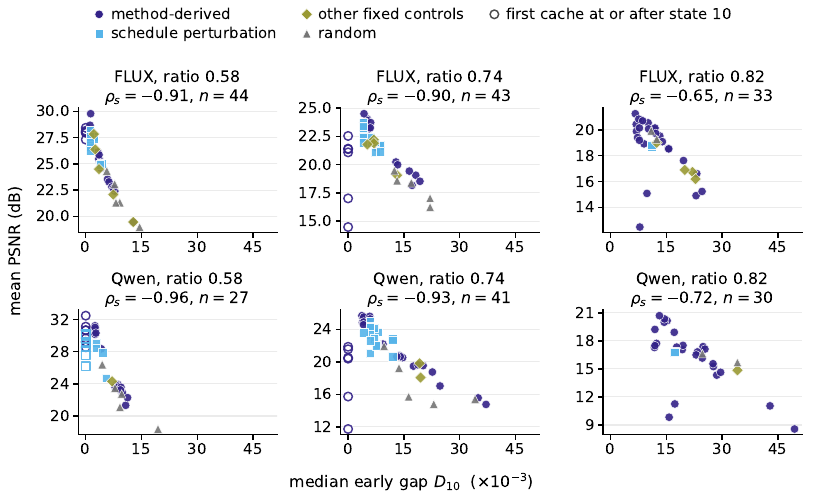}
  \caption{\textbf{Larger latent-state gaps at state 10 are associated with lower final-image PSNR.}
  Each point represents a fixed schedule and approximation policy.
  $D_{10}$ is the L2 distance between the cached and full-compute states at
  state 10, divided by the full-compute trajectory's chord length.
  The horizontal axis gives median $D_{10}$ over 49 prompts.
  The vertical axis gives mean PSNR in dB over 1,632 PartiPrompts.
  Colors and marker shapes distinguish schedules from caching methods,
  modified schedules, other fixed controls, and random schedules.
  Open markers indicate schedules whose first cached step is at or after
  step 10, so $D_{10}=0$.
  Each panel reports the Spearman correlation $\rho_s$ and number of
  combinations $n$ for filled markers.}
  \label{fig:early-drift}
\end{figure}

Figure~\ref{fig:early-drift} groups schedules by their source.  Method-derived
schedules are the fixed schedules and most frequent adaptive schedules used
in the experiment varying schedules and approximation policies.  Schedule perturbations swap cached and
full steps while preserving $K$.  Other fixed controls use trajectory geometry
or less frequent adaptive schedules.  Random schedules are sampled at the
same $K$.

The geometry-based controls use derivatives of the mean normalized off-chord
distance and speed along full-compute trajectories.
Speed is normalized step displacement per unit change in noise level.
These derivatives define costs for cached intervals, with larger costs
for longer noise-level gaps since the last full step.
Dynamic programming selects a schedule with the lowest total cost for
each model and cache ratio.
The FLUX.1-dev controls also include six schedules selected using
preliminary output quality. These variants change the derivative order
and how strongly the interval cost increases with the gap.

The video analysis retains 50 paired trajectories for each of 27 method and
cache-ratio combinations in each model.  The cached and full-compute trajectories
again agree before the first cached step and separate afterwards.

\subsection{Derivation and measurement of the four factors}
\label{app:pathwise-derivation}

Let $d_k^F$ denote the full-compute feature at step $k$ and let
$\widehat d_k=d_k^F+r_k$ be its approximation when only step $k$ is cached.
Let $G_k$ map this feature to the velocity prediction used by the sampler.  A
first-order expansion at $d_k^F$ gives
\begin{equation}
  G_k(d_k^F+r_k)-G_k(d_k^F)
  =B_k r_k+\mathcal O(\norm{r_k}^2),
  \qquad
  B_k=\left.\frac{\partial G_k}{\partial d}\right|_{d_k^F}.
\end{equation}
When only step $k$ is cached, it begins at the full-compute state.
The error in the next state is therefore
\begin{equation}
  e_{k+1}=H_kB_kr_k+\mathcal O(\norm{r_k}^2),
  \qquad
  H_k=\left.\frac{\partial\Psi_k}{\partial v}
  \right|_{(z_k^F,v_k^F)}.
\end{equation}
For each later full-compute step $j$, define
$T_j(z)=\Psi_j(z,v_\theta(z,\tau_j,c))$ and
$J_j=\left.\partial T_j/\partial z\right|_{z_j^F}$.  Propagating the
perturbation through the remaining transitions yields
\begin{equation}
  e_{50}^{(k)}\approx J_{49}\cdots J_{k+1}H_kB_kr_k
  =\Phi_{50,k+1}H_kB_kr_k.
\end{equation}
Taking the Euclidean norm of the error and operator norms of the linear maps
gives the scalar bound
\begin{equation}
  \norm{e_{50}^{(k)}}\lesssim
  \norm{\Phi_{50,k+1}}\,\norm{H_k}\,\norm{B_k}\,\norm{r_k}.
\end{equation}

The experiment uses FLUX.1-dev, 100 DrawBench prompts, 50 Euler steps,
and seed $42+i$ for prompt index $i$. We test steps 1--49 separately,
caching only the tested step under residual reuse, first-order Taylor
prediction, or second-order Hermite prediction. At early steps, predictions use the highest
order supported by the available feature history.
For the Euler update, write $\Delta\tau_k=\tau_{k+1}-\tau_k$, using the
sampler's actual noise levels.
We measure the velocity error and the resulting next-state error:
\begin{align}
  \delta v_k &=G_k(d_k^F+r_k)-G_k(d_k^F),\\
  \eta_k&=\Delta\tau_k\,\delta v_k.
\end{align}
We resume full computation from $z_{k+1}^F+\eta_k$ and denote its final state by $z_{50}'$.
The measured output error is $e_{50}^{(k)}=z_{50}'-z_{50}^F$.
The ratios $\norm{\delta v_k}_2/\norm{r_k}_2$ and
$\norm{e_{50}^{(k)}}_2/\norm{\eta_k}_2$ measure the model's response to the
feature error and the remaining trajectory's response to the next-state error.
These are measured gains along the finite perturbations used in the experiment.
For an evaluation prompt $i$, let $\operatorname{mean}_{-i}$ denote the
arithmetic mean over the other 99 prompts at the same step and with the
same approximation policy.
We estimate the propagation gain by averaging the two ratios separately:
\begin{equation}
  h_k^{\mathrm{LOO}}=
  \operatorname{mean}_{-i}\!\left[
    \frac{\norm{\delta v_k}_2}{\norm{r_k}_2+\epsilon}
  \right]
  |\Delta\tau_k|
  \operatorname{mean}_{-i}\!\left[
    \frac{\norm{e_{50}^{(k)}}_2}{\norm{\eta_k}_2+\epsilon}
  \right],
  \qquad \epsilon=10^{-12}.
  \label{eq:empirical-propagation-gain}
\end{equation}
The score for prompt $i$ is its feature-error norm multiplied by
$h_k^{\mathrm{LOO}}$.
Step 0 has no stored residual and step 49 has no later
transition for the amplification measurement.  The common comparison therefore uses
steps 1--48, giving 4,800 prompt--step observations per approximation policy.

We pool all 4,800 prompt--step observations within each approximation policy
and compute Spearman correlations with output-error magnitude.
For residual reuse, the complete product has a correlation of 0.886,
compared with 0.528 for local feature error $\norm{r_k}$ and 0.867 for the negated
step index $-k$.
For second-order Hermite prediction, the correlation is 0.896 for the complete
product, 0.586 for $\norm{r_k}$, and 0.880 for $-k$.
For first-order Taylor prediction, it is 0.899 for the complete product,
0.676 for $\norm{r_k}$, and 0.880 for $-k$.
The sampler step size $|\Delta\tau_k|$ and the measured future gain change in opposite
directions over denoising steps.
The feature-error norm multiplied by the step size, and the next-state
error estimate defined in Table~\ref{tab:four-factor-ladder-full}, are
negatively correlated with output-error magnitude.
Including the future gain in $\norm{r_k}h_k^{\mathrm{LOO}}$ gives a positive correlation.

\input{\arxivroot tables/four_factor_ladder_full}

We also split the 100 prompts into 50 even-indexed and 50 odd-indexed prompts.
For each approximation policy, we average the two gain ratios on one half,
form the propagation gain, and evaluate its product with $\norm{r_k}$
on the other half's 2,400 prompt--step observations.  We test both
transfer directions. Using the profile estimated from the other half changes
the correlation between this product and output-error magnitude by less than 0.003,
relative to using a profile estimated from the evaluation half.
Across approximation policies, the Spearman correlation between the two
halves' propagation profiles ranges from 0.9955 to 0.9981.
Thus, the propagation profile transfers across these prompt subsets.
The norm $\norm{r_k}$ captures differences in feature error between prompts at
the same step.

\subsection{Schedule and approximation policy comparisons}
\label{app:spx-details}

We distinguish the effects of schedule selection and feature approximation
through controlled comparisons.

\paragraph{Joint effects of schedule and approximation policy.}
The image experiment uses four schedules: BudCache, DPCache,
the uniform schedule shared by the fixed predictors, and DiCache's most
frequent schedule selected on the 544-prompt selection split.
Each is evaluated with residual reuse, first-order Taylor prediction,
second-order Hermite prediction, MeanCache's interval-average velocity
prediction, and two-anchor prediction in every
model--ratio combination, using 1,632 PartiPrompts under three base seeds.
For each of the 20 combinations, we average PSNR over prompts within each
seed, then average the three seed means. An additive
least-squares fit with equal weight for all 20 means separates the schedule
and approximation policy effects, with each set of effects constrained to sum to zero.
The interaction is the residual from this fit.  Its variance share is the sum
of squared residuals divided by the total sum of squared deviations of the
20 means from their grand mean.  Schedule and approximation policy shares use the
corresponding fitted effects and the same denominator.
Table~\ref{tab:spx-image-decomposition} reports the separate and joint effects.

\begin{figure}[!htbp]
  \centering
  \includegraphics[width=\linewidth]{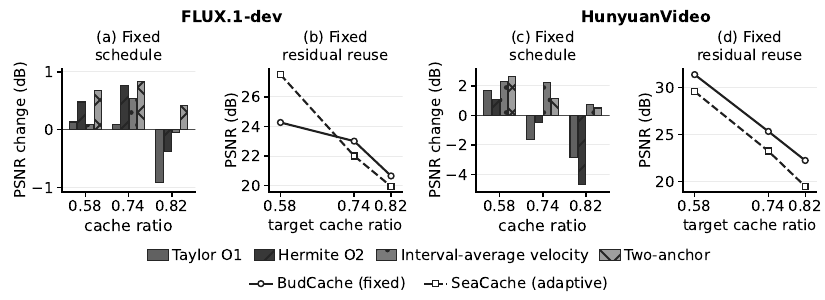}
  \caption{\textbf{Both the schedule and the approximation policy affect quality.}
  (a, c) The BudCache schedule is fixed at each cache ratio.
  For each approximation policy, we subtract the mean PSNR obtained with
  residual reuse from the mean PSNR obtained with that policy on the same
  prompts and seeds. Bars show these differences.
  Zero denotes residual reuse. Means use 1,632 PartiPrompts under three seeds
  for FLUX and 150 videos from each of Penguin599 and VBench944 for HunyuanVideo,
  with one seed per video prompt.
  (b, d) Fixed residual-reuse approximation policy. Circles show BudCache and squares show
  SeaCache. Means give equal weight to four image datasets or two video datasets.
  These axes use target ratios because SeaCache's cache counts vary.
  Vertical scales differ.}
  \label{fig:schedule-payload-separation}
\end{figure}

\begin{figure}[!htbp]
  \centering
  \includegraphics[width=\linewidth]{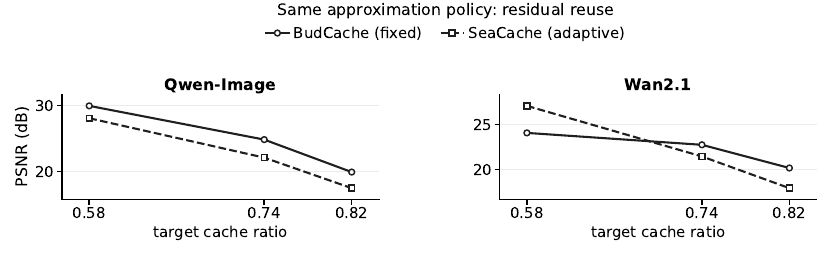}
  \caption{\textbf{Schedule choice also affects quality on Qwen-Image and Wan2.1.}
  Both methods use residual reuse.
  Circles show fixed BudCache schedules and squares show adaptive SeaCache.
  PSNR means give equal weight to four image datasets or two video datasets.
  The horizontal axes show target cache ratios because SeaCache's realized
  cache counts vary. Tables~\ref{tab:actual-k-flux}--\ref{tab:actual-k-wan21}
  in Appendix~\ref{app:ratio-alignment} report these counts. Table~\ref{tab:spx-axis-controls} gives the PSNR differences.}
  \label{fig:schedule-control-reuse}
\end{figure}

\paragraph{Holding one choice fixed.}
The fixed-schedule comparisons in Figure~\ref{fig:schedule-payload-separation}
evaluate the five approximation policies on BudCache schedules.
FLUX.1-dev uses 1,632 PartiPrompts under three seeds.
HunyuanVideo uses 300 prompts, split evenly between Penguin599 and VBench944,
with one seed per prompt.
The comparisons with a fixed approximation policy instead use the three-seed baseline results
of Appendix~\ref{app:complete-tables}.
They compare fixed BudCache schedules with adaptive SeaCache schedules under
residual reuse, averaging equally over four image datasets or two video datasets.
Figure~\ref{fig:schedule-control-reuse} gives this comparison for Qwen-Image and Wan2.1.
The two methods' mean PSNR differs by up to 3.2 dB on images and 3.0 dB on videos.
Table~\ref{tab:spx-axis-controls} also compares TaylorSeer, HiCache, and L2P,
which share a schedule but use different approximation policies.
Its PSNR range across approximation policies is the difference between their highest and lowest mean PSNR.

\input{\arxivroot tables/spx_partition_results}

%% file: tables/four_factor_ladder_full.tex
\begin{table}[!htbp]
\caption{Combining feature error with propagation gain gives the highest correlation with
output error magnitude.  We compute Spearman correlations over 100 prompts and
steps 1--48, giving 4,800 isolated cached steps per approximation policy.
The step-only score is the negative cached-step index $-k$.
The propagation gain $h_k^{\mathrm{LOO}}$ uses the other 99 prompts at the
same step and with the same approximation policy.
The next-state error estimate is $\norm{r_k}|\Delta\tau_k|$ multiplied by
the first mean ratio in Equation~\ref{eq:empirical-propagation-gain}.}
\label{tab:four-factor-ladder-full}
\centering
\footnotesize
\setlength{\tabcolsep}{3pt}
\begin{tabular}{lrrrrrr}
\toprule
\shortstack{Approximation\\policy} & \shortstack{Step only\\$-k$} &
\shortstack{Feature error\\$\norm{r_k}$} &
\shortstack{Feature error\\$\times$ step size\\$\norm{r_k}|\Delta\tau_k|$} &
\shortstack{Next-state error\\estimate} &
\shortstack{Propagation\\gain $h_k^{\rm LOO}$} &
\shortstack{Feature error\\$\times$ gain\\$\norm{r_k}h_k^{\rm LOO}$} \\
\midrule
Residual reuse & 0.867 & 0.528 & -0.382 & -0.215 & 0.803 & \textbf{0.886} \\
Hermite order 2 & 0.880 & 0.586 & -0.391 & -0.180 & 0.800 & \textbf{0.896} \\
Taylor order 1 & 0.880 & 0.676 & -0.350 & -0.115 & 0.788 & \textbf{0.899} \\
\bottomrule
\end{tabular}
\end{table}

%% file: tables/spx_partition_results.tex
\begin{table}[!htbp]
\caption{\textbf{Interactions between schedules and approximation policies account for 13--27\% of the
variation in mean PSNR at cache ratios 0.74 and 0.82.}
For each model and cache ratio, we fit an additive model to the 20 mean PSNR
values from four schedules and five approximation policies, with equal weight
for every combination.  The schedule and approximation policy columns give
the fractions of the total sum of squares explained by their separate effects.
The interaction column gives the remaining fraction.
The last column is the average of the 20 mean PSNR values, in dB.}
\label{tab:spx-image-decomposition}
\centering
\small
\begin{tabular}{lrrrrr}
\toprule
Model & $K$ & Schedule & \shortstack{Approximation\\policy} & Interaction & Mean PSNR (dB) \\
\midrule
FLUX.1-dev & 29 & .950 & .037 & .013 & 25.20 \\
FLUX.1-dev & 37 & .613 & .117 & .270 & 20.95 \\
FLUX.1-dev & 41 & .493 & .269 & .238 & 18.53 \\
Qwen-Image & 29 & .983 & .012 & .005 & 26.78 \\
Qwen-Image & 37 & .657 & .181 & .162 & 20.26 \\
Qwen-Image & 41 & .368 & .499 & .133 & 15.79 \\
\bottomrule
\end{tabular}
\end{table}

\begin{table}[!htbp]
\caption{\textbf{BudCache and SeaCache differ by up to 3.24 dB on images and
3.01 dB on video under residual reuse.}
``PSNR range across approximation policies''
is the difference between the highest and lowest mean PSNR of TaylorSeer,
HiCache, and L2P, which share a schedule.
``Schedule difference'' subtracts adaptive SeaCache's mean PSNR from the
mean PSNR obtained with the fixed BudCache schedule. Both use
residual reuse at the same target ratio, as in
Figures~\ref{fig:schedule-payload-separation} and~\ref{fig:schedule-control-reuse}.
We give equal weight to the mean PSNR of each dataset, using four image
datasets or two video datasets.  Positive schedule differences favor BudCache.}
\label{tab:spx-axis-controls}
\centering
\small
\begin{tabular}{lrrr}
\toprule
Model & $K$ & \shortstack{PSNR range across\\approximation policies (dB)} & Schedule difference (dB) \\
\midrule
FLUX.1-dev & 29 & 1.457 & -3.238 \\
FLUX.1-dev & 37 & 1.502 & +1.002 \\
FLUX.1-dev & 41 & 2.897 & +0.708 \\
Qwen-Image & 29 & 1.683 & +1.870 \\
Qwen-Image & 37 & 5.319 & +2.727 \\
Qwen-Image & 41 & 6.225 & +2.431 \\
HunyuanVideo & 29 & 0.606 & +1.806 \\
HunyuanVideo & 37 & 2.166 & +2.096 \\
HunyuanVideo & 41 & 3.683 & +2.758 \\
Wan2.1 & 29 & 1.995 & -3.007 \\
Wan2.1 & 37 & 4.189 & +1.295 \\
Wan2.1 & 41 & 3.030 & +2.240 \\
\bottomrule
\end{tabular}
\end{table}

%% file: tables/search_arbitration.tex
\begingroup
\scriptsize
\setlength{\tabcolsep}{4pt}
\setlength{\aboverulesep}{0.5pt}
\setlength{\belowrulesep}{0.5pt}
\setlength{\LTpre}{2pt}
\renewcommand{\arraystretch}{0.95}
\setlength{\LTcapwidth}{\linewidth}
\begin{longtable}{llrrrrr}
\caption{\textbf{Validation changes the selected candidate in eight search runs.}
The replaced candidate has the highest mean PSNR on the eight scoring runs.
The selected candidate has the highest mean PSNR on 50 validation prompts.
The first two numeric columns give the candidates' mean validation PSNR.
For each metric, we subtract the value for the replaced candidate's output
from the value for the selected candidate's output on the same prompt and seed.
The last three columns report the means of these differences over
4,896 PartiPrompts prompt--seed runs.  Positive PSNR and SSIM differences and
negative LPIPS differences favor the selected candidate.  All PSNR values are
in dB.  Brackets give a 95\% bootstrap interval for the PSNR difference.
Hill, anneal, and greedy denote hill climbing, annealing, and greedy coordinate ascent.}
\label{tab:search-arbitration}\\
\toprule
Model, cache ratio & Procedure & Selected PSNR & Replaced PSNR & $\Delta$PSNR (dB) & $\Delta$SSIM & $\Delta$LPIPS \\
\midrule
\endfirsthead
\caption[]{Continued.}\\
\toprule
Model, cache ratio & Procedure & Selected PSNR & Replaced PSNR & $\Delta$PSNR (dB) & $\Delta$SSIM & $\Delta$LPIPS \\
\midrule
\endhead
\bottomrule
\endfoot
FLUX.1-dev, ratio 0.58 & hill & 30.705 & 30.418 & +0.072 [+0.052, +0.091] & +0.0010 & -0.0005 \\
FLUX.1-dev, ratio 0.74 & anneal & 24.663 & 24.654 & -0.002 [-0.011, +0.006] & +0.0022 & -0.0008 \\
FLUX.1-dev, ratio 0.74 & greedy & 24.499 & 24.476 & +0.221 [+0.182, +0.255] & +0.0011 & +0.0019 \\
FLUX.1-dev, ratio 0.74 & hill & 24.832 & 24.825 & -0.012 [-0.020, -0.006] & +0.0006 & -0.0008 \\
Qwen-Image, ratio 0.58 & anneal & 30.381 & 30.201 & +0.149 [+0.128, +0.170] & +0.0011 & -0.0015 \\
Qwen-Image, ratio 0.82 & anneal & 20.739 & 20.662 & +0.001 [-0.014, +0.019] & -0.0002 & +0.0005 \\
Qwen-Image, ratio 0.82 & greedy & 20.481 & 20.423 & -0.015 [-0.024, -0.006] & +0.0044 & -0.0029 \\
Qwen-Image, ratio 0.82 & hill & 20.766 & 20.712 & +0.050 [+0.042, +0.058] & -0.0042 & +0.0045 \\
\end{longtable}
\endgroup

%% file: tables/search_paired_full.tex
\vskip .3cm
\begingroup
\normalsize
\setlength{\tabcolsep}{3pt}
\setlength{\LTcapwidth}{\linewidth}
\setlength{\LTpre}{0pt}
\renewcommand{\arraystretch}{0.7273}
\begin{longtable}{*{3}{>{\scriptsize}l}*{6}{>{\scriptsize}r}}
\caption{\textbf{Searched schedules improve PSNR over BudCache in every reported
model, cache ratio, and dataset combination.}  Each row compares the selected
schedule with one reference schedule under residual reuse.
For each metric, we subtract the value for the reference schedule's output
from the value for the searched schedule's output, using the same prompt and seed.
Entries are means of these differences over the listed prompt--seed runs.
Positive PSNR, SSIM, ImageReward, and CLIP differences and negative LPIPS
differences favor the searched schedule.  IR denotes ImageReward.
PSNR differences are in dB.  Brackets give 95\% bootstrap intervals obtained
by resampling prompt--seed runs individually.}
\label{tab:search-paired-full}\\
\toprule
Model, cache ratio & Dataset & Reference & Runs & $\Delta$PSNR (dB) & $\Delta$SSIM & $\Delta$LPIPS & $\Delta$IR & $\Delta$CLIP \\
\midrule
\endfirsthead
\multicolumn{9}{l}{\scriptsize Table~\thetable\ continued} \\
\toprule
Model, cache ratio & Dataset & Reference & Runs & $\Delta$PSNR (dB) & $\Delta$SSIM & $\Delta$LPIPS & $\Delta$IR & $\Delta$CLIP \\
\midrule
\endhead
\midrule
\multicolumn{9}{r}{\scriptsize Continued on next page} \\
\endfoot
\bottomrule
\endlastfoot
FLUX.1-dev, ratio 0.58 & DrawBench & MeanCache & 600 & +0.697 [0.55, 0.85] & -0.0062 & +0.0114 & -0.012 & -0.013 \\*
 & DrawBench & BudCache & 600 & +5.489 [5.18, 5.80] & +0.0558 & -0.0539 & -0.017 & -0.058 \\
FLUX.1-dev, ratio 0.58 & GenEval-style & MeanCache & 1,659 & +0.661 [0.56, 0.77] & -0.0049 & +0.0084 & -0.011 & +0.108 \\*
 & GenEval-style & BudCache & 1,659 & +6.214 [6.03, 6.40] & +0.0517 & -0.0535 & -0.014 & +0.121 \\
FLUX.1-dev, ratio 0.58 & PartiPrompts & MeanCache & 4,896 & +0.277 [0.22, 0.33] & -0.0124 & +0.0191 & -0.016 & +0.062 \\*
 & PartiPrompts & BudCache & 4,896 & +4.351 [4.25, 4.45] & +0.0375 & -0.0339 & -0.015 & +0.075 \\
FLUX.1-dev, ratio 0.58 & DiffusionDB & MeanCache & 30,000 & +0.370 [0.35, 0.39] & -0.0136 & +0.0229 & -0.026 & -0.037 \\*
 & DiffusionDB & BudCache & 30,000 & +3.259 [3.22, 3.29] & +0.0340 & -0.0312 & -0.025 & -0.038 \\
\midrule
FLUX.1-dev, ratio 0.74 & DrawBench & MeanCache & 600 & +0.492 [0.42, 0.56] & +0.0027 & +0.0029 & -0.003 & +0.044 \\*
 & DrawBench & BudCache & 600 & +1.139 [1.03, 1.25] & +0.0083 & -0.0044 & -0.013 & +0.048 \\
FLUX.1-dev, ratio 0.74 & GenEval-style & MeanCache & 1,659 & +0.503 [0.46, 0.55] & +0.0031 & -0.0000 & -0.013 & +0.050 \\*
 & GenEval-style & BudCache & 1,659 & +1.168 [1.10, 1.23] & +0.0080 & -0.0059 & -0.012 & +0.080 \\
FLUX.1-dev, ratio 0.74 & PartiPrompts & MeanCache & 4,896 & +0.357 [0.33, 0.38] & -0.0016 & +0.0086 & -0.017 & +0.036 \\*
 & PartiPrompts & BudCache & 4,896 & +0.906 [0.87, 0.94] & +0.0007 & +0.0043 & -0.020 & +0.052 \\
FLUX.1-dev, ratio 0.74 & DiffusionDB & MeanCache & 30,000 & +0.318 [0.31, 0.33] & -0.0030 & +0.0153 & -0.027 & -0.028 \\*
 & DiffusionDB & BudCache & 30,000 & +0.741 [0.73, 0.75] & -0.0021 & +0.0101 & -0.029 & -0.016 \\
\midrule
FLUX.1-dev, ratio 0.82 & DrawBench & MeanCache & 600 & +0.408 [0.31, 0.50] & -0.0044 & +0.0110 & -0.047 & +0.111 \\*
 & DrawBench & BudCache & 600 & +0.474 [0.35, 0.60] & -0.0165 & +0.0192 & -0.046 & +0.106 \\
FLUX.1-dev, ratio 0.82 & GenEval-style & MeanCache & 1,659 & +0.576 [0.51, 0.64] & +0.0008 & -0.0021 & -0.026 & +0.130 \\*
 & GenEval-style & BudCache & 1,659 & +0.695 [0.60, 0.78] & -0.0077 & +0.0036 & -0.026 & +0.130 \\
FLUX.1-dev, ratio 0.82 & PartiPrompts & MeanCache & 4,896 & +0.425 [0.40, 0.46] & -0.0068 & +0.0110 & -0.070 & +0.017 \\*
 & PartiPrompts & BudCache & 4,896 & +0.311 [0.27, 0.35] & -0.0233 & +0.0236 & -0.073 & -0.007 \\
FLUX.1-dev, ratio 0.82 & DiffusionDB & MeanCache & 30,000 & +0.441 [0.43, 0.45] & -0.0040 & +0.0183 & -0.130 & -0.405 \\*
 & DiffusionDB & BudCache & 30,000 & +0.117 [0.11, 0.13] & -0.0247 & +0.0341 & -0.137 & -0.373 \\
\midrule
Qwen-Image, ratio 0.58 & DrawBench & MeanCache & 600 & -0.021 [-0.09, 0.05] & -0.0012 & +0.0019 & +0.002 & +0.045 \\*
 & DrawBench & BudCache & 600 & +1.188 [0.98, 1.41] & +0.0032 & +0.0011 & -0.005 & +0.011 \\
Qwen-Image, ratio 0.58 & GenEval-style & MeanCache & 1,659 & -0.048 [-0.09, -0.00] & -0.0014 & +0.0019 & -0.000 & +0.019 \\*
 & GenEval-style & BudCache & 1,659 & +0.844 [0.71, 0.97] & -0.0026 & +0.0041 & -0.002 & +0.024 \\
Qwen-Image, ratio 0.58 & PartiPrompts & MeanCache & 4,896 & +0.040 [0.02, 0.06] & -0.0011 & +0.0020 & +0.000 & +0.012 \\*
 & PartiPrompts & BudCache & 4,896 & +1.075 [1.01, 1.15] & +0.0008 & +0.0044 & -0.004 & +0.013 \\
Qwen-Image, ratio 0.58 & DiffusionDB & MeanCache & 30,000 & +0.094 [0.09, 0.10] & -0.0008 & +0.0018 & -0.001 & +0.013 \\*
 & DiffusionDB & BudCache & 30,000 & +1.154 [1.13, 1.18] & +0.0041 & +0.0028 & -0.006 & +0.002 \\
\midrule
Qwen-Image, ratio 0.74 & DrawBench & MeanCache & 600 & +0.073 [-0.09, 0.25] & -0.0099 & +0.0173 & -0.006 & +0.040 \\*
 & DrawBench & BudCache & 600 & +0.413 [0.27, 0.57] & -0.0024 & +0.0053 & +0.003 & +0.022 \\
Qwen-Image, ratio 0.74 & GenEval-style & MeanCache & 1,659 & -0.086 [-0.17, 0.00] & -0.0115 & +0.0140 & -0.005 & -0.006 \\*
 & GenEval-style & BudCache & 1,659 & +0.279 [0.20, 0.36] & -0.0055 & +0.0038 & -0.002 & -0.026 \\
Qwen-Image, ratio 0.74 & PartiPrompts & MeanCache & 4,896 & -0.022 [-0.07, 0.03] & -0.0132 & +0.0209 & -0.019 & +0.034 \\*
 & PartiPrompts & BudCache & 4,896 & +0.350 [0.31, 0.39] & -0.0058 & +0.0090 & -0.010 & +0.021 \\
Qwen-Image, ratio 0.74 & DiffusionDB & MeanCache & 30,000 & -0.031 [-0.05, -0.02] & -0.0143 & +0.0291 & -0.027 & -0.037 \\*
 & DiffusionDB & BudCache & 30,000 & +0.298 [0.28, 0.31] & -0.0071 & +0.0163 & -0.016 & -0.019 \\
\midrule
Qwen-Image, ratio 0.82 & DrawBench & MeanCache & 600 & +0.700 [0.57, 0.83] & +0.0091 & -0.0123 & +0.009 & +0.085 \\*
 & DrawBench & BudCache & 600 & +1.574 [1.44, 1.73] & +0.0120 & -0.0159 & -0.028 & -0.010 \\
Qwen-Image, ratio 0.82 & GenEval-style & MeanCache & 1,659 & +0.957 [0.87, 1.04] & +0.0087 & -0.0132 & +0.003 & +0.005 \\*
 & GenEval-style & BudCache & 1,659 & +1.985 [1.89, 2.08] & +0.0123 & -0.0178 & -0.010 & -0.013 \\
Qwen-Image, ratio 0.82 & PartiPrompts & MeanCache & 4,896 & +0.790 [0.75, 0.83] & +0.0085 & -0.0108 & +0.007 & +0.019 \\*
 & PartiPrompts & BudCache & 4,896 & +1.514 [1.46, 1.56] & +0.0076 & -0.0079 & -0.019 & +0.030 \\
Qwen-Image, ratio 0.82 & DiffusionDB & MeanCache & 30,000 & +0.710 [0.70, 0.72] & +0.0084 & -0.0080 & -0.001 & -0.010 \\*
 & DiffusionDB & BudCache & 30,000 & +1.065 [1.05, 1.08] & +0.0030 & +0.0067 & -0.041 & -0.130 \\
\end{longtable}
\endgroup

%% file: tables/search_payloads.tex
\begin{table}[!htbp]
\caption{\textbf{Two-anchor prediction gives the highest mean PSNR on all six
searched schedules.}  Each row holds the PSNR-selected schedule fixed.
Mean PSNR is in dB and gives equal weight to the four evaluation datasets,
totaling 37,155 prompt--seed runs per row and approximation policy.
Bold marks the highest mean in each row.  Interval-average denotes
MeanCache's interval-average velocity prediction. Two-anchor uses the first
transformer block to extrapolate from two full-step residuals.}
\label{tab:search-payloads}
\centering
\small
\setlength{\tabcolsep}{4pt}
\begin{tabular}{llrrrrr}
\toprule
Model & Cache ratio & Reuse & Taylor O1 & Hermite O2 & Interval-average & Two-anchor \\
\midrule
FLUX.1-dev & 0.58 & 29.104 & 29.931 & 30.257 & 30.342 & \textbf{30.775} \\
FLUX.1-dev & 0.74 & 24.001 & 24.013 & 24.513 & 24.542 & \textbf{24.990} \\
FLUX.1-dev & 0.82 & 21.060 & 19.459 & 20.334 & 20.325 & \textbf{21.131} \\
Qwen-Image & 0.58 & 31.017 & 31.812 & 31.981 & 32.205 & \textbf{32.450} \\
Qwen-Image & 0.74 & 25.183 & 24.612 & 25.294 & 24.549 & \textbf{25.637} \\
Qwen-Image & 0.82 & 21.471 & 20.473 & 21.414 & 20.346 & \textbf{21.628} \\
\bottomrule
\end{tabular}
\end{table}

%% file: gph_arxiv.bbl
\begin{thebibliography}{40}
\providecommand{\natexlab}[1]{#1}
\providecommand{\url}[1]{\texttt{#1}}
\expandafter\ifx\csname urlstyle\endcsname\relax
  \providecommand{\doi}[1]{doi: #1}\else
  \providecommand{\doi}{doi: \begingroup \urlstyle{rm}\Url}\fi

\bibitem[Aliev et~al.(2026)Aliev, Neudachina, Bykov, Oganov, Struminsky, Alanov, and Rakitin]{recache2026}
Mishan Aliev, Eva Neudachina, Ilya Bykov, Aleksandr Oganov, Kirill Struminsky, Aibek Alanov, and Denis Rakitin.
\newblock {ReCache}: Learning budget-aware caching schedules for diffusion models via {REINFORCE}.
\newblock \emph{arXiv preprint arXiv:2606.06060}, 2026.

\bibitem[{Black Forest Labs}(2024)]{flux2024}
{Black Forest Labs}.
\newblock {FLUX}.
\newblock \url{https://github.com/black-forest-labs/flux}, 2024.

\bibitem[Bu et~al.(2025)Bu, Ling, Zhou, Wang, Zang, Lin, and Wang]{dicache2025}
Jiazi Bu, Pengyang Ling, Yujie Zhou, Yibin Wang, Yuhang Zang, Dahua Lin, and Jiaqi Wang.
\newblock {DiCache}: Let diffusion model determine its own cache.
\newblock \emph{arXiv preprint arXiv:2508.17356}, 2025.

\bibitem[Chen et~al.(2024{\natexlab{a}})Chen, Zhou, Wang, Shen, and Lyu]{chen2024trajectory}
Defang Chen, Zhenyu Zhou, Can Wang, Chunhua Shen, and Siwei Lyu.
\newblock On the trajectory regularity of {ODE}-based diffusion sampling.
\newblock In \emph{International Conference on Machine Learning}, 2024{\natexlab{a}}.
\newblock arXiv:2405.11326.

\bibitem[Chen et~al.(2026)Chen, Zheng, Lin, and Zhang]{svdcache2026}
Guantao Chen, Shikang Zheng, Yuqi Lin, and Linfeng Zhang.
\newblock Forecast the principal, stabilize the residual: Subspace-aware feature caching for efficient diffusion transformers.
\newblock \emph{arXiv preprint arXiv:2601.07396}, 2026.

\bibitem[Chen et~al.(2024{\natexlab{b}})Chen, Shen, Ye, Cao, Tu, Bouganis, Zhao, and Chen]{deltadit2024}
Pengtao Chen, Mingzhu Shen, Peng Ye, Jianjian Cao, Chongjun Tu, Christos-Savvas Bouganis, Yiren Zhao, and Tao Chen.
\newblock {$\Delta$-DiT}: A training-free acceleration method tailored for diffusion transformers.
\newblock \emph{arXiv preprint arXiv:2406.01125}, 2024{\natexlab{b}}.

\bibitem[Chu et~al.(2025)Chu, Wu, Feng, and Zhang]{chu2025omnicache}
Huanpeng Chu, Wei Wu, Guanyu Feng, and Yutao Zhang.
\newblock {OmniCache}: A trajectory-oriented global perspective on training-free cache reuse for diffusion transformer models.
\newblock In \emph{Proceedings of the IEEE/CVF International Conference on Computer Vision}, pages 16302--16312, 2025.
\newblock arXiv:2508.16212.

\bibitem[Chung et~al.(2026)Chung, Hyun, Lee, Han, Cha, Wee, Hong, and Heo]{seacache2026}
Jiwoo Chung, Sangeek Hyun, MinKyu Lee, Byeongju Han, Geonho Cha, Dongyoon Wee, Youngjun Hong, and Jae-Pil Heo.
\newblock {SeaCache}: Spectral-evolution-aware cache for accelerating diffusion models.
\newblock In \emph{Proceedings of the IEEE/CVF Conference on Computer Vision and Pattern Recognition}, pages 14283--14294, 2026.
\newblock arXiv:2602.18993.

\bibitem[Cui et~al.(2026)Cui, Wang, Xu, Chen, Zhang, Jiang, Jin, Liu, and Huang]{dpcache2026}
Bowen Cui, Yuanbin Wang, Huajiang Xu, Biaolong Chen, Aixi Zhang, Hao Jiang, Zhengzheng Jin, Xu~Liu, and Pipei Huang.
\newblock Denoising as path planning: Training-free acceleration of diffusion models with {DPCache}.
\newblock In \emph{Proceedings of the IEEE/CVF Conference on Computer Vision and Pattern Recognition}, pages 43632--43642, 2026.
\newblock arXiv:2602.22654.

\bibitem[Feng et~al.(2026)Feng, Zheng, Liu, Lin, Zhou, Cai, Wang, Chen, Zou, Ma, and Zhang]{hicache2026}
Liang Feng, Shikang Zheng, Jiacheng Liu, Yuqi Lin, Qinming Zhou, Peiliang Cai, Xinyu Wang, Junjie Chen, Chang Zou, Yue Ma, and Linfeng Zhang.
\newblock {HiCache}: A plug-in scaled-hermite upgrade for taylor-style cache-then-forecast diffusion acceleration.
\newblock In \emph{International Conference on Learning Representations}, 2026.
\newblock arXiv:2508.16984.

\bibitem[Gao et~al.(2026{\natexlab{a}})Gao, Chen, Shi, Wu, Li, Hui, et~al.]{meancache2026}
Huanlin Gao, Ping Chen, Fuyuan Shi, Ruijia Wu, Yantao Li, Qiang Hui, et~al.
\newblock {MeanCache}: From instantaneous to average velocity for accelerating flow matching inference.
\newblock In \emph{International Conference on Learning Representations}, 2026{\natexlab{a}}.
\newblock arXiv:2601.19961.

\bibitem[Gao et~al.(2026{\natexlab{b}})Gao, Zhao, Hui, Shi, Zhao, Li, Tan, Lu, You, Wang, and Lian]{otcache2026}
Huanlin Gao, Fang Zhao, Qiang Hui, Fuyuan Shi, Shaoan Zhao, Yantao Li, Chao Tan, Ting Lu, Yuren You, Kai Wang, and Shiguo Lian.
\newblock {OTCache}: Optimal transport for geometry-aware caching in diffusion models.
\newblock \emph{arXiv preprint arXiv:2606.31026}, 2026{\natexlab{b}}.

\bibitem[Ghosh et~al.(2023)Ghosh, Hajishirzi, and Schmidt]{ghosh2023geneval}
Dhruba Ghosh, Hannaneh Hajishirzi, and Ludwig Schmidt.
\newblock {GenEval}: An object-focused framework for evaluating text-to-image alignment.
\newblock In \emph{Advances in Neural Information Processing Systems}, 2023.

\bibitem[Haghighi and Alahi(2026)]{sencache2026}
Yasaman Haghighi and Alexandre Alahi.
\newblock {SenCache}: Accelerating diffusion model inference via sensitivity-aware caching.
\newblock In \emph{Proceedings of the IEEE/CVF Conference on Computer Vision and Pattern Recognition}, pages 14295--14304, 2026.
\newblock arXiv:2602.24208.

\bibitem[Ho et~al.(2020)Ho, Jain, and Abbeel]{ho2020denoising}
Jonathan Ho, Ajay Jain, and Pieter Abbeel.
\newblock Denoising diffusion probabilistic models.
\newblock In \emph{Advances in Neural Information Processing Systems}, 2020.

\bibitem[Huang et~al.(2024)Huang, He, Yu, Zhang, Si, Jiang, Zhang, Wu, Jin, Chanpaisit, et~al.]{vbench2024}
Ziqi Huang, Yinan He, Jiashuo Yu, Fan Zhang, Chenyang Si, Yuming Jiang, Yuanhan Zhang, Tianxing Wu, Qingyang Jin, Nattapol Chanpaisit, et~al.
\newblock {VBench}: Comprehensive benchmark suite for video generative models.
\newblock In \emph{Proceedings of the IEEE/CVF Conference on Computer Vision and Pattern Recognition}, 2024.

\bibitem[Kong et~al.(2024)Kong, Tian, Zhang, Min, Dai, Zhou, Xiong, Li, Wu, Zhang, et~al.]{hunyuanvideo2024}
Weijie Kong, Qi~Tian, Zijian Zhang, Rox Min, Zuozhuo Dai, Jin Zhou, Jiangfeng Xiong, Xin Li, Bo~Wu, Jianwei Zhang, et~al.
\newblock {HunyuanVideo}: A systematic framework for large video generative models.
\newblock \emph{arXiv preprint arXiv:2412.03603}, 2024.

\bibitem[Lei et~al.(2026)Lei, Zhao, Yuan, and Zhang]{budcache2026}
Mingkun Lei, Tong Zhao, Liangyu Yuan, and Chi Zhang.
\newblock Budget-constrained step-level diffusion caching.
\newblock In \emph{International Conference on Machine Learning}, 2026.
\newblock arXiv:2606.13496.

\bibitem[Lin et~al.(2014)Lin, Maire, Belongie, Hays, Perona, Ramanan, Doll{\'a}r, and Zitnick]{lin2014coco}
Tsung-Yi Lin, Michael Maire, Serge Belongie, James Hays, Pietro Perona, Deva Ramanan, Piotr Doll{\'a}r, and C.~Lawrence Zitnick.
\newblock Microsoft coco: Common objects in context.
\newblock In \emph{European Conference on Computer Vision}, pages 740--755, 2014.

\bibitem[Lipman et~al.(2022)Lipman, Chen, Ben-Hamu, Nickel, and Le]{lipman2022flow}
Yaron Lipman, Ricky T.~Q. Chen, Heli Ben-Hamu, Maximilian Nickel, and Matt Le.
\newblock Flow matching for generative modeling.
\newblock \emph{arXiv preprint arXiv:2210.02747}, 2022.

\bibitem[Liu et~al.(2025{\natexlab{a}})Liu, Zhang, Wang, Wei, Qiu, Zhao, Zhang, Ye, and Wan]{teacache2025}
Feng Liu, Shiwei Zhang, Xiaofeng Wang, Yujie Wei, Haonan Qiu, Yuzhong Zhao, Yingya Zhang, Qixiang Ye, and Fang Wan.
\newblock Timestep embedding tells: It's time to cache for video diffusion model.
\newblock In \emph{Proceedings of the IEEE/CVF Conference on Computer Vision and Pattern Recognition}, 2025{\natexlab{a}}.
\newblock arXiv:2411.19108.

\bibitem[Liu et~al.(2025{\natexlab{b}})Liu, Zou, Lyu, Chen, and Zhang]{taylorseer2025}
Jiacheng Liu, Chang Zou, Yuanhuiyi Lyu, Junjie Chen, and Linfeng Zhang.
\newblock From reusing to forecasting: Accelerating diffusion models with {TaylorSeers}.
\newblock In \emph{Proceedings of the IEEE/CVF International Conference on Computer Vision}, 2025{\natexlab{b}}.
\newblock arXiv:2503.06923.

\bibitem[Ma et~al.(2024)Ma, Fang, and Wang]{ma2024deepcache}
Xinyin Ma, Gongfan Fang, and Xinchao Wang.
\newblock {DeepCache}: Accelerating diffusion models for free.
\newblock In \emph{Proceedings of the IEEE/CVF Conference on Computer Vision and Pattern Recognition}, 2024.

\bibitem[Radford et~al.(2021)Radford, Kim, Hallacy, Ramesh, Goh, Agarwal, Sastry, Askell, Mishkin, Clark, et~al.]{radford2021clip}
Alec Radford, Jong~Wook Kim, Chris Hallacy, Aditya Ramesh, Gabriel Goh, Sandhini Agarwal, Girish Sastry, Amanda Askell, Pamela Mishkin, Jack Clark, et~al.
\newblock Learning transferable visual models from natural language supervision.
\newblock In \emph{International Conference on Machine Learning}, 2021.

\bibitem[Saharia et~al.(2022)Saharia, Chan, Saxena, Li, Whang, Denton, Ghasemipour, Ayan, Mahdavi, Gontijo-Lopes, et~al.]{saharia2022drawbench}
Chitwan Saharia, William Chan, Saurabh Saxena, Lala Li, Jay Whang, Emily Denton, Seyed Kamyar~Seyed Ghasemipour, Burcu~Karagol Ayan, S.~Sara Mahdavi, Raphael Gontijo-Lopes, et~al.
\newblock Photorealistic text-to-image diffusion models with deep language understanding.
\newblock In \emph{Advances in Neural Information Processing Systems}, 2022.

\bibitem[Selvaraju et~al.(2024)Selvaraju, Ding, Chen, Zharkov, and Liang]{selvaraju2024fora}
Pratheba Selvaraju, Tianyu Ding, Tianyi Chen, Ilya Zharkov, and Luming Liang.
\newblock {FORA}: Fast-forward caching in diffusion transformer acceleration.
\newblock \emph{arXiv preprint arXiv:2407.01425}, 2024.

\bibitem[Shen et~al.(2026)Shen, Huang, Liu, Zou, Cai, Zheng, Shi, Feng, and Zhang]{l2p2026}
Zhirong Shen, Rui Huang, Jiacheng Liu, Chang Zou, Peiliang Cai, Shikang Zheng, Zhengyi Shi, Liang Feng, and Linfeng Zhang.
\newblock Beyond fixed formulas: Data-driven linear predictor for efficient diffusion models.
\newblock In \emph{Proceedings of the IEEE/CVF Conference on Computer Vision and Pattern Recognition}, pages 30792--30801, 2026.

\bibitem[Son et~al.(2026)Son, Jeon, Choi, and Ham]{rfc2026}
Byunggwan Son, Jeimin Jeon, Jeongwoo Choi, and Bumsub Ham.
\newblock Relational feature caching for accelerating diffusion transformers.
\newblock \emph{arXiv preprint arXiv:2602.19506}, 2026.

\bibitem[{Team Wan}(2025)]{wan2025}
{Team Wan}.
\newblock {Wan}: Open and advanced large-scale video generative models.
\newblock \emph{arXiv preprint arXiv:2503.20314}, 2025.

\bibitem[{Tencent Hunyuan}(2025)]{penguin2025}
{Tencent Hunyuan}.
\newblock Penguin video benchmark.
\newblock \url{https://github.com/Tencent-Hunyuan/HunyuanVideo/blob/main/assets/PenguinVideoBenchmark.csv}, 2025.

\bibitem[Wang et~al.(2004)Wang, Bovik, Sheikh, and Simoncelli]{wang2004ssim}
Zhou Wang, Alan~C. Bovik, Hamid~R. Sheikh, and Eero~P. Simoncelli.
\newblock Image quality assessment: From error visibility to structural similarity.
\newblock \emph{IEEE Transactions on Image Processing}, 13\penalty0 (4):\penalty0 600--612, 2004.

\bibitem[Wang et~al.(2022)Wang, Montoya, Munechika, Yang, Hoover, and Chau]{wang2022diffusiondb}
Zijie~J. Wang, Evan Montoya, David Munechika, Haoyang Yang, Benjamin Hoover, and Duen~Horng Chau.
\newblock {DiffusionDB}: A large-scale prompt gallery dataset for text-to-image generative models.
\newblock \emph{arXiv preprint arXiv:2210.14896}, 2022.

\bibitem[Wu et~al.(2025)Wu, Li, Zhou, Lin, Gao, Yan, Yin, Bai, Xu, Chen, et~al.]{qwenimage2025}
Chenfei Wu, Jiahao Li, Jingren Zhou, Junyang Lin, Kaiyuan Gao, Kun Yan, Sheng-ming Yin, Shuai Bai, Xiao Xu, Yilei Chen, et~al.
\newblock {Qwen-Image} technical report, 2025.
\newblock arXiv:2508.02324.

\bibitem[Xu et~al.(2023)Xu, Liu, Wu, Tong, Li, Ding, Tang, and Dong]{xu2023imagereward}
Jiazheng Xu, Xiao Liu, Yuchen Wu, Yuxuan Tong, Qinkai Li, Ming Ding, Jie Tang, and Yuxiao Dong.
\newblock {ImageReward}: Learning and evaluating human preferences for text-to-image generation.
\newblock In \emph{Advances in Neural Information Processing Systems}, 2023.

\bibitem[Yu et~al.(2022)Yu, Xu, Koh, Luong, Baid, Wang, Vasudevan, Ku, Yang, Ayan, et~al.]{yu2022parti}
Jiahui Yu, Yuanzhong Xu, Jing~Yu Koh, Thang Luong, Gunjan Baid, Zirui Wang, Vijay Vasudevan, Alexander Ku, Yinfei Yang, Burcu~Karagol Ayan, et~al.
\newblock Scaling autoregressive models for content-rich text-to-image generation.
\newblock \emph{arXiv preprint arXiv:2206.10789}, 2022.

\bibitem[Zhang et~al.(2018)Zhang, Isola, Efros, Shechtman, and Wang]{zhang2018lpips}
Richard Zhang, Phillip Isola, Alexei~A. Efros, Eli Shechtman, and Oliver Wang.
\newblock The unreasonable effectiveness of deep features as a perceptual metric.
\newblock In \emph{Proceedings of the IEEE Conference on Computer Vision and Pattern Recognition}, 2018.

\bibitem[Zheng et~al.(2026{\natexlab{a}})Zheng, Chen, He, Liu, Lin, Zou, and Zhang]{fresco2026}
Shikang Zheng, Guantao Chen, Lixuan He, Jiacheng Liu, Yuqi Lin, Chang Zou, and Linfeng Zhang.
\newblock From sketch to fresco: Efficient diffusion transformer with progressive resolution.
\newblock In \emph{Proceedings of the IEEE/CVF Conference on Computer Vision and Pattern Recognition}, 2026{\natexlab{a}}.
\newblock arXiv:2601.07462.

\bibitem[Zheng et~al.(2026{\natexlab{b}})Zheng, Chen, Zhou, Lin, He, Zou, Cai, Liu, and Zhang]{hyca2026}
Shikang Zheng, Guantao Chen, Qinming Zhou, Yuqi Lin, Lixuan He, Chang Zou, Peiliang Cai, Jiacheng Liu, and Linfeng Zhang.
\newblock Let features decide their own solvers: Hybrid feature caching for diffusion transformers.
\newblock In \emph{International Conference on Learning Representations}, 2026{\natexlab{b}}.
\newblock arXiv:2510.04188.

\bibitem[Zheng et~al.(2026{\natexlab{c}})Zheng, Feng, Wang, Zhou, Cai, Zou, Liu, Lin, Chen, Ma, and Zhang]{foca2026}
Shikang Zheng, Liang Feng, Xinyu Wang, Qinming Zhou, Peiliang Cai, Chang Zou, Jiacheng Liu, Yuqi Lin, Junjie Chen, Yue Ma, and Linfeng Zhang.
\newblock Forecast then calibrate: Feature caching as {ODE} for efficient diffusion transformers.
\newblock In \emph{AAAI Conference on Artificial Intelligence}, 2026{\natexlab{c}}.
\newblock arXiv:2508.16211.

\bibitem[Zou et~al.(2025)Zou, Liu, Liu, Huang, and Zhang]{toca2025}
Chang Zou, Xuyang Liu, Ting Liu, Siteng Huang, and Linfeng Zhang.
\newblock Accelerating diffusion transformers with token-wise feature caching.
\newblock In \emph{International Conference on Learning Representations}, 2025.
\newblock arXiv:2410.05317.

\end{thebibliography}
